\documentclass{elsarticle}

\usepackage[english]{babel}
\usepackage{amsmath, amsfonts, amssymb, bm} 
\usepackage{graphicx}
\usepackage[letterpaper,top=2cm,bottom=2cm,left=3cm,right=3cm,marginparwidth=1.75cm]{geometry}
\usepackage{esvect}
\usepackage{physics}
\usepackage{afterpage}
\usepackage{float}
\usepackage{listings} %
\usepackage{subcaption} 
\usepackage[table,xcdraw]{xcolor} 
\usepackage{booktabs}
\usepackage{algpseudocode}
\usepackage{algorithm}
\usepackage[colorlinks=true, allcolors=blue]{hyperref}
\usepackage{tikz}
\usepackage{array} 
\usepackage{multirow} 
\usepackage{hhline}
\usepackage{colortbl}
\usepackage{xcolor}
\usepackage{makecell} 
\usepackage[normalem]{ulem}
\allowdisplaybreaks 
\usetikzlibrary{shapes.geometric, arrows}
\tikzstyle{startstop} = [rectangle, rounded corners, minimum width=3cm, minimum height=1cm,text centered, draw=black, fill=red!30]
\tikzstyle{io} = [trapezium, trapezium left angle=70, trapezium right angle=110, minimum width=3cm, minimum height=1cm, text centered, draw=black, fill=blue!30]
\tikzstyle{process} = [rectangle, minimum width=3cm, minimum height=1cm, text centered, draw=black, fill=orange!30]
\tikzstyle{decision} = [diamond, minimum width=3cm, minimum height=1cm, text centered, draw=black, fill=green!30]
\tikzstyle{arrow} = [thick,->,>=stealth]

\begin{document}
\begin{frontmatter}


\title{Optimizing the Optimizer for Physics-Informed Neural Networks and Kolmogorov-Arnold Networks}

\author[1]{Elham Kiyani}

\author[1]{Khemraj Shukla}

\author[2]{Jorge F. Urbán}

\author[1]{Jérôme Darbon}

\author[1]{George Em Karniadakis}

\address[1]{Division of Applied Mathematics, Brown University, 182 George Street, Providence, RI 02912, USA}

\address[2]{Departament de F\'{\i}sica Aplicada, Universitat d’Alacant, Ap. Correus 99, Alacant, 03830, Comunitat Valenciana, Spain}

\begin{abstract}
Physics-Informed Neural Networks (PINNs) have revolutionized the computation of PDE solutions by integrating partial differential equations (PDEs) into the neural network's training process as soft constraints, becoming an important component of the scientific machine learning (SciML) ecosystem. More recently, physics-informed Kolmogorv-Arnold networks (PIKANs) have also shown to be effective and comparable in accuracy with PINNs. In their current implementation, both PINNs and PIKANs are mainly optimized using first-order methods like Adam, as well as quasi-Newton methods such as BFGS and its low-memory variant, L-BFGS.
However, these optimizers often struggle with highly non-linear and non-convex loss landscapes, leading to challenges such as slow convergence, local minima entrapment, and (non)degenerate saddle points. 
In this study, we investigate the performance of Self-Scaled BFGS (SSBFGS), Self-Scaled Broyden (SSBroyden) methods and other advanced quasi-Newton schemes, including BFGS and L-BFGS with different line search strategies. These methods dynamically rescale updates based on historical gradient information, thus enhancing training efficiency and accuracy. We systematically compare these optimizers -- using both PINNs and PIKANs --  on key challenging PDEs, including the Burgers, Allen-Cahn, Kuramoto-Sivashinsky, Ginzburg-Landau, and Stokes equations. Additionally, we evaluate the performance of SSBFGS and SSBroyden for Deep Operator Network (DeepONet) architectures, demonstrating their effectiveness for data-driven operator learning. Our findings provide state-of-the-art results with orders-of-magnitude accuracy improvements without the use of adaptive weights or any other enhancements typically employed in PINNs. More broadly, our results reveal insights into the effectiveness of quasi-Newton optimization strategies in significantly improving the convergence and accurate generalization of PINNs and PIKANs.

Code is available at \url{https://github.com/EliKiani/Optimizing_the_Optimizer_PINNs}

%
\end{abstract}
\end{frontmatter}

\section{Introduction}

This section begins by providing the necessary background and motivation for our study, highlighting the key challenges and objectives. Following this, we present an overview of the BFGS (Broyden–Fletcher–Goldfarb–Shanno), SSBFGS (Self-Scaled BFGS), and SSBroyden (Self-Scaled Broyden) optimization methods, discussing their theoretical foundations and main characteristics.
Section~\ref{sec:comuptaional_exp} presents a comprehensive study comparing different optimization methods with various line search and trust-region strategies. Specifically, in Subsections~\ref{subsubsec:burger_tf} and~\ref{JAX_burgers}, we discuss a wide range of scenarios for the Burgers equation. We extend our study to more complex PDEs, including the Allen–Cahn equation in Section~\ref{sec:allen-cahn}, the Kuramoto–Sivashinsky equation in Subsection~\ref{subsec:Kuramoto-Sivashinsky}, and the Ginzburg–Landau equation in Subsection~\ref{Sec:Ginzburg-Landau}. 
To demonstrate the efficacy of the proposed optimizers, we also solve the Stokes equation for viscous flow in a lid-driven wedge in Subsection~\ref{sec:Stokes equation}. Furthermore, the performance of SSBFGS and SSBroyden is investigated for data-driven operator learning using Deep Operator Network (DeepONet) in Section~\ref{sec:deeponet}.
The conclusions of our study are summarized in Section~\ref{summary}. Finally, in Appendix~\ref{appendix:Lorenz system}, we review the performance of the optimizers on the Lorenz system. In Appendix~\ref{appendix:rosenbrock}, we provide a quantitative analysis of BFGS and SSBroyden on the multi-dimensional Rosenbrock function, serving as a pedagogical example to illustrate the effect of dimensionality on optimization performance.

\subsection{Background and Motivation}\label{subsec:Background_Motivation}

Physics-Informed Neural Networks (PINNs), introduced in 2017, are a groundbreaking development in scientific machine learning (SciML)~\cite{cuomo2022scientific, raissi2017machine}. By seamlessly integrating the fundamental physical principles of a system with neural networks, PINNs offer a versatile and mesh-free framework for solving nonlinear partial differential equations (PDEs). Unlike traditional numerical methods, PINNs directly incorporate initial and boundary conditions as well as PDE residuals into their loss functions, enabling them to address both forward problems (predicting solutions) and inverse problems (e.g.,estimating unknown parameters or unknown functions). Their adaptability and scalability make PINNs particularly well-suited for tackling challenges in high-dimensional spaces and complex geometries that conventional methods struggle to handle. 
The network parameters are updated during training to minimize the loss function, resulting in a solution that meets the constraints applied in the loss function. More recently, physics-informed Kolmogorov-Anrold Networks (PIKANs) were introduced in \cite{Liu2024KANKN,shukla2024comprehensive,ji2024comprehensive} to solve PDEs,
so in the present study, we aim to investigate both PINNS and PIKANs with respect to their optimization performance.

Since their introduction, PINNs have undergone numerous extensions and adaptations to enhance their applicability and address limitations in the original framework. These advancements include uncertainty quantification~\cite{zhang2019quantifying}, domain decomposition techniques~\cite{jagtap2020conservative,jagtap2020extended,shukla2021parallel,kiyani2023framework}, and the incorporation of alternative network architectures such as convolutional neural networks (CNNs)~\cite{gao2021phygeonet} and recurrent neural networks (RNNs)~\cite{ren2022phycrnet}.
Innovative approaches like Generative Adversarial PINNs (GA-PINNs)~\cite{li2022revisiting}, Physics-Informed Graph Convolutional Networks (GCNs)~\cite{peng2024data,rodrigo2024physics}, and Bayesian PINNs (B-PINNs)~\cite{yang2021b,graf2022error,kianiharchegani2023data,kiyani2024characterization} have further broadened the scope of PINNs. Physics-Informed Extreme Learning Machines (PIELM)~\cite{dwivedi2020physics} combine the computational efficiency of Extreme Learning Machines (ELMs) with the physics-informed principles of PINNs. The hp-VPINN method~\cite{kharazmi2021hp} integrates variational principles with neural networks, utilizing high-order polynomial test spaces for improved accuracy.

Extensive studies have also focused on error analysis and theoretical underpinnings of PINNs. Research on error estimates and convergence properties is presented in~\cite{de2022error,mishra2022estimates,shin2020convergence,shin20202convergence}. Wang et al.~\cite{wang2022respecting} proposed a reformulation of the PINN loss function to explicitly incorporate physical causality during training, although its performance was limited to simple benchmark problems. Additionally, Wang et al.~\cite{wang2022and} established a theoretical foundation by deriving and analyzing the limiting Neural Tangent Kernel (NTK) of PINNs. This analysis has been extended in subsequent studies to justify and refine various PINN extensions~\cite{WANGFou,McClennyBragaNeto2023,Bai2023,Jha2024}.
Furthermore, Anagnostopoulos et al.~\cite{Anagnostopoulos2024SNR} investigated the learning dynamics of PINNs using the Information Bottleneck theory~\cite{IBtheory}, identifying distinct training phases and proposing a loss weighting scheme to reduce generalization error. 

One of the greatest challenges in neural network frameworks lies in the inherently non-convex nature of their optimization problems. 
As a result, a growing body of research has been dedicated to understanding their training dynamics (see e.g., \cite{2020Lee, Cohen2021GradientDO, 2022arXiv220714484C}).
Optimization methods are broadly categorized into first-order (e.g., stochastic gradient descent)~\cite{robbins1951stochastic, jain2018parallelizing}, high-order (e.g., Newton’s method)~\cite{shanno1970conditioning, hu2019structured, dennis1977more}, and heuristic derivative-free approaches~\cite{larson2019derivative,kramer2011derivative}. Compared to first-order methods, high-order optimization achieves faster convergence by leveraging curvature information but faces challenges in handling and storing the Hessian's inverse.
To address this, Newton's method variants employ various techniques to approximate the Hessian matrix. For example, BFGS and L-BFGS (Limited-memory BFGS) approximate the Hessian or its inverse using rank-one or rank-two updates based on gradient information from previous iterations~\cite{nawi2006improved,saputro2017limited, BroydenBFGS,FletcherBFGS,Goldfarb1970AFO,ShannoBFGS,LiuNocedal}.

Currently, the standard optimization algorithms used in PINNs are Adam and quasi-Newton methods, such as BFGS or its low-memory variant, L-BFGS. BFGS and L-BFGS enhance significantly the performance of PINNs by incorporating second-order information of the trainable variable space to precondition it. Although these two optimizers are very commonly employed not only in PINNs but also in other optimization problems because of their theoretical and numerical properties, recent experience in PINNs has shown that other algorithms could outperform them for a great variety of problems~\cite{RLFLU24,GaussNewtonPINNs,urban2024unveiling}. Among all of them we can highlight the \emph{Self-Scaled Broyden} algorithms~\cite{BroydenMethodsReview}, which update the approximation to the inverse Hessian matrix using a self-scaling technique. This advanced optimization technique enhances convergence by adaptively scaling gradient updates based on historical error information.
Furthermore,~\cite{AlBaaliKhalfan2005} investigates self-scaling within quasi-Newton methods from the Broyden family, introducing innovative scaling schemes and updates. Urban et al.~\cite{urban2024unveiling} further demonstrated that these advanced optimizers, when combined with rescaled loss functions, significantly enhance the efficiency and accuracy of PINNs, enabling smaller networks to solve problems with significantly greater accuracy. As PINNs often suffer from conflicting gradients between multiple loss components (e.g., data and physics losses), resolving these conflicts is essential for stable and accurate training. Recent work by Wang et al.~\cite{wang2025gradient} demonstrates that quasi-Newton optimizers, particularly Newton-type methods like SOAP, exhibit positive gradient alignment and can effectively mitigate such conflicts, leading to improved convergence in multi-task optimization settings, like PINNs. 
Quasi-Newton optimization is also discussed in~\cite{jnini2024gauss}, where the authors present an implementation of a quasi-Newton optimizer and demonstrate improvements in both step time and wall-clock time across multiple large-scale tasks spanning diverse domains.

In the seminal paper by \cite{shin2020convergence}, PINN $(h_n)$ is formally described as a method that seeks to find a neural network minimizing a prescribed loss function $(\mathcal{L})$ within a class of neural networks $\mathcal{H}_n$, where $n$ represents the number of parameters in the network. The resulting minimizer serves as an approximation to the solution of the PDE. A crucial question posed by~\cite{shin2020convergence} is: Does this sequence of neural networks converge to the solution of the PDE? The answer to this question depends on the regularity of the loss functions and the three types of errors observed during the network's training or optimization process.

For completeness, we define these three errors, which also serve as one of the motivations for this work. The errors, as outlined in~\cite{niyogi1999generalization, bottou2007tradeoffs}, are: 
\begin{enumerate}
\item Approximation error.  
\item  Estimation error.  
\item Optimization error.  
\end{enumerate}
For example, consider a function class \(\mathcal{H}_n\) and let \(u^*\) be the solution to the underlying PDE. Let \(m\) represent the number of training data points. The function \(h_m\) is the minimizer of the loss with \(m\) data points, while \(\hat{h}\) is the function in \(\mathcal{H}_n\) that minimizes the loss with infinitely many data points. The error between \(h_m\) and \(u^*\) represents the \textit{approximation error}, while the error between \(\hat{h}\) and \(u^*\) is the \textit{estimation error}. In practice, \(\tilde{h}_m\) denotes the approximation obtained after a finite number of training iterations, such as the result of one million iterations of gradient-based optimization. The error between \(\tilde{h}_m\) and \(\hat{h}\) is referred to as the \textit{optimization error}. While the approximation error is well understood, the optimization error remains a challenging area due to the highly nonlinear and nonconvex nature of the objective function. These challenges are exacerbated by the inclusion of PDE terms in the loss function, leading to degenerate and non-degenerate saddle points. As a result, optimization often requires ad hoc tricks and tedious parameter fine-tuning through trial and error. 

\begin{figure}[!tbh]
\centering
 \includegraphics[width=\linewidth]{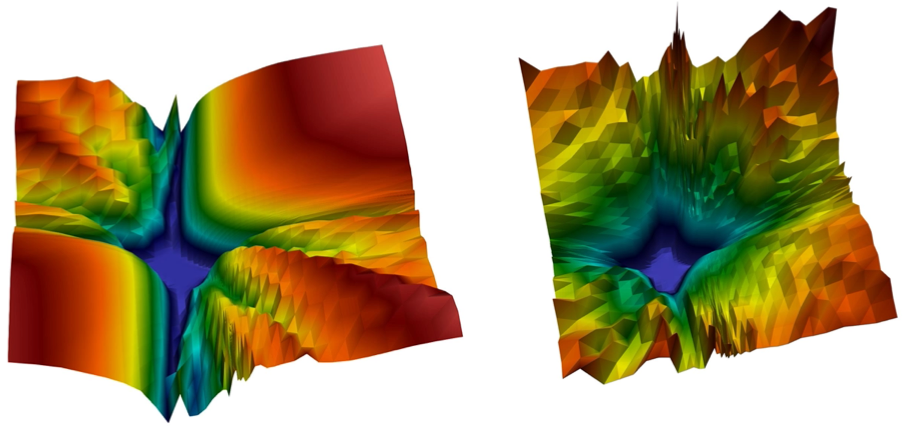}
\caption{Projection of the loss landscape \(\mathcal{L} = \mathcal{L}_{\text{Data}} + \lambda \mathcal{L}_{\text{PDE}}\) for the viscous Burgers equation (\ref{eq:burgers}) onto a 2D parameter subspace extracted from SVD of global parameter space,  as described in \cite{li2018visualizing}. The vertical axis represents the scalar loss value $\mathcal{L}(\theta)$, while the horizontal axes correspond to randomly chosen linear combinations of network parameters.
In the left subfigure, the landscape is shown for $\lambda = 0$, where only data loss is used. The resulting landscape is relatively smooth and convex but fails to enforce the PDE constraints within the domain. In contrast, the right subfigure shows the landscape for $\lambda = 1.0$, where the PDE residual is included. This introduces a highly non-convex landscape with numerous local minima and saddle points, reflecting the increased stiffness and complexity introduced by the physics term.}
\label{fig:lls_pinn_nopde}
\end{figure}

To provide an intuitive explanation, Figure~\ref{fig:lls_pinn_nopde} illustrates the loss landscape \(\mathcal{L} = \mathcal{L}_D + \lambda \mathcal{L}_{PDE}\) for the viscous Burgers Equation~\eqref{eq:burgers}, using the method proposed by \cite{li2018visualizing}. The left subfigure of Figure~\ref{fig:lls_pinn_nopde} corresponds to \(\lambda = 0\), considering only the initial and boundary conditions. This loss landscape is smooth and exhibits well-defined convexity. However, it fits only the initial and boundary condition data and fails to converge to the solution within the domain. In contrast, the right subfigure depicts the loss landscape for \(\lambda = 1.0\). Unlike the smooth landscape on the left, this landscape is crinkled with numerous local minima and saddle points, including potentially degenerate and non-degenerate saddle points~\cite{kawaguchi2016deep}. Consequently, the presence of multiple local minima can trap the optimizer, making it extremely challenging to minimize the loss function effectively. Motivated by this, in this work, we propose and analyze a suite of quasi-Newton optimizers and their applicability to linear, nonlinear, stiff, and chaotic PDEs whose solutions are computed by PINN and PIKAN types of approximation. Furthermore, we introduce a set of optimizers tailored to various classes of PDEs, achieving unprecedented convergence to machine precision with both single and double precision arithmetic. To the best  of our knowledge, the current study is the most systematic and comprehensive investigation of how optimizers affect the performance of both PINNs and PIKANs.
\subsection{Overview of BFGS and SSBroyden}\label{subsec:Overview_BFGS_SSBroyden}
In optimization, a fundamental challenge is determining the optimal direction and step size to transition from the current point \( \mathbf{x}_k \) to an improved solution. In general, two primary approaches address this problem: \textbf{ line-search methods} and \textbf{ trust-region methods}. Both rely on a quadratic model to approximate the objective function around the current iterate. The key distinction lies in how they utilize this approximation; line-search methods determine the step size along a chosen direction, while trust-region methods restrict the step to a pre-defined neighborhood where the model is considered reliable. 
The details of the line-search and trust-region methods are provided in~\ref{search_methods_trust-region_techniques}.

\subsubsection{Quasi-Newton methods }
A well-known example of a line-search direction is the steepest descent direction, which is given by the negative gradient at \( \mathbf{x}_k \), that is,

\[
    \mathbf{p}_k = -\nabla f_k,
\]

which ensures locally the steepest decrease in \( f \) at \( \mathbf{x}_k \). Another well-known example is Newton's method, which stands out for its remarkable quadratic rate of convergence near the solution. By leveraging both first- and second-order derivatives of the function, Newton’s method computes the following search direction \( \mathbf{p}_k \) by
\[
    \mathbf{p}_k = -(\nabla^2 f_k)^{-1} \nabla f_k,
\]
where 
\( \mathbf{H}_k \approx \left( \nabla^2 f_k \right)^{-1} \)
is the Hessian matrix evaluated at the current iterate \( \mathbf{x}_k \). In both cases, the search direction is then used to determine an appropriate step length \( \alpha \), ensuring a sufficient decrease in \( f \). 

Since Newton’s search direction is derived from minimizing a local quadratic approximation of the objective function, it is highly efficient near the solution where the loss landscape is approximately convex and well-behaved. This quadratic model enables rapid convergence, often in just a few iterations, especially when the Hessian accurately captures local curvature. However, Newton’s method requires computing and inverting the Hessian matrix, which becomes computationally prohibitive in high-dimensional settings, such as those encountered in deep learning or PINNs. Moreover, when the current iterate is far from the solution, the Hessian may be indefinite or poorly conditioned, leading to search directions that are not guaranteed to be descent directions and may even increase the loss.
In contrast, the steepest descent (or gradient descent) method avoids the need for second-order information by using the negative gradient as the update direction. This makes each iteration relatively inexpensive and guarantees a descent direction at every step. However, its efficiency drastically deteriorates in the presence of \emph{ill-conditioned}, which arises when the Hessian has eigenvalues that differ by several orders of magnitude. In such cases, the optimization trajectory may exhibit zigzagging behavior along the narrow valleys of the loss surface, causing slow convergence. This is particularly problematic in multi-objective or physics-informed settings, where different loss terms may induce conflicting gradients and anisotropic curvature. As a result, the steepest descent method often struggles in scenarios where curvature-aware updates, like those in quasi-Newton or second-order methods are essential for efficient and stable convergence~\cite{wang2025gradient}.

To balance computational efficiency with convergence performance, quasi-Newton methods have emerged as an attractive alternative to full Newton’s method. These approaches iteratively approximate the Hessian matrix using only first-order derivative information, thereby avoiding the explicit computation of second derivatives.   Specifically, quasi-Newton methods achieve superlinear convergence by updating an approximation \( \mathbf{B}_k \) of the Hessian matrix, where $\mathbf{H}_k \equiv \mathbf{B}_k^{-1}$, using gradient information from previous iterations. These methods rely on update formulas, which ensure that the updated approximation satisfies the secant equation:

\begin{equation}
    \mathbf{B}_{k+1} \mathbf{s}_k = \mathbf{y}_k,
\end{equation}

where \( \mathbf{s}_k = \mathbf{x}_{k+1} - \mathbf{x}_k \) represents the change in the iterates, and \( \mathbf{y}_k = \nabla f_{k+1} - \nabla f_k \) represents the corresponding change in gradients.
The specific form of the update rule defines the type of quasi-Newton method employed. Notable examples include the Symmetric Rank-One (SR1) and the widely used BFGS (Broyden–Fletcher–Goldfarb–Shanno) methods.
Among these, the BFGS method stands out due to its ability to maintain symmetry and ensure positive definiteness of the Hessian approximation under mild conditions. This makes BFGS one of the most robust and commonly used quasi-Newton techniques.
Once the search direction is determined—typically defined as
%
\begin{equation}\label{eq:pk_bfgs}
    \mathbf{p}_k = -\mathbf{H}_k \nabla f_k,
\end{equation}
%
%
a suitable step size $\alpha_k$ is selected (often via a line search strategy).  The next iterate is then updated as $\mathbf{x}_{k+1} = \mathbf{x}_k + \alpha_k \mathbf{p}_k$ followed by computing the $\textbf{s}_k$ and $\textbf{y}_k$, and finally updating the Hessian approximation using the the matrix \( \mathbf{B}_k \) as follows

\begin{equation}\label{eq:Bkp1BFGS}
    \mathbf{B}_{k+1} = \mathbf{B}_k - \frac{\mathbf{B}_k \mathbf{s}_k \mathbf{s}_k^{\top}  \mathbf{B}_k}{\mathbf{s}_k^{\top} \mathbf{B}_k \mathbf{s}_k} + \frac{\mathbf{y}_k \mathbf{y}_k^{\top} }{\mathbf{y}_k^{\top}  \mathbf{s}_k}. \tag{2.19}
\end{equation}

\subsection{Broyden Family of Optimizers}

Many quasi-Newton methods fall under the Broyden family of updates, characterized by the following general formula:

\begin{equation}\label{eq:Bkp1}
    \mathbf{B}_{k+1} = \mathbf{B}_k - \frac{\mathbf{B}_k \mathbf{s}_k \mathbf{s}_k^{\top} \mathbf{B}_k}{\mathbf{s}_k^{\top} \mathbf{B}_k \mathbf{s}_k} + \frac{\mathbf{y}_k \mathbf{y}_k^{\top}}{\mathbf{y}_k^{\top} \mathbf{s}_k} + \theta_k (\mathbf{s}_k^{\top} \mathbf{B}_k \mathbf{s}_k) \mathbf{w}_k \mathbf{w}_k^{\top},
\end{equation}

where \( \theta_k \) is a scalar parameter that may vary at each iteration, and

\begin{equation*}
    \mathbf{w}_k = \frac{\mathbf{y}_k}{\mathbf{y}_k^{\top} \mathbf{s}_k} - \frac{\mathbf{B}_k \mathbf{s}_k}{\mathbf{s}_k^{\top} \mathbf{B}_k \mathbf{s}_k}.
\end{equation*}

The BFGS and DFP methods are special cases of the Broyden class. Specifically, setting \( \theta_k = 0 \) recovers the BFGS update, while \( \theta_k = 1 \) yields the DFP update. 
In practice, quasi-Newton methods typically work directly with \( \mathbf{H}_k \), avoiding explicit matrix inversion. By applying the inverse to both sides of Equation \eqref{eq:Bkp1}, and using the Sherman–Morrison formula, we obtain the following update for \( \mathbf{H}_k \):

\begin{equation}\label{eq:Hkp1}
    \mathbf{H}_{k+1} = \mathbf{H}_k - \frac{\mathbf{H}_k \mathbf{y}_k \mathbf{y}_k^{\top} \mathbf{H}_k}{\mathbf{y}_k^{\top} \mathbf{H}_k \mathbf{y}_k} + \frac{\mathbf{s}_k \mathbf{s}_k^{\top}}{\mathbf{y}_k^{\top} \mathbf{s}_k} + \phi_k (\mathbf{y}_k^{\top} \mathbf{H}_k \mathbf{y}_k) \mathbf{v}_k \mathbf{v}_k^{\top},
\end{equation}

where the intermediate quantities are defined as:

\begin{align*}
    \mathbf{v}_k &= \frac{\mathbf{s}_k}{\mathbf{y}_k^{\top} \mathbf{s}_k} - \frac{\mathbf{H}_k \mathbf{y}_k}{\mathbf{y}_k^{\top} \mathbf{H}_k \mathbf{y}_k}, \\
    \phi_k &= \frac{1 - \theta_k}{1 + (h_k b_k - 1) \theta_k}, \\
    b_k &= \frac{\mathbf{s}_k^{\top} \mathbf{B}_k \mathbf{s}_k}{\mathbf{y}_k^{\top} \mathbf{s}_k}, \\
    h_k &= \frac{\mathbf{y}_k^{\top} \mathbf{H}_k \mathbf{y}_k}{\mathbf{y}_k^{\top} \mathbf{s}_k}.
\end{align*}

Additional useful expressions include:

\begin{align*}
    a_k &= b_k h_k - 1, \\
    c_k &= \left( \frac{a_k}{1 + a_k} \right)^{1/2}, \\
    \rho_k^- &= \min\left(1, h_k(1 - c_k)\right), \\
    \theta_k^- &= \frac{\rho_k^- - 1}{a_k}, \\
    \theta_k^+ &= \frac{1}{\rho_k^-}, \\
    \theta_k &= \max\left(\theta_k^-, \min\left(\theta_k^+, \frac{1 - b_k}{b_k} \right)\right), \\
    \rho_k^+ &= \min\left(1, \frac{1}{b_k}\right), \\
    \sigma_k &= 1 + \theta_k a_k, \\
    \sigma_k^{(1-N)} &= |\sigma_k|^{\frac{1}{1 - N}}, \\
    \tau_k &=
    \begin{cases}
        \min\left(\rho_k^+ \sigma_k^{(1-N)}, \sigma_k \right), & \text{if } \theta_k \leq 0, \\[6pt]
        \rho_k^+ \min\left(\sigma_k^{(1-N)}, \frac{1}{\theta_k} \right), & \text{otherwise}.
    \end{cases}
\end{align*}

If \( \theta_k \) depends explicitly on \( \mathbf{B}_k \), then both equations \eqref{eq:Bkp1} and \eqref{eq:Hkp1} must be used at each iteration to update the inverse Hessian estimates. However, when \( \mathbf{B}_k \) appears only through the product \( \mathbf{B}_k \mathbf{s}_k \), the update can avoid direct use of \eqref{eq:Bkp1}, since:

\begin{equation}
    \mathbf{B}_k \mathbf{s}_k = -\alpha_k \nabla f_k.
\end{equation}

Among the methods in the Broyden class, the BFGS method is particularly effective for small- and medium-scale unconstrained optimization problems \cite{Nocedal_1992, Fletcher1994}. However, its performance may deteriorate for ill-conditioned problems \cite{Powell1986, Byrd1992}. To address this limitation, the \textit{self-scaled} BFGS (SSBFGS) method was proposed by Oren and Luenberger \cite{OrenLuenberger}. In SSBFGS, the Hessian approximation \( \mathbf{B}_k \) is scaled by a positive scalar \( \tau_k \) prior to the BFGS update, i.e.,

\begin{equation}
    \mathbf{B}_{k+1} = \tau_k \left[ \mathbf{B}_k - \frac{\mathbf{B}_k \mathbf{s}_k \mathbf{s}_k^{\top} \mathbf{B}_k}{\mathbf{s}_k^{\top} \mathbf{B}_k \mathbf{s}_k} \right] + \frac{\mathbf{y}_k \mathbf{y}_k^{\top}}{\mathbf{y}_k^{\top} \mathbf{s}_k}.
\end{equation}

The motivation behind scaling is to reduce the condition number of \( \mathbf{H}_k^{1/2} \nabla^2 f_k \mathbf{H}_k^{1/2} \), which reflects the convergence rate. While initial results by Nocedal and Yuan \cite{Nocedal1993} were discouraging for the scaling \( \tau_k = 1/b_k \), later work by Al-Baali \cite{AlBaali1993} showed promising results using the modified scaling \( \tau_k = \min \{1, 1/b_k\} \).
Al-Baali \cite{Al-Baali1998} extended the idea of scaling to other Broyden updates, proving favorable theoretical and numerical results for various \( \theta_k \in [0, 1] \), while previously assuming \( \tau_k \leq 1 \). Additional scaling strategies have been developed for BFGS and other Broyden methods in works such as \cite{Contreras1993, Luksan1994, Wolkowicz1995, AlBaali2007, Yabe2007}.
The self-scaled versions of the direct and inverse updates for general Broyden family methods—termed \textit{SSBroyden} methods—are:

\begin{align}
    \mathbf{B}_{k+1} &= \tau_k \left[ \mathbf{B}_k - \frac{\mathbf{B}_k \mathbf{s}_k \mathbf{s}_k^{\top} \mathbf{B}_k}{\mathbf{s}_k^{\top} \mathbf{B}_k \mathbf{s}_k} + \theta_k (\mathbf{s}_k^{\top} \mathbf{B}_k \mathbf{s}_k) \mathbf{w}_k \mathbf{w}_k^{\top} \right] + \frac{\mathbf{y}_k \mathbf{y}_k^{\top}}{\mathbf{y}_k^{\top} \mathbf{s}_k}, \\
    \mathbf{H}_{k+1} &= \frac{1}{\tau_k} \left[ \mathbf{H}_k - \frac{\mathbf{H}_k \mathbf{y}_k \mathbf{y}_k^{\top} \mathbf{H}_k}{\mathbf{y}_k^{\top} \mathbf{H}_k \mathbf{y}_k} + \phi_k (\mathbf{y}_k^{\top} \mathbf{H}_k \mathbf{y}_k) \mathbf{v}_k \mathbf{v}_k^{\top} \right] + \frac{\mathbf{s}_k \mathbf{s}_k^{\top}}{\mathbf{y}_k^{\top} \mathbf{s}_k}.
\end{align}

In the context of Physics-Informed Neural Networks (PINNs) based on Multi-Layer Perceptrons (MLPs), recent research has begun to explore various families of SSBroyden updates and modifications to the loss formulation to address known challenges such as gradient conflicts and ill-conditioning. These second-order strategies significantly enhance both convergence speed and prediction accuracy compared to classical BFGS.
In this work, we extend these findings to more complex problems beyond those presented in~\cite{urban2024unveiling}, employing not only SSBroyden but also other techniques proposed in the PINN literature. Additionally, we investigate whether these BFGS modifications enhance the performance of the recently introduced Kolmogorov-Arnold Networks (KANs)~\cite{Liu2024KANKN}. This study will further enable fair comparisons between KANs and MLPs using more advanced optimization algorithms.

In addition to PINNs, we also evaluate the performance of quasi-Newton optimization methods in the context of Deep Operator Networks (DeepONets)~\cite{lu2019deeponet}. DeepONet is a neural architecture designed to learn nonlinear operators—mappings between infinite-dimensional function spaces—and has shown strong performance in solving parametric PDEs in purely data-driven settings. While PINNs incorporate the governing physics directly into the loss function, DeepONets rely solely on data to learn the underlying operator. We demonstrate that the benefits of quasi-Newton methods are not limited to physics-informed frameworks, but also extend to operator learning tasks. This inclusion broadens the impact of our work by showing that improved optimization strategies can enhance training stability and accuracy across diverse neural PDE solvers. This extension allows us to evaluate whether  quasi-Newton optimizers like SSBroyden can consistently improve training performance across different PDE learning paradigms.

In the following section, we present a comprehensive study comparing the performance of different optimization methods with various line-search and trust-region strategies for solving PDEs. For each example, we analyze and discuss the performance of both PINNs and PIKANs.

\section{Computational 
Experiments}\label{sec:comuptaional_exp}
In this section, we conduct various computational experiments to evaluate the performance of optimizers across a diverse range of steady-state and time-dependent PDEs. To ensure a comprehensive assessment, we select PDEs that encompass a wide spectrum of classes, including parabolic, hyperbolic, elliptic, and hyperbolic-parabolic equations. 
It is worth noting that the computations for BFGS and SSBroyden with Wolfe line-search are performed on an NVIDIA RTX A6000 GPU, while the computations for BFGS with backtracking and trust-region methods are conducted on an NVIDIA RTX 3090 GPU.
\subsection{Burgers equation}\label{subsec:burgereq}
To begin with, we consider the viscous Burgers' equation in a spatially periodic domain,  given by,
\begin{equation}\label{eq:burgers}
\centering
    \frac{\partial u}{\partial t} + u \frac{\partial u}{\partial x} = \nu \frac{\partial^2 u}{\partial x^2},
\end{equation}
with an initial condition 
\begin{equation}
\centering
    u(x, 0) = -\sin(\pi x)
\end{equation}
and periodic boundary conditions. The viscosity is given by $\nu = \frac{0.01}{\pi} \approx 0.003$, and the spatio-temporoal domain for computing the solution is \((x,~t) \in [-1, 1] \times [0, 1]\).
In the following subsections, we present the performance of L-BFGS, BFGS, SSBFGS, and SSBroyden optimizers using Strong Wolfe line-search, backtracking, and trust-region strategies (see Subsections~\ref{subsubsec:burger_tf} and~\ref{JAX_burgers}). All results in these sections are reported in double precision. To provide a more comprehensive study, we also include single-precision results in \ref{Burgers_Equation_Single_Precision}.

\subsubsection{BFGS and SSBroyden optimization with Strong Wolfe line-search}\label{subsubsec:burger_tf}
A comparative study of the performance of L-BFGS, BFGS, SSBFGS and SSBroyden optimization with Strong Wolfe line-search for solving the Burgers' Equation has been conducted using seven different case studies. In all PINNs cases, the hidden layers utilize the hyperbolic tangent (\( \tanh \)) activation function. Model training is performed in two stages: first, the Adam optimizer is employed with a learning rate of \( 10^{-3} \) for 1000 iterations.  Moreover, the adaptive sampling strategy (RAD) algorithm defined in \cite{WuSampling2023} is employed to dynamically resample points in regions with high errors.
%
%

\begin{figure}[!tbh]
    \centering
    \begin{minipage}[b]{\linewidth}
        \centering
        \includegraphics[width=\linewidth]{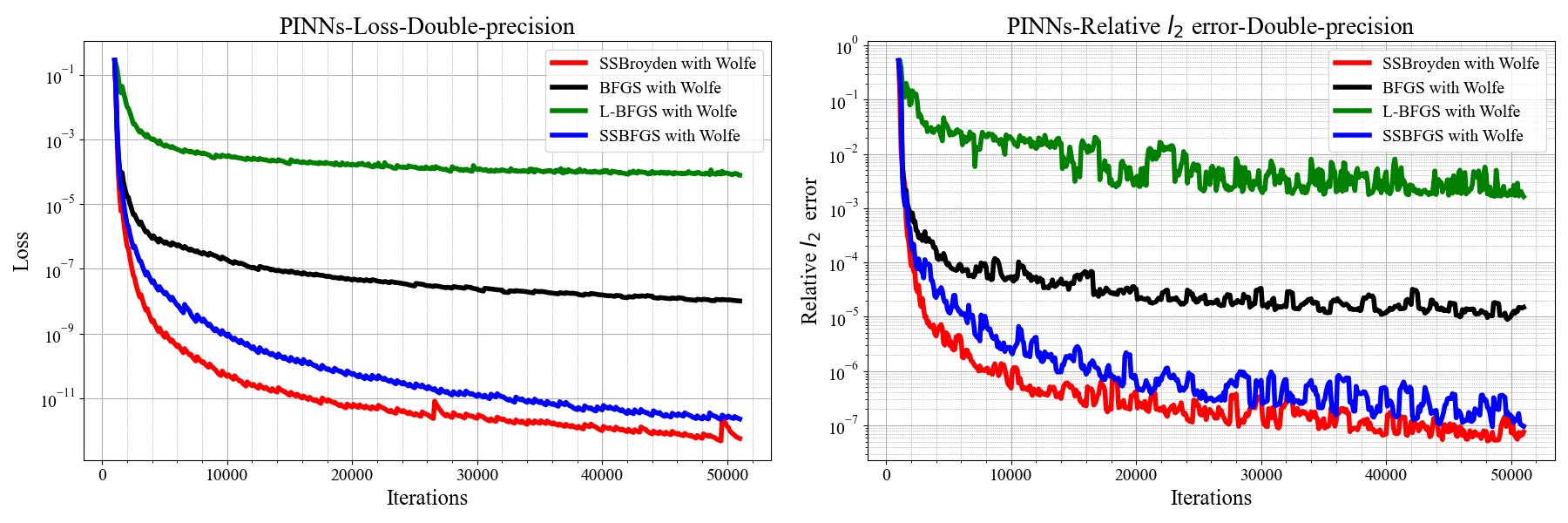}
        \caption{\textbf{Double-Precision PINNs for the Burgers equation}: The evolution of the loss function (left) and corresponding \( l_2 \) relative errors (right) over iterations for \textbf{Case 1}. }
        \label{fig:PINNs_burger_double_combined}
    \end{minipage}
    \vspace{1em}
    \begin{minipage}[b]{\linewidth}
        \centering
        \rowcolors{2}{cyan!15}{white}
        \scalebox{0.75}{
        \begin{tabular}{|c|c|c|c|c|}
        \hline
        \rowcolor{cyan!40} 
        \textbf{Case}  & \textbf{Optimizer[\# Iters.], Line-search [\#Iters.]}     & \textbf{Relative $l_2$ error} & \textbf{Training time (s)} & \textbf{Total params} \\ \hline
        1 & Adam [1000] + BFGS with Wolfe [50000]                 & \( 1.50 \times 10^{-5} \)  & 1292 &  1,341  \\ \hline
        1 & Adam [1000] + SSBroyden with Wolfe [50000]            & \( 7.57 \times 10^{-8} \)  & 1354 &  1,341 \\ \hline
        1 & Adam [1000] + SSBFGS with Wolfe [50000]               & \( 9.62 \times 10^{-8} \)  & 1293  &  1,341 \\ \hline
        1 & Adam [1000] + L-BFGS with Wolfe [50000]               & \( 2.05 \times 10^{-3} \)  & 713  &  1,341 \\ \hline
        2 & Adam [1000] + BFGS with Wolfe [30000]                 & \( 2.21 \times 10^{-5} \)  & 774  &  1,341 \\ \hline
        2 & Adam [1000] + SSBroyden with Strong Wolfe [30000]            & \( 2.12 \times 10^{-7} \)  & 819  &  1,341 \\ \hline
        3 &  BFGS with Wolfe [30000]                              & \( 1.73 \times 10^{-5} \)  & 863  &  1,341 \\ \hline
        3 &  SSBFGS with Wolfe [30000]                              & \( 1.54 \times 10^{-7} \)  & 711  &  1,341 \\ \hline
        3 &  SSBroyden with  Wolfe [30000]                         & \( 3.59 \times 10^{-7} \)  & 862  &  1,341 \\ \hline
        4 & Adam [1000] + BFGS with Wolfe [30000]                 & \( 8.52 \times 10^{-6} \)  & 3049 &  3,021 \\ \hline
        4 & Adam [1000] + SSBroyden with Wolfe [30000]            & \( 1.62 \times 10^{-8} \)  & 2812 &  3,021 \\ \hline
        4 & Adam [1000] + SSBFGS with Wolfe [30000]            & \( 4.19 \times 10^{-8} \)  & 2878 &  3,021 \\ \hline
        \end{tabular}}
        \captionof{table}{\textbf{Double-Precision PINNs for the Burgers equation}: 
        \textbf{Case 1:} A PINN with four layers, each containing 20 neurons, is trained for 1000 iterations using the Adam optimizer, followed by 50,000 iterations with BFGS, SSBroyden, SSBFGS, and L-BFGS optimizers with Strong Wolfe line-search.  
        \textbf{Case 2:} The same network structure as \textbf{Case 1} is used, but the number of iterations with BFGS,  and SSBroyden is reduced to 30,000.  
        \textbf{Case 3:} The same network structure as in \textbf{Case 1} and \textbf{Case 2} is used; however, instead of starting with the Adam optimizer, the training begins directly with quasi-Newtonoptimizers, namely BFGS, SSBFGS, and SSBroyden.
        \textbf{Case 4:} A deeper PINN with eight layers (20 neurons per layer) is trained for 1000 iterations using Adam, followed by 30,000 iterations with BFGS and SSBroyden.}
        \label{tab:PINNs_burger_double_precision}
    \end{minipage}
\end{figure}

\begin{figure}[!tbh]
    \centering
    \begin{minipage}[b]{\linewidth}
        \centering
        \includegraphics[width=\linewidth]{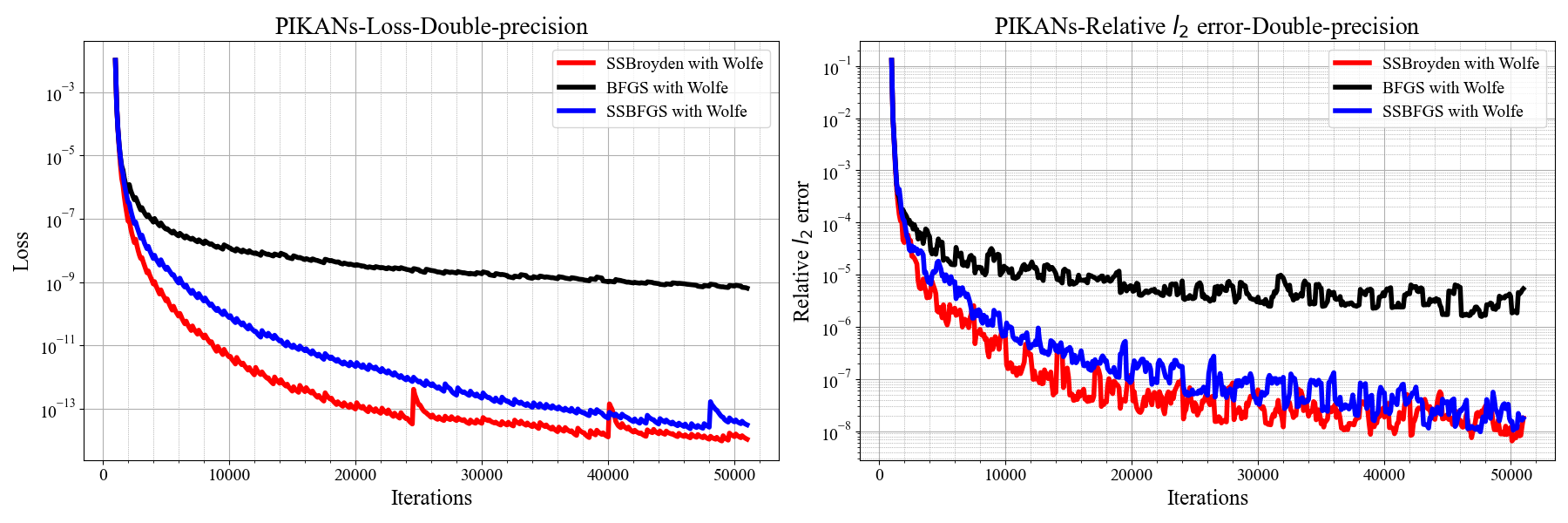}
        \caption{\textbf{Double-Precision PIKANs for the Burgers Equation}: Evolution of the loss function (left) and \( l_2 \) relative errors (right) for \textbf{Case 5}. The results compare the performance of PIKANs trained with SSBroyden and BFGS optimizers.}
        \label{fig:PIKANs_burger}
    \end{minipage}
    \vspace{1em}
    \begin{minipage}[b]{\linewidth}
        \centering
        \rowcolors{2}{cyan!15}{white}
        \scalebox{0.75}{
        \begin{tabular}{|l|r|r|r|r|}
        \hline
        \rowcolor{cyan!40} 
        \textbf{Case}  &  \textbf{Optimizer[\# Iters.], Line-search [\#Iters.]} & \textbf{Relative $l_2$ error} & \textbf{Training time (s)} & \textbf{Total params} \\ \hline
        5 & Adam [1000] + BFGS with Wolfe [50000] (degree 3)  & \( 5.38 \times 10^{-6} \)  & 9276  & 5,040 \\ 
        5 & Adam [1000] + SSBFGS with Wolfe [50000] (degree 3)  & \( 1.77 \times 10^{-8} \)  & 12796  & 5,040 \\ 
        5 & Adam [1000] + SSBroyden with Wolfe [50000] (degree 3) & \( 1.79 \times 10^{-8} \)  & 14169 & 5,040 \\ 
        6 & Adam [1000] + BFGS with Wolfe [50000] (degree 3)  & \( 3.18 \times 10^{-5} \)  & 891 & 920 \\ 
        6 & Adam [1000] + SSBroyden with Wolfe [50000] (degree 3)  & \( 2.41 \times 10^{-6} \)  & 849 & 920 \\ 
        6 & Adam [1000] + SSBFGS with Wolfe [50000] (degree 3)  & \( 4.21 \times 10^{-6} \)  & 810  & 920 \\ 
        7 & Adam [1000] + BFGS with Wolfe [50000] (degree 5)  & \( 4.29 \times 10^{-5} \)  & 1364 & 1,380 \\ 
        7 & Adam [1000] + SSBroyden with Wolfe [50000] (degree 5) & \( 1.08 \times 10^{-6} \)  & 1465 & 1,380 \\ 
        \hline
        \end{tabular}}
        \captionof{table}{\textbf{Double-Precision PIKANs for the Burgers Equation}: Relative \( l_2 \) error and training time for solving the Burgers equation using \textbf{double precision} with different optimizers. \textbf{Case 5} highlights a PIKAN with four layers of 20 neurons each and degree 3. A PIKAN with three layers, each containing 10 neurons, is trained using Chebyshev polynomials of degree 3 in \textbf{Case 6} and degree 5 in \textbf{Case 7}. The training consists of 1000 iterations with the Adam optimizer, followed by 50,000 iterations with BFGS and SSBroyden optimizers. \textbf{Case 5} demonstrates improved performance despite a significant increase in training time.}
        \label{tab:PIKANs_burger}
    \end{minipage}
\end{figure}

The performance of PINNs for \textbf{Case 1} and \textbf{Case 2} is presented in Table~\ref{tab:PINNs_burger_double_precision}. Both cases use the same network architecture, consisting of four hidden layers with 20 neurons each, and are evaluated in double precision. After the initial Adam optimization stage, the training proceeds with BFGS, SSBFGS, and SSBroyden optimizers for two different iteration counts: \textbf{Case 1} utilizes 50,000 iterations, while \textbf{Case 2} uses 30,000 iterations.  In \textbf{Case 3}, the same network structure as in \textbf{Case 1} and \textbf{Case 2} is used; however, training begins directly with quasi-Newton optimizers, BFGS, SSBFGS, and SSBroyden, instead of the Adam optimizer.
The results, as shown in the table, demonstrate that increasing the iteration count to 50,000 significantly raises the training time. While this reduces SSBFGS and SSBroyden's error to \(10^{-8}\), it does not result in any noticeable improvement in the error achieved by BFGS. 
In \textbf{Case 4}, a deeper network with eight hidden layers is used, trained for 30,000 iterations. This configuration further improves the error, though at the cost of a substantial increase in training time.  
Figure~\ref{fig:PINNs_burger_double_combined} illustrates the loss function and relative error over iterations for \textbf{Case 1} using four optimizers, BFGS, SSBFGS, SSBroyden, and L-BFGS. As summarized in Table~\ref{tab:PINNs_burger_double_precision}, the relative errors achieved are \(10^{-8}\) for SSBroyden, \(10^{-5}\) for BFGS, and \(10^{-3}\) for L-BFGS.

Additionally, Figure~\ref{fig:PIKANs_burger} compares the performance of BFGS and SSBroyden for PIKANs in \textbf{Case 5}. A PIKAN with four layers, each containing 20 neurons, is trained using Chebyshev polynomials of degree 3. As reported in Table~\ref{tab:PIKANs_burger}, the error in this case is very close to that of \textbf{Case 1} in Table~\ref{tab:PINNs_burger_double_precision}.  
Furthermore, a PIKAN with three layers, each containing 10 neurons, is trained using Chebyshev polynomials of degree 3 in \textbf{Case 6} and degree 5 in \textbf{Case 7}. The results demonstrate that SSBroyden consistently outperforms BFGS. Although the networks are trained for 50,000 iterations, minimal improvement is observed beyond approximately 30,000 iterations.  
As summarized in Table~\ref{tab:PIKANs_burger}, increasing the polynomial degree from 3 to 5 does not improve the results, while the training time nearly doubles with the higher degree.

The conclusion highlights that the best error for the Burgers equation was achieved using SSBroyden with double precision. Across all seven case studies, SSBroyden consistently outperformed both BFGS and L-BFGS. A comparison between PINNs and PIKANs reveals that, for the same number of parameters, PIKANs did not perform as well. However, when trained with a higher number of parameters—at the cost of significantly increased training time—PIKANs can achieve results comparable to PINNs. In general, a comparison between \textbf{Case 1} and \textbf{Case 5} shows that, although both achieve nearly identical errors, PINNs with SSBroyden demonstrate more efficient training times compared to PIKANs.

The performance of SSBroyden, SSBFGS, and BFGS using single-precision is presented in ~\ref{Burgers_Equation_Single_Precision}.
It is important to note that the previous tables compare the performance of different optimizers under a fixed iteration budget. To further assess the efficiency of our proposed quasi-Newton methods, Table~\ref{tab:threshold_comparison} presents results based on a fixed relative \(L_2\) error threshold of \(10^{-4}\). This setup enables a fair comparison of SSBFGS, SSBroyden, and BFGS in terms of both convergence speed and overall computational cost, independent of a predefined number of iterations. As shown in the table, both SSBFGS and SSBroyden reach the target accuracy significantly faster than standard BFGS, requiring fewer iterations and substantially less training time. Notably, SSBroyden achieves the target relative \(L_2\) errors of \(10^{-4}\) and \(10^{-5}\) in just 1,307 and 2,864 iterations, respectively, compared to 5,621 and 69,332 iterations for BFGS. This results in a more than threefold reduction in training time. BFGS exhibits slow convergence and eventual stagnation in the later stages, suggesting that it struggles to make further progress. In contrast, the self-scaled variants (SSBFGS and SSBroyden) achieve the target error more efficiently and consistently.

\begin{table}[ht]
\centering
\footnotesize
\rowcolors{2}{cyan!15}{white}
\begin{tabular}{|l|c|c|c|}
\hline
\rowcolor{cyan!40}
\textbf{Optimizer} & \textbf{Iterations} & \textbf{Final Rel. \(L_2\) Error} & \textbf{Training Time (s)} \\
\hline
SSBFGS with Wolfe     & 1,567  & \(9.91 \times 10^{-5}\)  & 167 \\
SSBroyden with Wolfe  & 1,307  & \(8.98 \times 10^{-5}\)  & 171 \\
BFGS with Wolfe       & 5,621  & \(9.79 \times 10^{-5}\)  & 586 \\
SSBFGS with Wolfe     & 4,766  & \(9.97 \times 10^{-6}\)  & 496 \\
SSBroyden with Wolfe  & 2,864  & \(9.74 \times 10^{-6}\)  & 310 \\
BFGS with Wolfe       & 69,332  & \(9.99 \times 10^{-6}\)  & 7189 \\
\hline
\end{tabular}
\caption{\textbf{Comparison of optimizers for the Burgers equation with a target relative \(L_2\) error of \(10^{-4}\) and \(10^{-5}\)}. A PINN with four layers, each containing 20 neurons, is trained directly using a quasi-Newton  optimizer. Each optimizer is run until the error threshold is reached. The table reports the number of iterations, final relative error, and total training time. Both SSBFGS and SSBroyden achieve the target accuracy significantly faster than standard BFGS.
}
\label{tab:threshold_comparison}
\end{table}

\subsubsection{Performance of optimzers based on composition operator in the JAX ecosystem}\label{JAX_burgers}
Approximating solutions of PDEs with neural networks involves randomly initializing the network parameters using a normal probability distribution \cite{glorot2010understanding}. The random sampling of the parameters depends on the random seeds, which also defines the reproducibility of the results. The framework presented above is based on Tensorflow, which generates the random number using a stateful random number generators. A stateful generator has the following three characteristics:  
\begin{enumerate}
\item It maintains an internal state that is updated after each random number generation.
\item The next random number depends on the current internal state.
\item The generator needs to be reseeded explicitly to reset or reproduce results.
\end{enumerate}
Therefore, the main drawback with stateful random number generators can result in non-neproducibility as the internal state can cause issues when working in parallel systems. Secondly, it requires careful management in multithreaded contexts to avoid race conditions, which is very common in GPU based architectures, as stateful random number generator consider previous state for future number generation.
To overcome these issues, the JAX framework~\cite{jax2018github} is based on a stateless random number generator which does not maintain any internal state between calls. Instead, it uses the provided input (such as a seed or key) to generate each random number independently, without relying on previous outputs. The state must be explicitly passed along with the call, and the RNG (Random Number Generator) does not store any internal data between calls. Especially, it becomes essential when true randomness or independent streams of numbers are required, e.g.,  parallel random number generation, GPU based programming models such as MPI + X $\in$ {OpenMP, CUDA, HIP, } etc.

\begin{figure}[!tbh]
\includegraphics[width=1\textwidth]{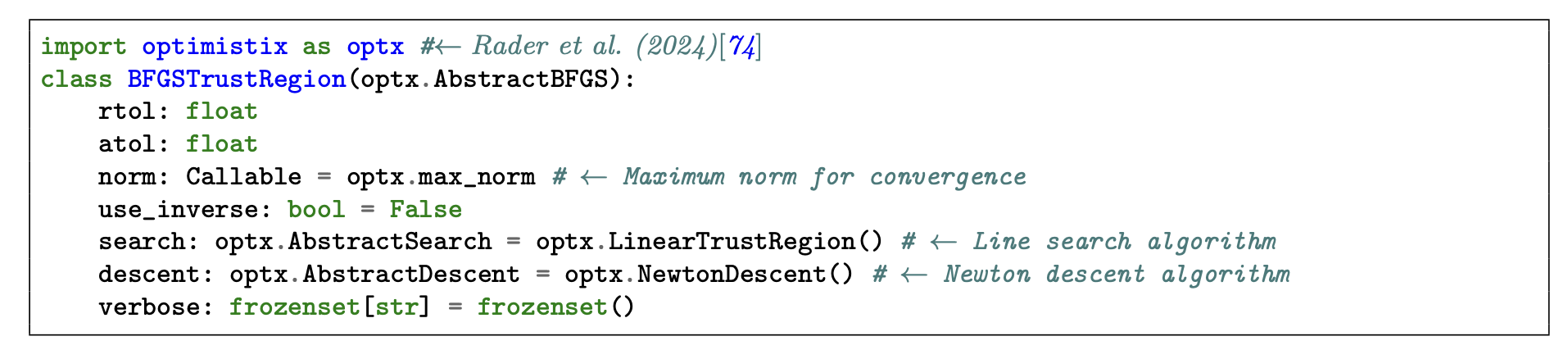}
\caption{A Python class for BFGS optimizers paired with trust-region line search and Newton descent. Equation~\eqref{eq:Bkp1BFGS} with $L_{\infty}$ norm chosen as convergence criteria.}
\label{code:BFGS line search}
\end{figure}

\begin{figure}[!tbh]
  \begin{minipage}[b]{\linewidth}
    \centering
    \includegraphics[trim={0cm 0 0 0},clip, width=\textwidth]{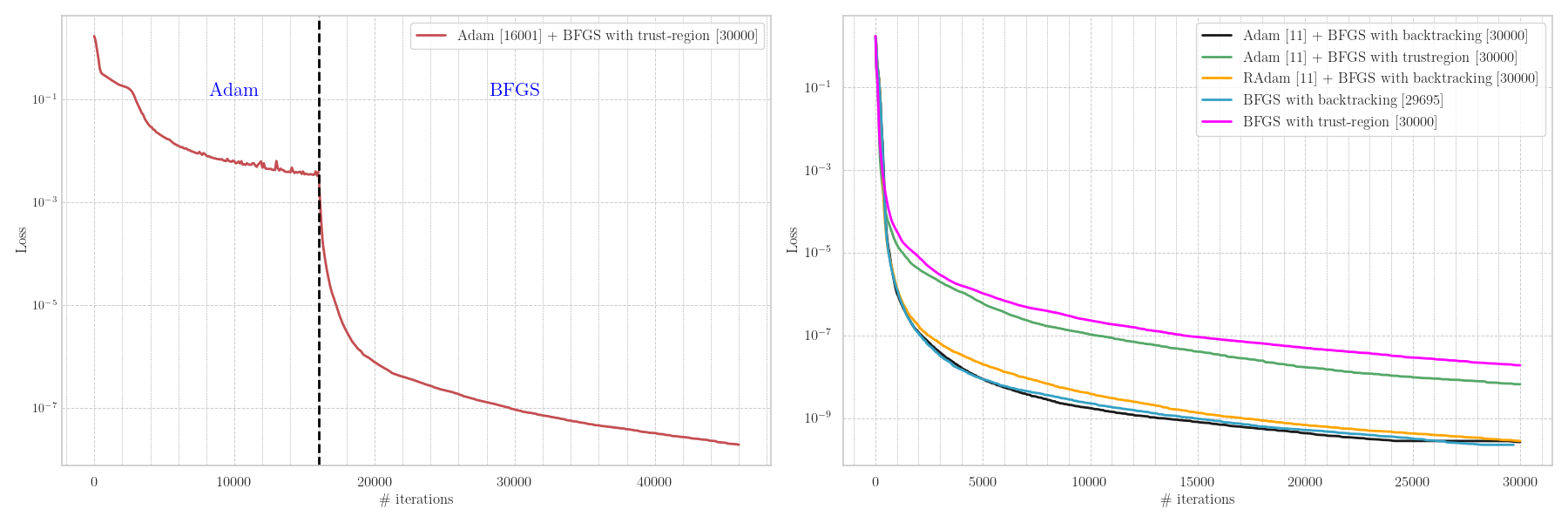}
        \captionof{figure}{\textbf{Scenario 1: Double-Precision PINNs for the Burgers Equation.} 
    Results are shown with backtracking and trust-region strategies. The GPU is fully saturated with model parameters to avoid the effect of latency on compute time. 
 The left panel illustrates a scenario where the iteration count of the first-order Adam algorithm dominates, and combined with BFGS using the trust-region approach. This pairing unlike Figure~\ref{fig:sp_optax_mp} does not exhibit degeneracy during BFGS iterations due to double-precision and converges to much better loss value. The right panel presents the convergence behavior of BFGS paired with trust-region and backtracking line search methods. Inline to the Figure~\ref{fig:sp_optax_mp}. in this setup as well, Adam is applied for 11 iterations as a warmup and then switched to BFGS algorithms. The BFGS-based optimizer requires significantly more iterations than first-order optimization methods like Adam and Rectified Adam (RAdam)~\cite{liu2019variance}. }
    \label{fig:dp_optax_mp}
  \end{minipage}\hfill
  \begin{minipage}[b]{\linewidth}
   \centering
    \rowcolors{2}{cyan!15}{white} 
    \scalebox{0.78}{
    \begin{tabular}{|c|c|c|c|}
    \hline
    \rowcolor{cyan!40} 
    \textbf{Optimizer[\# Iters.], Line-search algorithm [\#Iters.]} & \textbf{Relative $l_2$ error} & \textbf{Training time (s)} & \textbf{Total params} \\ 
    \hline
    Adam [16001] + BFGS with trust-region [30000] &  \(2.8 \times 10^{-5}\)     &  2030    & 3501              \\ 
    \hline
    Adam [11] + BFGS with backtracking [30000]  &  \(3.0\times 10^{-6}\)     & 1607        &    3501            \\ 
    \hline
    Adam [11] + BFGS with trust-region [30000] & \(1.3\times 10^{-5}\)   &  1726                & 3501              \\ 
    \hline
    RAdam [11] + BFGS with backtracking [30000]  &  \(\mathbf{2.0 \times 10^{-6}}\)    & 1587     & 3501               \\ 
    \hline
    BFGS with backtracking [29695]  & \(4.0\times 10^{-6}\)          & 1555        &  3501               \\ 
    \hline
    BFGS with trust-region [30000] & \(1.6\times 10^{-5}\)       & 1722                 & 3501              \\ 
    \hline
    \end{tabular}}
    \captionof{table}{[Corresponds to Figure~\ref{fig:dp_optax_mp}] \textbf{Scenario 1: Double-Precision PINNs for the Burgers Equation.} The relative $L_2$ error, training time, and number of training parameters are evaluated for solving the viscous Burgers equation with various optimizers and combinations of line search algorithms. The convergence criteria are based on absolute and relative tolerances in norm of the gradient of the loss function and number of iterations, and set as $[\text{ATOL}, \text{RTOL}]=\{[10^{-8}, 10^{-8}]~~||~~ (\#~~\text{of steps}=30000$).\}. This experiments shows that RAdam combined with BFGS with backtarcking linesearch provides better accuracy (hghlighted in bold fonts). The computation is performed on Nvidia-GPU Card RTX-3090 with persistence memory usage of 76 \% (Total 30 GB) and compute usage of 99\% i.e. 550.44 GFLOPs of total theoretical peak of 556 GFLOPs.}
    \label{tab:dp_optax_mp}
\end{minipage}
\end{figure}

\begin{figure}[!tbh]
  \begin{minipage}[b]{\linewidth}
    \centering
   \includegraphics[trim={0cm 0 0 0},clip, width=\textwidth]{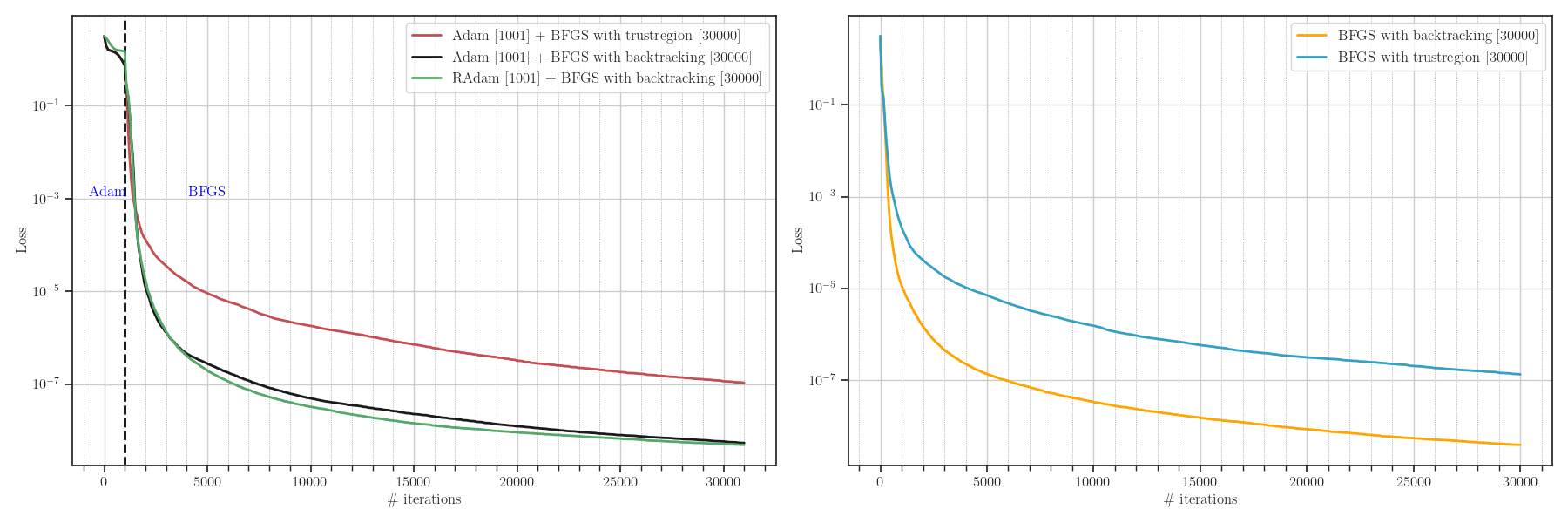}
    \captionof{figure}{\textbf{Scenario 2: Double-Precision PINNs for the Burgers Equation.} Results are shown with backtracking and trust-region strategies. The GPU is fully saturated by collocation points and data points (initial and boundary conditions) to minimize the impact of latency on computation time. The left panel illustrates a case where the iteration count for the first-order Adam and RAdam optimizers is fixed at 1001. Unlike Figure~\ref{fig:sp_optax_cp}, this combination does not exhibit degeneracy during BFGS iterations due to the use of double-precision arithmetic and achieves a significantly lower loss value. The right panel presents the convergence behavior of the standalone BFGS optimizer with trust-region and backtracking line-search methods, without warmup from a first-order optimizer. Notably, algorithms employing the backtracking line-search method demonstrate better convergence rates and higher accuracy.}
    \label{fig:dp_optax_cp}
  \end{minipage}\hfill
  \begin{minipage}[b]{\linewidth}
   \centering
\rowcolors{2}{cyan!15}{white} 
\scalebox{0.86}{
\begin{tabular}{|c|c|c|c|}
\hline
\rowcolor{cyan!40} 
\textbf{Optimizer[\# Iters.], Line-search [\#Iters.]} & \textbf{Relative $l_2$ error} & \textbf{Training time (s)} & \textbf{Total params} \\ 
\hline
Adam [1001] + BFGS with trust-region [30000] &  \(4.4 \times 10^{-5}\)     & 1141 & 1341             \\ 
\hline
Adam [1001] + BFGS with backtracking [30000] &  \(8 \times 10^{-6}\)     & 1371    &  1341            \\ 
\hline
RAdam [1001] + BFGS with backtracking [30000] &  \(\mathbf{6} \times \mathbf{10^{-6}}\)     & 1070   &     1341               \\ 
\hline
BFGS with backtracking [30000] &  \(9 \times 10^{-6}\)     & 855    &  1341                \\ 
\hline
BFGS with trust-region [30000] &  \(4.9 \times 10^{-5}\)     & 1423    & 1341               \\ 
\hline
\end{tabular}}
\captionof{table}{[Corresponding to Figure~\ref{fig:dp_optax_cp}] \textbf{Scenario 2: Double-Precision PINNs for the Burgers Equation.} 
The relative $L_2$ error, training time, and number of training parameters are reported for solving the viscous Burgers equation with various optimizers and combinations of line search algorithms. The convergence criteria here is based on union of absolute and relative tolerance of norm of gradient and considered as: $[\text{ATOL}, \text{RTOL}]=\{[10^{-7}, 1\times10^{-7}]~~||~~ (\#~~\text{of steps}=30000$).\}. The results demonstrate that RAdam paired with BFGS with the backtracking line search achieves the highest accuracy (highlighted in bold) among all cases presented in Figure~\ref{fig:dp_optax_cp}. The computation is performed on Nvidia-GPU Card RTX-3090 with persistence memory usage of 75 \% (Total 24.56 GB) and compute usage of 99\% i.e. 429 GFLOPs of total theoretical peak of 433.9  GFLOPs.}
\label{tab:dp_optax_cp}
\end{minipage}
\end{figure}

Therefore, to show the reproducibility of the proposed optimization method, in this paper we develop the same code in the JAX framework with the Optax library~\cite{deepmind2020jax}. Optax is a library of gradient transformations paired with composition operators (e.g., chain) that allow implementing many standard and new optimisers (e.g., RMSProp \cite{hinton2012rmsprop}, Adam \cite{kingma2014adam}, Lion \cite{chen2023lion}, RAdam \cite{liu2019variance}, etc.) in just a single line of code. The compositional nature of Optax naturally supports recombining the same basic ingredients in custom optimisers.  As we elaborated earlier, BFGS paired with a line search algorithm consists of the following components:

\begin{enumerate}
\item Relative tolerance: atol
\item Absolute tolerance: rtol
\item norm: $L_1,~L_2,~L_{\infty}$ norm of gradient for convergence criteria.
\item Line search direction: Backtracking, Trust-region
\item Newton descent: Equation~\eqref{eq:Bkp1BFGS}
\end{enumerate}

To customize an instance of the BFGS optimizer, the choices for items (1)–(5) should be made based on the specific problem being addressed. For this purpose, we employed the state-based Optimistix library~\cite{optimistix2024}, which integrates seamlessly with the Optax library, to implement various variants of the BFGS optimizer by inheriting the \texttt{AbstractBFGS} class. These variants are tailored according to the selected line search algorithms, tolerance levels, convergence criteria, and types of Newton descent methods. For example, the BFGS optimizer with a trust-region line search algorithm and \( L_{\infty} \)-norm convergence is shown in code listing Figure~\ref{code:BFGS line search}. To evaluate the efficiency of the optimizers in terms of both accuracy and runtime on a GPU, we conducted computational experiments under two distinct settings. In \textbf{Scenario 1}, GPU saturation is achieved by increasing the number of model parameters to a sufficiently large value, thereby minimizing the impact of latency on compute time. In \textbf{Scenario 2}, saturation is obtained by increasing the number of collocation points while keeping the model size fixed.

In \textbf{Scenario 1}, the convergence history for double-precision arithmetic is presented in Figure \ref{fig:dp_optax_mp}. This convergence is achieved using a neural network with 8 hidden layers, each containing 20 neurons, a \(\tanh\) activation function, 200 randomly sampled data points for the initial and boundary conditions, and 10,000 collocation points to calculate the residual loss. The performance of five different optimizer combinations is summarized in Table \ref{tab:dp_optax_mp}. It is worth noticing that reslts with the same network strukture for single persison is reported in~\ref{Burgers_Equation_Single_Precision}.
The convergence criteria are based on absolute and relative tolerances in norm of the gradient of the loss function and number of iterations, and set as $[\text{ATOL}, \text{RTOL}]=\{[10^{-8}, 10^{-8}]~~||~~ (\#~~\text{of steps}=30000$)\}. The convergence plot with Adam having 16001 iterations and then using BFGS paired with trust-region linesearch algorithm is shown in left subfigure of Figure \ref{fig:dp_optax_mp}. Unlike single precision, the BFGS method achieves better accuracy in this case, converging to an \(L_2\) error of \(2.8 \times 10^{-5}\). Convergence results for other scenarios, similar to those in Figure~\ref{fig:sp_optax_mp}, are displayed in the right subfigure of Figure \ref{fig:dp_optax_mp}. Performance metrics for these runs are provided in Table~\ref{tab:dp_optax_mp}. 
Notably, the degeneration of the loss function is not observed and no longer impacts the convergence of the BFGS optimizer for any of the runs. Furthermore, relative $L_2$ errors for all the cases are reduced by two orders of magnitude compared to single-precision performance. For the remaining cases, the relative \(L_2\) error is reduced to a minimum of \(2 \times 10^{-6}\). However, runs in Table~\ref{tab:dp_optax_mp} with 30,000 BFGS iterations indicate that the optimizer failed to converge, likely due to the increased sensitivity of the tolerance limit in double precision. 

In \textbf{Scenario 2}, where the GPU is fully utilized with collocation points and the PINN employs a smaller neural network, the convergence history for double precision arithmetic is shown in Figure~\ref{fig:dp_optax_cp}. Convergence was achieved using a neural network with four hidden layers, each containing 20 neurons, a \(\tanh\) activation function, 250 randomly sampled data points for initial and boundary conditions, and 50,000 collocation points. The performance metrics, including the relative \(L_2\) error and training time are summarized in Table~\ref{tab:dp_optax_cp}. The interpretation of these results is similar to that of Scenario 1, though the overall accuracy is slightly reduced. However, in double precision, the BFGS optimizer failed to converge within 30,000 iterations, likely due to the limited representational capacity of the neural network.

\subsection{Takeaways from~\ref{subsec:burgereq}}

In Table~\ref{tab:optimizer_comparison}, we provide our assessment of the most effective optimizer for PINN representation with double-precision arithmetic. The evaluation compares Wolfe and backtracking line searches alongside the trust-region approach, implemented in both TensorFlow and the JAX ecosystem. \textbf{Case 1\textsuperscript{*}} corresponds to SSBroyden with Wolfe line search, implemented in TensorFlow using double precision, while \textbf{Case 2\textsuperscript{*}} refers to BFGS with backtracking line search, implemented in JAX.

\begin{figure}
    \centering
    \rowcolors{2}{cyan!15}{white} 
    \scalebox{0.75}{
    \begin{tabular}{|c|c|c|c|c|}
        \hline
        \rowcolor{cyan!40} 
        \textbf{Case}  & \textbf{Optimizer}  &   \textbf{Training time (s)}  & \textbf{Relative $l_2$ error} & \textbf{Total parameters, Verdict}
        \\ \hline
        1\textsuperscript{*}  & Adam [1000] + SSBroyden with Wolfe [30000] & 2812  & \( 1.62 \times 10^{-8} \)  & 3,021, \textbf{Winner} 
        \\ \hline
        2\textsuperscript{*}  &  RAdam [11] + BFGS with backtracking [30000]  & 1587    & \( 2 \times 10^{-6} \)  & 3,501, \textbf{Runner-up}\\ \hline
       
    \end{tabular}
    }
    \captionof{table}{\textbf{Verdict on the choice of optimizer in double precision}: The winner and runner-up are determined based on the best relative $L_2$ error and training time for different optimizers applied to Burgers' equation. \textbf{Case 1\textsuperscript{*}} represents SSBroyden with Wolfe line search, implemented in TensorFlow using double precision, while \textbf{Case 2\textsuperscript{*}} corresponds to BFGS with backtracking line search, implemented in JAX.}
    \label{tab:optimizer_comparison}
\end{figure}

\subsection{Allen-Cahn equation} \label{sec:allen-cahn}
Next, we consider the Allen-Cahn equation, given by
\[
\frac{\partial u}{\partial t} - \epsilon \frac{\partial^2 u}{\partial x^2} + \kappa \left(u^3 - u \right) = 0,
\]

with the initial condition \( u(x, 0) = x^2 \sin \left(2 \pi x \right) \) and periodic boundary conditions:

\[
u(t, -1) = u(t, 1), \quad u_x(t, -1) = u_x(t, 1),
\]

where \( \epsilon = 10^{-4} \), \( \kappa = 5 \),
and the spatio-temporoal domain for computing the solution is \([x,~t] \in [-1, 1] \times [0, 1]\). The periodicity at the boundaries is enforced using interpolation polynomials \( \{ v^{(i)}(x) \}_{i=1}^n \) (see~\cite{DongNi2021Periodic}) and defined as

\begin{equation}\label{eq:polys}
    v^{(i)}(x) = s_0^{(i)} + s_1^{(i)}(x-a)(b-x)(a+b-2x) + \left(r_0^{(i)} + r_1^{(i)}x \right)(x-a)^2(x-b)^2,
\end{equation}

where \( a \) and \( b \)  represent the spatial domain boundaries, and the coefficients \( \{ s_0^{(i)}, s_1^{(i)}, r_0^{(i)}, r_1^{(i)} \} \) are defined for \( i = 1, 2 \) corresponding to \( n = 2 \). For all case studies involving the Allen-Cahn equation, the PINNs architecture utilizes the hyperbolic tangent activation function (\(\tanh\)) in its hidden layers. The training process begins with the Adam optimizer at \(\kappa = 1\), incorporating a learning rate decay schedule. Following this initial phase, \(\kappa\) is increased to \(5\), and the model is further refined using the BFGS and SSBroyden algorithms. This training strategy starts with Adam on a simplified problem by reducing the parameter \(\kappa\), and subsequently resumes with BFGS or SSBroyden at the original \(\kappa\) value until convergence. We found this strategy to be highly robust in achieving convergence to the global minimum. In contrast, maintaining $\kappa = 5$ throughout training occasionally led to stagnation in the loss function at relatively higher values for certain initializations, suggesting that the PINN converged to a stationary point distinct from the global minimum.

\begin{figure}[!tbh]
    \centering
        \begin{subfigure}[b]{0.99\linewidth}
        \centering
        \includegraphics[width=\linewidth]{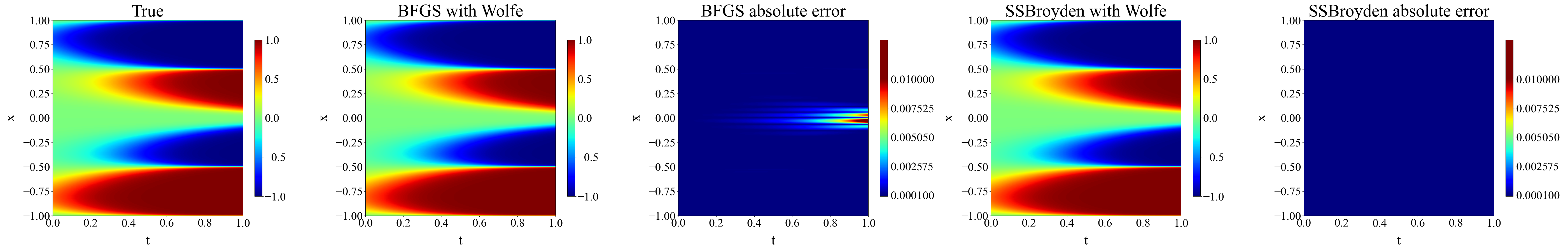}
        \label{fig:Allen-Cahn}
    \end{subfigure}
\caption{\textbf{Double-Precision PINNs for Allen-Cahn equation}: Comparison of Allen-Cahn equation predictions using PINNs optimized with BFGS and SSBroyden methods for \textbf{Case 3}. Absolute error plots are provided for each case, emphasizing the differences in error magnitudes between BFGS and SSBroyden.}
\label{fig:Allen_predict}
\end{figure}

\begin{figure}[!tbh]
    \centering
    \begin{minipage}[b]{\linewidth}
        \centering
        \includegraphics[width=\linewidth]{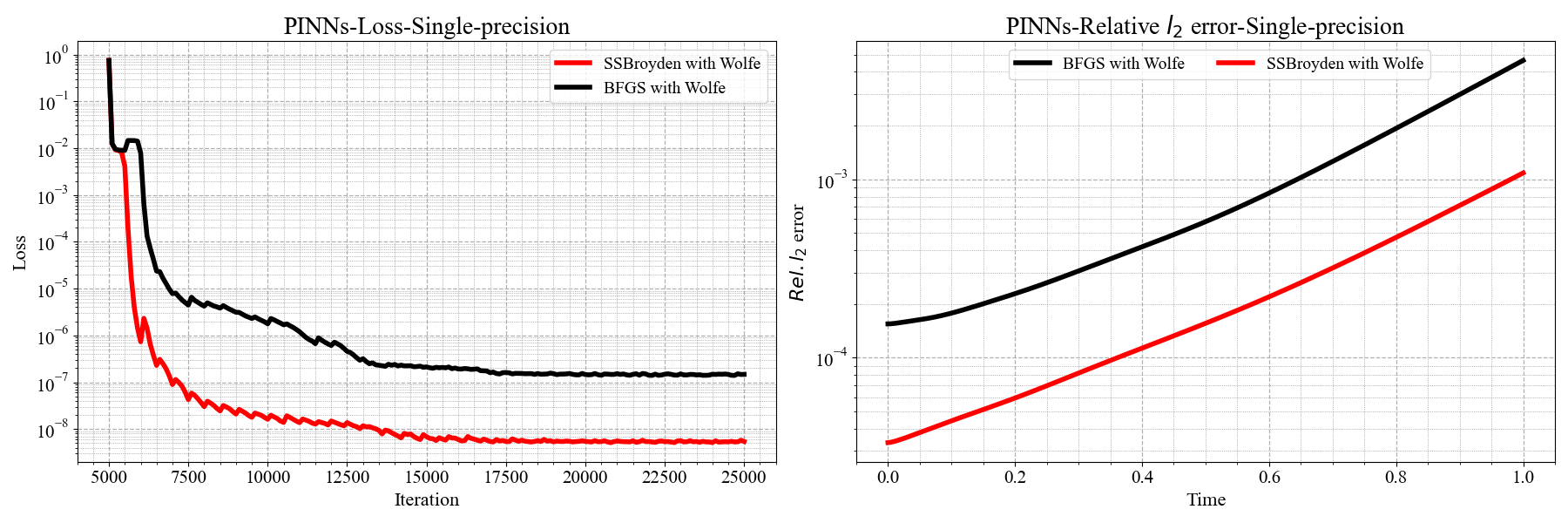}
        \caption{\textbf{Single-Precision PINNs for the Allen-Cahn equation}: Evolution of the loss function (left) and \( l_2 \) relative errors (right) for \textbf{Case 1}. The results compare the performance of PINNs trained with SSBroyden and BFGS optimizers with single precision.}
        \label{fig:PINNs_allen_Single_Allen-Cahn}
    \end{minipage}
    \vspace{1em}
    \begin{minipage}[b]{\linewidth}
        \centering
        \rowcolors{2}{cyan!15}{white}
        \scalebox{0.82}{
        \begin{tabular}{|l|r|r|r|r|}
        \hline
        \rowcolor{cyan!40} 
        \textbf{Case} &  \textbf{Optimizer[\# Iters.], Line-search [\#Iters.]} & \textbf{Relative $l_2$ error} & \textbf{Training Time (s)} & \textbf{Total Params} \\ 
        \hline
        1 &  Adam [5000] + BFGS with Wolfe [20000]                &  \(1.97 \times 10^{-3} \)   & 908  & 2,019              \\ 
        \hline
        1 &  Adam [5000] + SSBroyden with Wolfe [20000]           &   \(4.73e \times 10^{-4} \)    & 762   & 2,019     \\ 
        \hline
        \end{tabular}}
    \captionof{table}{\textbf{PINNs with single precision for the Allen-Cahn equation}: Comparison of the loss function (left) and \( l_2 \) relative errors (right) for Allen-Cahn equation using SSBroyden and BFGS optimizers. 
    }
        \label{tab:PINNs_allen_Sngle_Allen-Cahn}
    \end{minipage}
\end{figure}


\begin{figure}[!tbh]
    \centering
    \begin{minipage}[b]{\linewidth}
        \centering
        \includegraphics[width=\linewidth]{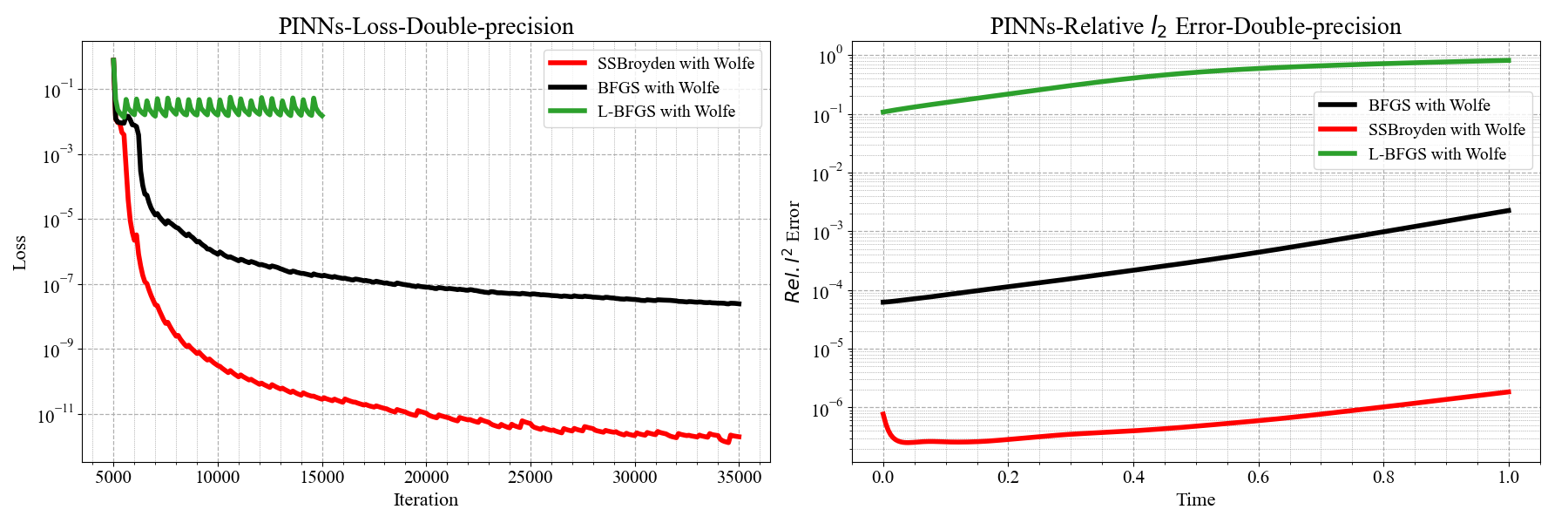}
        \caption{\textbf{Double-Precision PINNs for the Allen-Cahn equation}: Evolution of the loss function (left) and \( l_2 \) relative errors (right) for \textbf{Case 3}. The results compare the performance of PINNs trained with SSBroyden and BFGS optimizers with double precision.}
        \label{fig:PINNs_allen_Double_Precision}
    \end{minipage}
    \vspace{1em}
    \begin{minipage}[b]{\linewidth}
        \centering
        \rowcolors{2}{cyan!15}{white}
        \scalebox{0.8}{
        \begin{tabular}{|l|r|r|r|r|}
        \hline
        \rowcolor{cyan!40} 
        \textbf{Case} &  \textbf{Optimizer[\# Iters.], Line-search [\#Iters.]} & \textbf{Relative $l_2$ error} & \textbf{Training time (s)} & \textbf{Total params} \\ 
        \hline
        2 &  Adam [5000] + BFGS with Wolfe [20000]                &  \(7.59 \times 10^{-4} \)   & 733  & 2,019 \\ 
        \hline
        2 &  Adam [5000] + SSBroyden with Wolfe [20000]           &   \(1.15 \times 10^{-6} \)    & 973 & 2,019     \\ 
        \hline
        3 &  Adam [5000] + BFGS with Wolfe [30000]                &  \(9.84 \times 10^{-4} \)   & 1493 & 2,019   \\ 
        \hline
        3 &  Adam [5000] + SSBroyden with Wolfe [30000]           &   \(9.43e \times 10^{-7} \)    & 2000   & 2,019     \\ 
        \hline
        4 &   BFGS with Wolfe [30000]                &  \(3.78 \times 10^{-4} \)   & 1404  & 2,019  \\ 
        \hline
        4 &   SSBroyden with Wolfe [30000]           &   \(1.28 \times 10^{-6} \)    & 1488   & 2,019     \\ 
        \hline
        \end{tabular}}
    \captionof{table}{\textbf{Double-Precision PINNs for the Allen-Cahn equation}:       
    \textbf{Case 2:} A PINN with three layers, each containing 30 neurons, is trained for 1000 iterations using the Adam optimizer, followed by 20,000 iterations with BFGS, SSBroyden, and L-BFGS optimizers with Wolfe line-search.  
        \textbf{Case 3:} The same network structure as \textbf{Case 2} is used, but the number of iterations with BFGS and SSBroyden is reduced to 30,000. 
        \textbf{Case 4:} The same network structure as in \textbf{Case 2} and \textbf{Case 3} is used; however, instead of starting with the Adam optimizer, the training begins directly with quasi-Newton optimizers, namely BFGS and SSBroyden.}
        \label{tab:PINNs_allen_Double_Precision}
    \end{minipage}
\end{figure}


\begin{figure}[!tbh]
    \centering
    \begin{minipage}[b]{\linewidth}
        \centering
        \includegraphics[width=\linewidth]{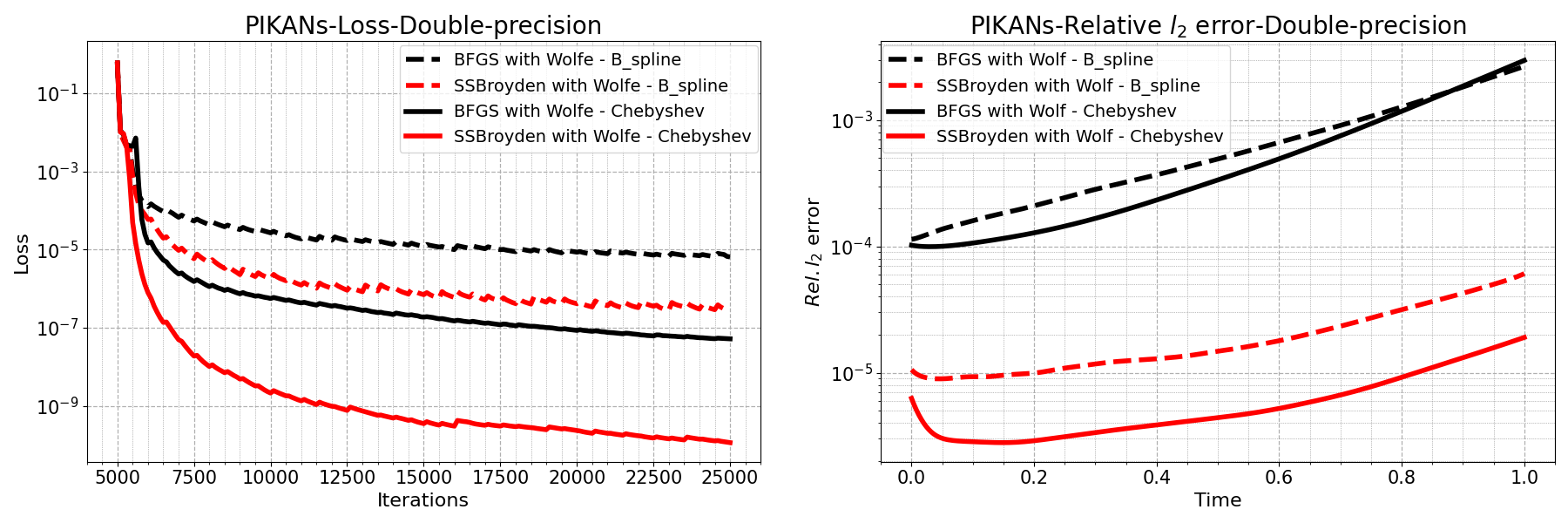}
        \caption{\textbf{Double-Precision PIKANs for the Allen-Cahn equation:} Evolution of the loss function (left) and \( L_2 \) relative errors (right) for \textbf{Case 5} (PIKAN using Chebyshev polynomials) and \textbf{Case 6} (PIKAN using B-spline polynomials). The results highlight the performance comparison of PIKANs trained using SSBroyden and BFGS optimizers.}
        \label{fig:PIKANs_Allen-Cahn}
    \end{minipage}
    \vspace{1em}
    \begin{minipage}[b]{\linewidth}
        \centering
        \rowcolors{2}{cyan!15}{white}
        \scalebox{0.75}{
        \begin{tabular}{|l|r|r|r|r|}
        \hline
        \rowcolor{cyan!40} 
        \textbf{Case} &  \textbf{Optimizer[\# Iters.], Line-search [\#Iters.]} & \textbf{Relative $l_2$ error} & \textbf{Training time (s)} & \textbf{Total parameters} \\ 
        \hline
        5  & Adam [5000] + BFGS with Wolfe-Chebyshev    & \(1.23 \times 10^{-3}\)     & 1804.82    & 1,368 \\ 
        \hline
        5  & Adam [5000] + SSBroyden with Wolfe-Chebyshev & \(9.01 \times 10^{-6}\)     & 2239.38    & 1,368 \\ 
        \hline
        6  & Adam [5000] + BFGS with Wolfe-Bspline       & \(1.17 \times 10^{-2}\)     & 11018.23   & 5,540 \\ 
        \hline
        6  & Adam [5000] + SSBroyden with Wolfe-Bspline  & \(4.52 \times 10^{-4}\)     & 12025.20   & 5,540 \\ 
        \hline
        \end{tabular}}
        \captionof{table}{\textbf{Double-Precision PIKANs for the Allen-Cahn equation:} 
        \textbf{Case 5:} PIKANs with Chebyshev polynomials consisting of four hidden layers, each with 10 neurons, and polynomial degree 5. Training involves 5000 Adam steps followed by optimization using BFGS and SSBroyden. 
        \textbf{Case 6:} PIKANs with B-splines consisting of four hidden layers with dimensions of 20, grid sizes of 10, and spline orders of 3. Training involves 5000 Adam steps followed by optimization using BFGS and SSBroyden. 
        In all cases, SSBroyden consistently outperforms BFGS in terms of accuracy.}
        \label{tab:PIKANs_Allen-Cahn_equation}
    \end{minipage}
\end{figure}

Figure \ref{fig:Allen_predict} presents contour plots comparing the  PINNs solutions against highly acurated specral solutions. For each case, the corresponding absolute errors for BFGS and SSBroyden are plotted as well. The numerical solution is computed using MATLAB's Chebfun package~\cite{Battles2004Chebfun} with 1000 Fourier modes per spatial dimension, combined with the ETDRK4 algorithm~\cite{KassamEDTRK4} for temporal integration and a time step of \(dt = 10^{-5}\).
Figure~\ref{fig:PINNs_allen_Single_Allen-Cahn} illustrates the progression of the loss function over iterations for the BFGS and SSBroyden optimizers applied to PINNs with single precision. As summarized in Table~\ref{tab:PINNs_allen_Sngle_Allen-Cahn}, SSBroyden achieves a relative error of approximately \(10^{-4}\), while BFGS converges to around \(10^{-3}\). Notably, the loss function shows minimal change beyond 15,000 epochs.

Table~\ref{tab:PINNs_allen_Double_Precision} presents three cases, all of which share the same network architecture. The network consists of three hidden layers, each with 30 neurons.
For \textbf{Case 2} and \textbf{Case 3}, the training process begins with the Adam optimizer, run for 5000 epochs at \(\kappa = 1\), using a learning rate decay schedule. Following this initial phase, \(\kappa\) is increased to \(5\), and the model is further optimized using the BFGS and SSBroyden algorithms for 20,000 and 30,000 iterations, respectively.
\textbf{Case 4} shows the results for the case that the network is trained using second optimizer BFGS and SSBroyden directly without ADAM optimizer. 
Figure~\ref{fig:PINNs_allen_Double_Precision} illustrates the evolution of the \( l_2 \) relative error over time and the loss function over iterations, comparing PINNs with double precision for \textbf{Case 3}. 

We also demonstrate the performance of SSBroyden using PIKANs. The PIKANs architecture incorporates Chebyshev polynomials and B-spline basis function. The KAN architecture consists of four hidden layers, each containing 10 nodes, with cubic spline functions (order 3) defined over 10 grid points. Also, for Chebyshev polynomials, degree 5 is used. Figure~\ref{fig:PIKANs_Allen-Cahn} illustrates the evolution of the \( l_2 \) relative error over time and the loss function over iterations, comparing PIKANs with double precision for \textbf{Case 5} and \textbf{Case 6}.
Table~\ref{tab:PIKANs_Allen-Cahn_equation} summarizes the \( l_2 \) relative error for BFGS and SSBroyden optimizers applied to PINNs and PIKANs architectures with B-spline and Chebyshev polynomials. As shown, the \( l_2 \) relative error obtained using SSBroyden for both KAN architectures is consistently lower than that obtained with BFGS. The results also confirm that PIKANs architectures with Chebyshev polynomials outperform those with B-splines in terms of both error and training time.

Overall, the results for the Allen-Cahn equation demonstrate that both PINNs and PIKANs with Chebyshev polynomials achieved a relative error of \(10^{-6}\). However, the training time to reach this accuracy is significantly shorter for PINNs (973 s) compared to PIKANs (1804 s). The \( l_2 \) relative error is computed across all time and domain points.
In terms of training time, BFGS is generally faster than SSBroyden for all cases (PINNs and both PIKANs architectures), though the difference is minimal. Notably, there is a substantial difference in training times between PINNs and PIKANs with B-splines. Additionally, Chebyshev polynomials are shown to be considerably more efficient than B-splines in both accuracy and training time.

\subsection{Kuramoto-Sivashinsky equation}\label{subsec:Kuramoto-Sivashinsky}

In this example, we illustrate the effectiveness of the SSBroyden method for tackling spatio-temporal chaotic systems, with a particular focus on the one-dimensional Kuramoto–Sivashinsky equation. Known for its intricate spatial patterns and unpredictable temporal dynamics, the Kuramoto–Sivashinsky equation poses a significant challenge to conventional numerical approaches. The governing equation is given by:

\begin{equation}\label{Eq:Ks_eq}
    \frac{\partial u}{\partial t} + \alpha u \frac{\partial u}{\partial x} + \beta \frac{\partial^2 u}{\partial x^2} + \gamma \frac{\partial^4 u}{\partial x^4} = 0,
\end{equation}

where \( \alpha = \frac{100}{16} \), \( \beta = \frac{100}{16^2} \), and \( \gamma = \frac{100}{16^4} \). 
The initial condition is defined as
\begin{center}
    $u_0(x) = \cos(x) \left( 1 + \sin(x) \right),$
\end{center}

as presented in \cite{Wang2024Causality}. The solution domain is defined as \( (t, x) \in [0, 1] \times [ x_0 =0, x_f=2\pi] \). To evaluate the performance of BFGS and SSBroyden in solving the Kuramoto-Sivashinsky equation and addressing the complexity of this spatio-temporal problem, a time-marching strategy, which divides the temporal domain into smaller intervals, enables stable training and accurate predictions over time. However, it introduces computational overhead due to sequential training for each time window. The time domain is divided into subdomains with a time increment of \( dt = 0.05 \), resulting in 20 time windows to train the equation over the interval \([0, 1]\).  Moreover, we discuss the use of PINNs for 5 time windows over the interval \([0, 0.5]\) with \( dt = 0.1 \), as described in~\cite{wang2022respecting}. 
Periodic boundary conditions are implemented following the approach in~\cite{DongNi2021Periodic}, while the initial condition is softly enforced. The periodic nature of the problem is seamlessly encoded into the model by leveraging Fourier basis functions for the spatial domain. Therefore, the inputs to the PINN model are extended to incorporate these periodic boundary conditions. The total input to the neural network, \( X_{\text{input}} \), is defined as:

\[
X_{\text{input}} = \left[t, \cos\left(\frac{2\pi m x}{L_x}\right), \sin\left(\frac{2\pi m x}{L_x}\right)\right],
\]

where \( L_x = x_f - x_0 \) is the length of the spatial domain, and \( M \) is the number of Fourier modes. Here, \( M = 10 \) is used.

In all case studies for the Kuramoto-Sivashinsky equation, the neural network architecture comprises five fully connected hidden layers, each containing 30 neurons, with the hyperbolic tangent (\(\tanh\)) activation function applied across all hidden layers. To improve computational efficiency, the collocation points are dynamically resampled using the RAD method, with updates performed every 500 iterations.
We explored two training scenarios. In the first scenario, training begins with the Adam optimizer. A learning rate decay schedule is implemented, starting at \(5 \times 10^{-3}\) and reducing by a factor of 0.98 every 1000 iterations. In the second scenario, training is initiated directly using a second optimizer, either BFGS or SSBroyden.

\begin{figure}[!tbh]
    \centering
    \includegraphics[width=0.999\linewidth]{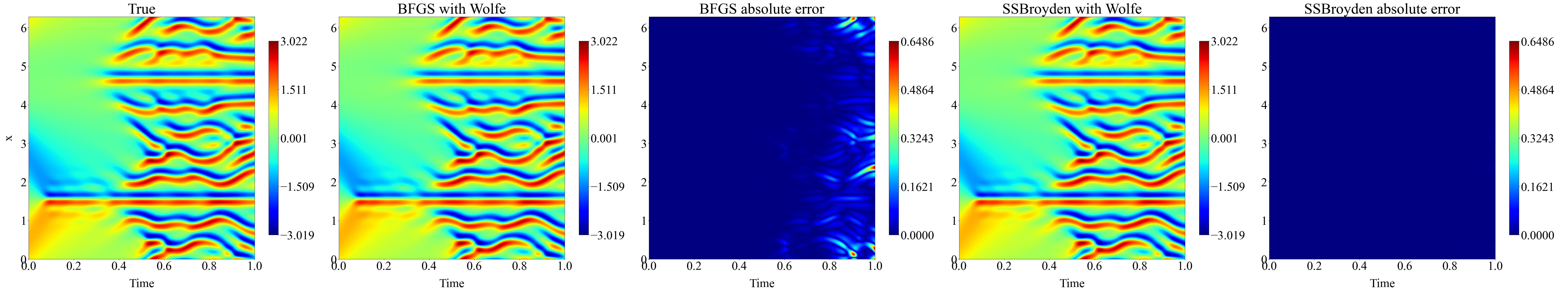}
    \caption{\textbf{Double-Precision PINNs for the Kuramoto-Sivashinsky equation:} Performance comparison of BFGS and SSBroyden in solution prediction for \textbf{Case 2}. Contour plots illustrate the absolute error for each method, revealing a notable difference in error magnitude between the two. The \(x\)-axis represents time, while the \(y\)-axis corresponds to the spatial domain.}    
    \label{fig:KS_solution}
\end{figure}

\begin{figure}[!tbh]
    \centering
    \includegraphics[width=0.995\linewidth]{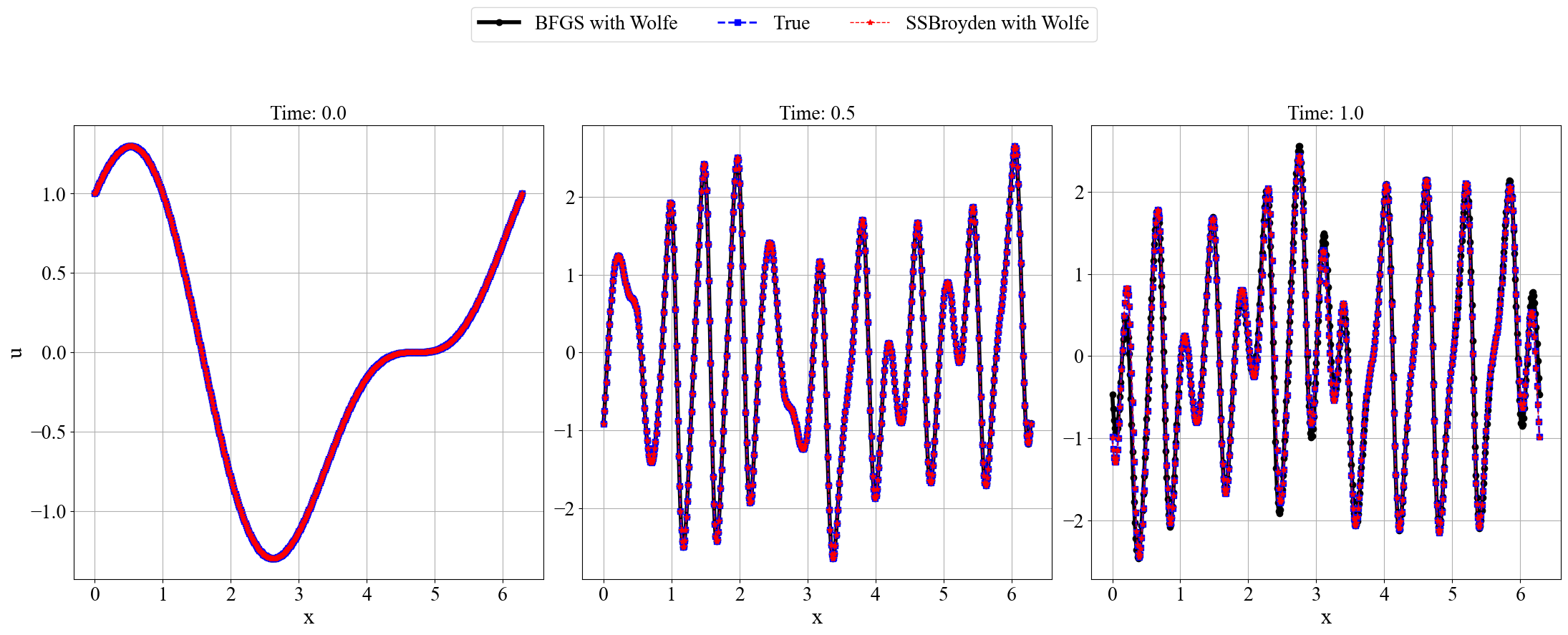}
    \caption{\textbf{Double-Precision PINNs for Kuramoto-Sivashinsky equation:} Comparison of numerical solution of the Kuramoto-Sivashinsky equation and predicted solution using BFGS and SSBroyden at three time steps $t=0.0$, $t=0.5$, and $t=1.0$ for \textbf{Case 2}. In each plot, the \(x\)-axis represents the spatial domain, and the \(y\)-axis represents the solution of the Equation~\eqref{Eq:Ks_eq}.}
    \label{fig:KS_final_time}
\end{figure}

Figures~\ref{fig:KS_solution} and~\ref{fig:KS_final_time} present the results for the Kuramoto-Sivashinsky equation for \textbf{Case 2}. Figure~\ref{fig:KS_solution} compares the performance of BFGS and SSBroyden in predicting the Kuramoto-Sivashinsky equation using double-precision PINNs with 20 time windows, displayed through contour plots. The absolute error for each prediction is shown, highlighting a significant difference in error magnitude between the two optimizers, particularly around \(t = 1.0\), where BFGS struggles to accurately predict the solution.
Figure~\ref{fig:KS_final_time} illustrates the performance of BFGS and SSBroyden at three time steps: \(t = 0.0\), \(t = 0.5\), and \(t = 1.0\) across the spatial domain for \textbf{Case 2}. The figure demonstrates the complexity of the Kuramoto-Sivashinsky equation solution as it evolves over time. Although predicting the solution around \(t = 1.0\) is challenging, the network successfully predicts the solution, with the prediction and the true solution matching closely.

\begin{figure}[!tbh]
    \centering
    \includegraphics[width=0.995\linewidth]{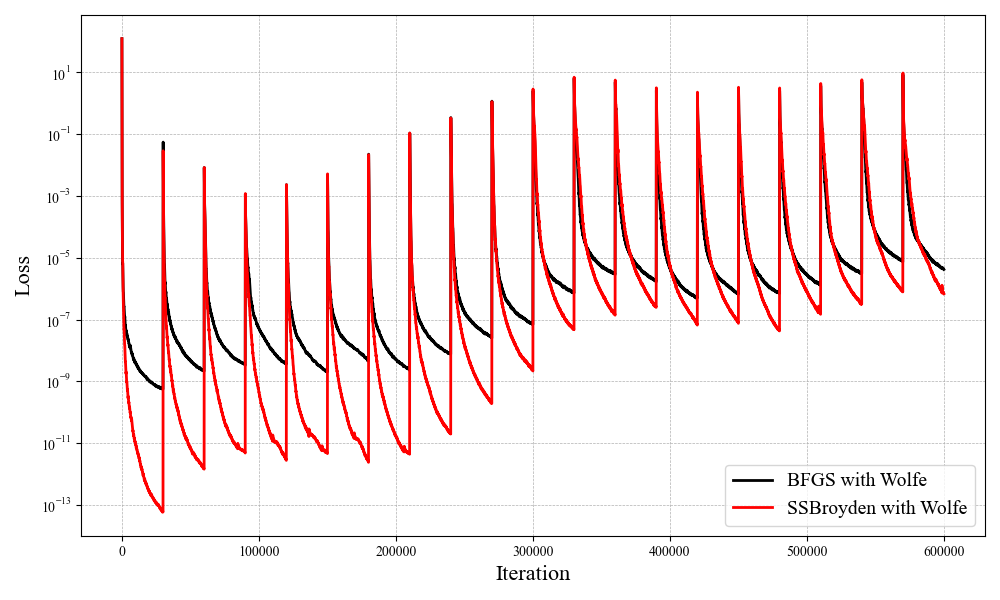}
    \caption{\textbf{Double-Precision PINNs for Kuramoto-Sivashinsky equation:}: Comparison of loss functions for BFGS and SSBroyden over 20 time windows within the time interval \([0, 1]\) with \(\Delta t = 0.05\), consisting of 20 separate PINNs (\textbf{Case 2}). Each PINNs was trained for 30,000 iterations, starting with the second optimizer BFGS or SSBroyden.}  
    \label{fig:loss_KS}
\end{figure}

\begin{figure}[!tbh]
    \centering
    \begin{minipage}[b]{\linewidth}
        \centering     
        \includegraphics[width=\linewidth]{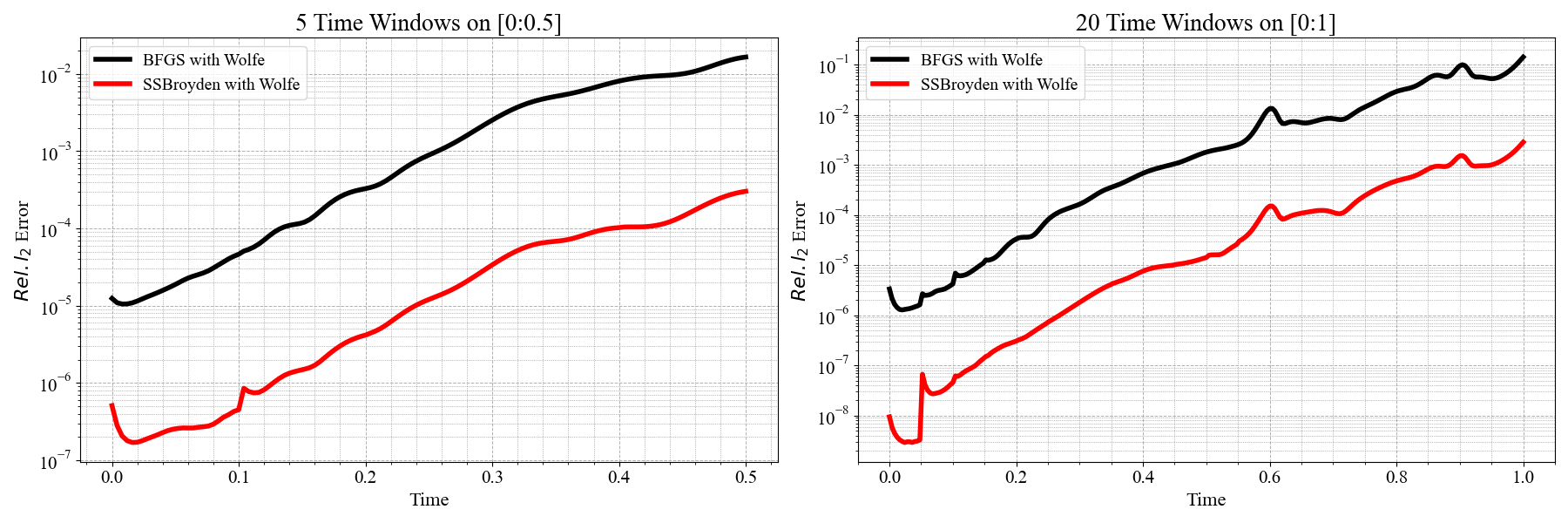}
        \caption{\textbf{Double-Precision PINNs for Kuramoto-Sivashinsky equation:}
        The evolution of \( l_2 \) relative errors is illustrated for two different time intervals and time window configurations in \textbf{Case 2}. Specifically, the plot on the left displays the \( l_2 \) relative errors over 5 time windows on the time interval \([0, 0.5]\), while the plot on the right shows the \( l_2 \) relative errors over 20 time windows on the time interval \([0, 1]\).}
        \label{fig:PINNs_equation_double_precision_KS}
    \end{minipage}
    \vspace{2em}
    \begin{minipage}[b]{\linewidth}
        \centering
        \rowcolors{2}{cyan!15}{white}
        \scalebox{0.70}{
        \begin{tabular}{|l|r|r|r|r|}
        \hline
        \rowcolor{cyan!40} 
        \textbf{Case} &  \textbf{Optimizer[\# Iters.], Line-search [\#Iters.]} & \textbf{Relative \( l_2 \) error} & \textbf{Training time (s)} & \textbf{Total parameters} \\ \hline
        1 & BFGS with Wolfe [20000] (20 windows) & \( 8.75e \times 10^{-2} \) & 102177 &  4,411\\ \hline
        1 & SSBroyden with Wolfe [20000] (20 windows) & \( 2.13 \times 10^{-3} \) & 113214 &  4,411\\ \hline
        2 & BFGS with Wolfe [30000] (20 windows) & \( 3.65 \times 10^{-2} \) & 119286 &  4,411\\ \hline
        2 & SSBroyden with Wolfe [30000] (20 windows) & \( 6.51 \times 10^{-4} \) & 130222 &  4,411\\ \hline
        3 & Adam [1000] + BFGS with Wolfe [30000] (20 windows) & \( 6.15 \times 10^{-2} \) & 119353 &  4,411\\ \hline
        3 & Adam [1000] + SSBroyden with Wolfe [30000] (20 windows) & \( 7.53 \times 10^{-4} \) & 127372 & 4,411 \\ \hline       4 &  BFGS with Wolfe [20000] (5 windows) & \( 7.54 \times 10^{-3} \) & 17793 &  4,411\\ \hline
        4 &  SSBroyden with Wolfe [20000] (5 windows) & \( 1.24 \times 10^{-4} \) & 19048 & 4,411 \\ \hline  
        5 &  BFGS with Wolfe [30000] (5 windows) & \( 2.39 \times 10^{-3} \) & 26697 &  4,411\\ \hline
        5 &  SSBroyden with Wolfe [30000] (5 windows) & \( 2.65 \times 10^{-5} \) & 30883 & 4,411 \\ \hline  
        \end{tabular}}
        \captionof{table}{\textbf{Double-Precision PINNs for the Kuramoto-Sivashinsky equation:} 
        Summary of results comparing relative \( l_2 \) errors and training times for PINNs trained using BFGS and SSBroyden optimizers over 20 time windows on the time interval \([0, 1]\) and 5 time windows on the time interval \([0, 0.5]\). 
        \textbf{Case 1:} The PINN architecture consists of five hidden layers with 30 neurons each. Optimization is performed directly using BFGS and SSBroyden for 30,000 iterations. 
        \textbf{Case 2:} The same architecture as \textbf{Case 1} is used. Training includes 30,000 iterations of BFGS and SSBroyden optimization. 
        \textbf{Case 3:} The architecture is consistent with \textbf{Case 1} and \textbf{Case 2}. Training consists of 1,000 steps of Adam optimization, followed by 30,000 iterations of BFGS and SSBroyden.        
        Similarly, for \textbf{Case 4} and \textbf{Case 5}, the architecture remains unchanged, i.e., it  employs five hidden layers with 30 neurons each and utilizes five time windows over the time interval \([0, 0.5]\), with training conducted for 20,000 and 30,000 iterations of BFGS and SSBroyden with Strong Wolfe line-search, respectively.      
        }
        \label{tab:PINNs_equation_double_precision_KS}
    \end{minipage}
\end{figure}

\begin{figure}[!tbh]
    \centering
    \begin{minipage}[b]{\linewidth}
        \centering     
        \includegraphics[width=\linewidth]{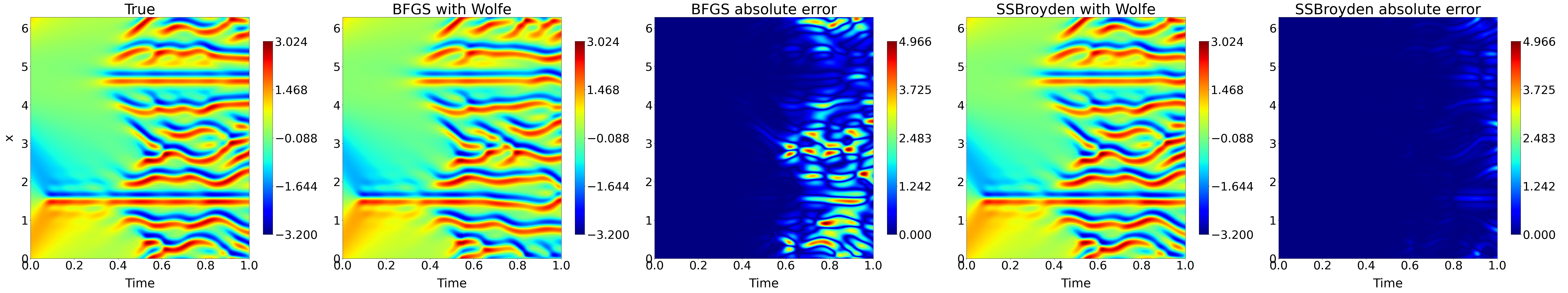}
        \caption{\textbf{Double-Precision PIKANs for Kuramoto-Sivashinsky equation:} 
        Evolution of the loss function (left) and \( l_2 \) relative errors (right) for \textbf{Case 6}. 
        The results compare the performance of PIKANs trained using SSBroyden and BFGS optimizers.}
        \label{fig:PIKANs_equation_double_precision_KS}
    \end{minipage}
    \vspace{2em}
    \begin{minipage}[b]{\linewidth}
        \centering
        \rowcolors{2}{cyan!15}{white}
        \scalebox{0.73}{
        \begin{tabular}{|l|r|r|r|r|}
        \hline
        \rowcolor{cyan!40} 
        \textbf{Case} & \textbf{Optimizer[\# Iters.], Line-search [\#Iters.]} & \textbf{Relative \( l_2 \) error} & \textbf{Training time (s)} & \textbf{Total parameters} \\ \hline
        6 &  BFGS with Wolfe [20000]-Chebyshev (degree 3) & \( 7.73 \times 10^{-1} \) & 50387 & 2,480 \\ \hline
        6 &  SSBroyden with Wolfe [20000]-Chebyshev (degree 3) & \( 1.27 \times 10^{-1} \) &   54130 & 2,480 \\ \hline
        7 & BFGS with Wolfe [20000]-Chebyshev (degree 3) & \( 1.68 \times 10^{-1} \) & 219627 &  8,160\\ \hline
        7 & SSBroyden with Wolfe [20000]-Chebyshev (degree 3) & \( 1.55 \times 10^{-1} \) & 238393 & 8,160\\ \hline 
        \end{tabular}}
        \captionof{table}{\textbf{Double-Precision PIKANs for Kuramoto-Sivashinsky equation:} 
        Summary of results highlighting the relative \( L_2 \) errors and training durations for PIKANs optimized with BFGS and SSBroyden methods.            
        \textbf{Case 6:} PIKANs with Chebyshev polynomials of degree 3 consist of five hidden layers, each containing 10 neurons. Training is initiated directly using BFGS and SSBroyden optimizers.
        \textbf{Case 7:} This PIKAN is trained similarly to \textbf{Case 6}, utilizing Chebyshev polynomials of degree 3 and consisting of five hidden layers, each with 20 neurons.}
        \label{tab:PIKANs_equation_double_precision_KS}
    \end{minipage}
\end{figure}

Figure~\ref{fig:loss_KS} illustrates the evolution of the loss function, which includes contributions from both the PDE residual and the initial conditions, across 20 time windows (\textbf{Case 2}). The time interval \([0, 1]\) is divided into 20 sequential windows, with each window trained independently using PINNs for 30,000 iterations. The comparison highlights the performance of the BFGS and SSBroyden optimization algorithms. Results show that SSBroyden consistently outperforms BFGS in all time slices, achieving faster convergence and delivering more accurate predictions across the entire domain. 
The figure also demonstrates that predicting the solution becomes increasingly challenging as we approach \(t = 1\), as evidenced by higher loss values for the final time windows compared to the initial ones.

Furthermore, Figure~\ref{fig:PINNs_equation_double_precision_KS} illustrates the relative errors obtained using BFGS and SSBroyden for the Kuramoto-Sivashinsky equation over time. The results are presented for two case studies, when the time interval \([0, 0.5]\) was divided into 5 windows (left), and when the time interval \([0, 1]\) was divided into 20 windows (right). The \(y\)-axis represents the relative error, while the \(x\)-axis represents time.
As the complexity of the Kuramoto-Sivashinsky equation solution increases over time, predictions become more challenging, particularly after \(t = 0.5\). This increased complexity leads to a noticeable rise in relative error. For the 5-window case, the relative error grows to approximately \(10^{-4}\) at \(t = 0.5\). For the 20-window case, the relative error increases to nearly \(10^{-2}\) at \(t = 1\). 

Table~\ref{tab:PINNs_equation_double_precision_KS} summarizes the relative \( l_2 \) errors and training times for the Kuramoto-Sivashinsky equation across three different cases, where PINNs are trained using double precision over 20 time windows on \([0, 1]\) and 5 time windows on the time interval \([0, 0.5]\). The architecture of the PINNs remains consistent across all cases. 
In \textbf{Case 1} and \textbf{Case 2}, the PINNs are trained directly using 20,000 and 30,000 iterations of the BFGS and SSBroyden optimizers, respectively, without prior Adam optimization. 
In \textbf{Case 3}, training involves 1,000 iterations of the Adam optimizer, followed by 30,000 iterations of BFGS and SSBroyden.
The results highlight the superior accuracy and computational efficiency of the SSBroyden optimizer compared to BFGS across all cases. Moreover, the nearly identical results for \textbf{Case 2} and \textbf{Case 3} suggest that starting optimization directly with BFGS and SSBroyden or including an initial phase of Adam optimization yields similar error levels. 
Similarly, for \textbf{Case 4} and \textbf{Case 5}, the architecture remains unchanged, employing five hidden layers with 30 neurons each and utilizing five time windows over the time interval \([0, 0.5]\). Training is conducted for 20,000 and 30,000 iterations of BFGS and SSBroyden, respectively.

Table~\ref{tab:PIKANs_equation_double_precision_KS} summarizes the performance of PIKANs using Chebyshev polynomials of degree 3 with two architectures: one consisting of five hidden layers with 10 neurons per layer (\textbf{Case 6}) and another with five hidden layers and 20 neurons per layer (\textbf{Case 7}). The table highlights the relative accuracy and efficiency of each configuration. It confirms that while increasing the number of neurons to 20 significantly raises the number of parameters and training time, it does not substantially reduce the error.
Figure~\ref{fig:PIKANs_equation_double_precision_KS} presents the results for \textbf{Case 6}. The figure illustrates that the SSBroyden optimizer successfully captures the solution, whereas the BFGS optimizer fails to achieve accurate predictions, demonstrating the superior performance of SSBroyden in this scenario. 
Comparing the prediction figures obtained using BFGS and SSBroyden with the true solution, it is evident that although the relative error reported in the table for both optimizers is approximately \(10^{-1}\), SSBroyden's predictions are significantly more accurate than those of BFGS.

\subsection{Ginzburg-Landau equation}\label{Sec:Ginzburg-Landau}
In this section, we illustrate the effectiveness of  the two-dimensional Ginzburg-Landau equation, expressed as:
\begin{equation}
    \frac{\partial A}{\partial t} = \epsilon \nabla^2 A + \left(\nu - \gamma \left| A \right|^2 \right)A,
\end{equation}
where \( A \) is a complex-valued function, and \(\epsilon\), \(\nu\), and \(\gamma\) are constant coefficients. Here, \(\epsilon\) and \(\nu\) are real numbers, while \(\gamma\) is a complex constant (\(\gamma \in \mathbb{C}\)).
Representing \( A \) in terms of its real and imaginary components, \( A = u + iv \), where \( u \) and \( v \) are real-valued functions and \( i \) is the imaginary unit, we derive the following system of PDEs:
\begin{align}
    & \frac{\partial u}{\partial t} = \epsilon \nabla^2 u + \nu u - (u^2 + v^2)\left(\Re(\gamma)u - \Im(\gamma) v \right), \\
    & \frac{\partial v}{\partial t} = \epsilon \nabla^2 v + \nu v - (u^2 + v^2)\left(\Re(\gamma)v + \Im(\gamma) u \right),
\end{align}
 where $\Re(\cdot)$ and $\Im(\cdot)$ denotes the real and the imaginary parts, respectively. Specific values for these coefficients are $\epsilon = 0.004$, $\nu = 10$, $\gamma = 10 + 15i$, which are the same values as the ones chosen in \cite{Wang2024Pirate}. 
The initial condition is 

\begin{center}
    $A_0(x,y) = 10 \left(y + ix \right) e^{-25(x^2 + y^2)}$.
\end{center}

The solution domain is defined as \( (t, x) \in [0, 1] \times [ -1,  1]^2 \). Periodic boundary conditions are enforced using a hard-enforcement method with polynomials defined in Equation~\eqref{eq:polys} for \( n = 4 \), applied to each spatial variable \( x \) and \( y \). Initial conditions, on the other hand, are applied using a soft-enforcement method. Consequently, the loss function integrates contributions from the PDE residuals and the initial conditions.
The temporal domain is divided into 5 time windows, each of length \( \Delta t = 0.2 \), with a separate PINN trained for each window. A fully connected neural network is designed to predict the solution fields \( u \) and \( v \). The network architecture consists of five dense layers, each with 30 neurons, where all hidden layers utilize the hyperbolic tangent (\(\tanh\)) activation function. The output layer contains two neurons, providing simultaneous predictions for \( u \) and \( v \).
To enhance training efficiency, the model incorporates the adaptive sampling strategy (RAD) algorithm.

\begin{figure}[!tbh]
    \centering
    \begin{subfigure}[b]{0.999\linewidth}
        \centering
        \includegraphics[width=\linewidth]{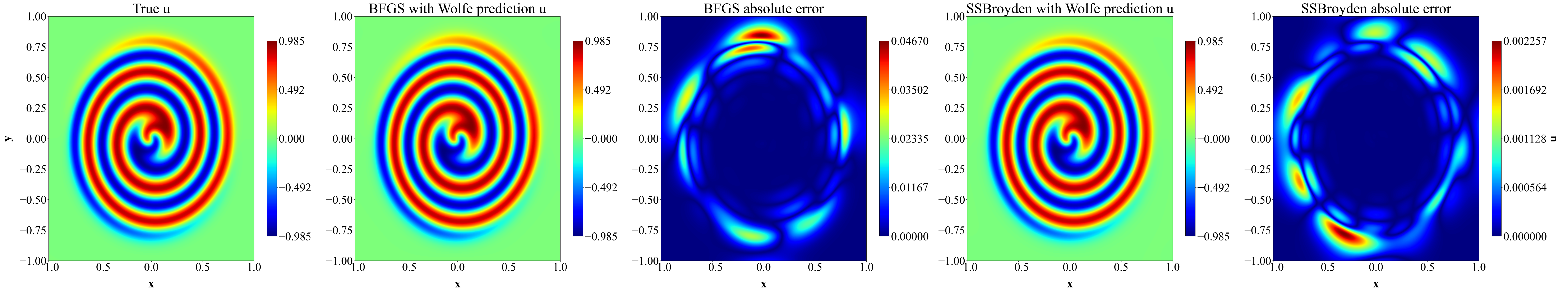}
        \caption{Real part of the Ginzburg-Landau equation}
        \label{fig:GL_solution_u}
    \end{subfigure}
    \hfill
    \begin{subfigure}[b]{0.999\linewidth}
        \centering
        \includegraphics[width=\linewidth]{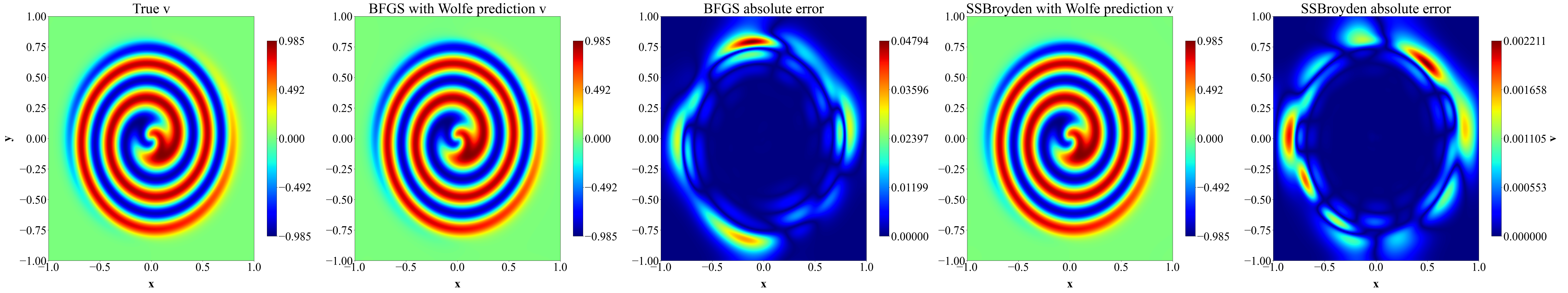}
        \caption{Imaginary part of the Ginzburg-Landau equation}
        \label{fig:GL_solution_v}
    \end{subfigure}
\caption{\textbf{Double-Precision PINNs for Ginzburg-Landau equation}: Comparison of the numerical solution for the Ginzburg-Landau equation with predictions obtained using double-precision PIKANs with 5 time windows, each of length \( \Delta t = 0.2 \). (a) Real part. (b) Imaginary part. 
In each case, the absolute error is also plotted, demonstrating the superior performance of SSBroyden compared to BFGS for both the real and imaginary components.}
    \label{fig:GL_solution}
\end{figure}

\begin{figure}[!tbh]
    \centering
    \begin{subfigure}[b]{0.999\linewidth}
        \centering
        \includegraphics[width=\linewidth]{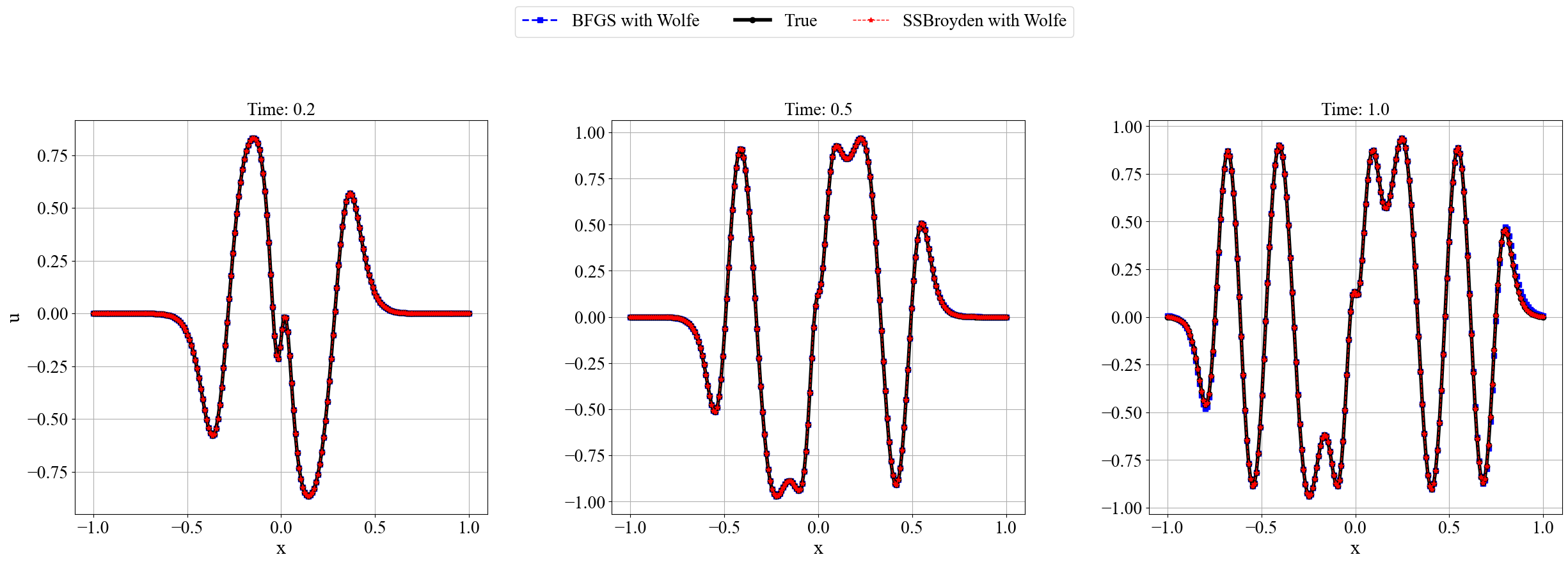}
        \caption{Real part of the Ginzburg-Landau equation}
        \label{fig:GL_solution_u2}
    \end{subfigure}
    \hfill
    \begin{subfigure}[b]{0.999\linewidth}
        \centering
        \includegraphics[width=\linewidth]{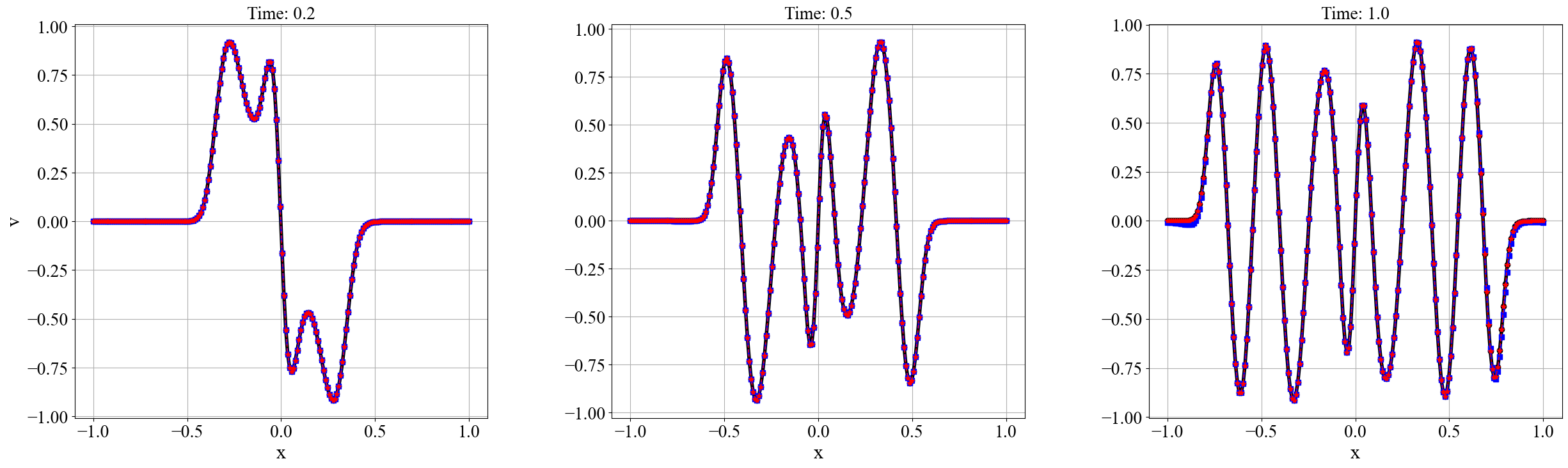}
        \caption{Imaginary part of the Ginzburg-Landau equation}
        \label{fig:GL_solution_v2}
    \end{subfigure}
    \caption{\textbf{Double-Precision PINNs for Ginzburg-Landau equation}: Comparison of the numerical solution and the predicted solution using BFGS and SSBroyden. (a) real part. (b) imaginary part. Results are shown at time steps $t=0.2$, $t=0.5$, and $t=1.0$ obtained using double-precision PIKANs with 5 time windows, each of length \( \Delta t = 0.2 \).}
    \label{fig:GL_solution_over_time}
\end{figure}

\begin{figure}[!tbh]
    \centering
    \includegraphics[width=0.7\linewidth]{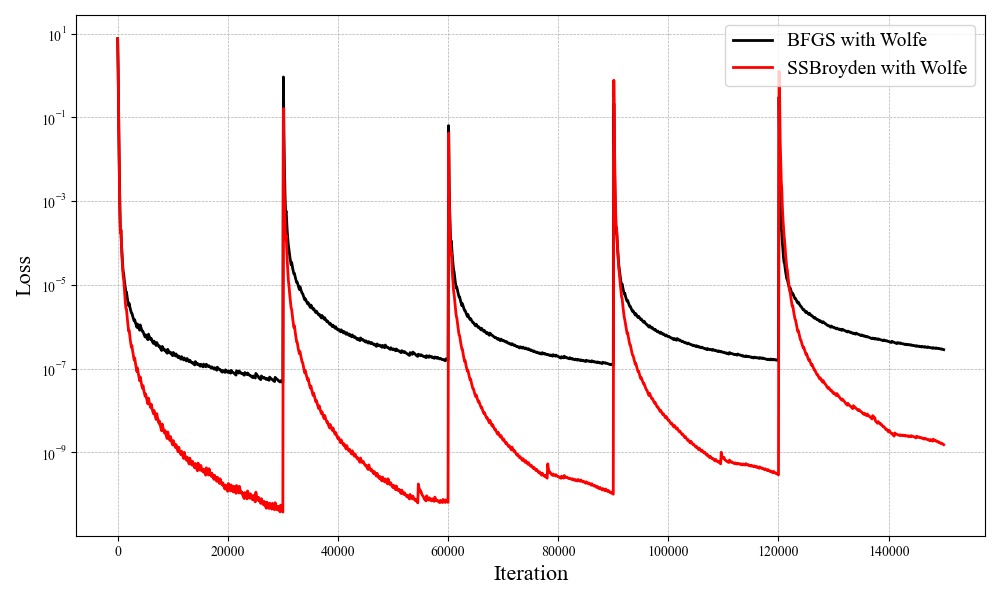}
    \caption{\textbf{Double-Precision PINNs for Ginzburg-Landau equation}: 
    Comparison of loss functions for BFGS and SSBroyden over 5 time windows within the time interval \([0, 1]\) with \(\Delta t = 0.2\), consisting of 5 separate PINNs (\textbf{Case 2}). Each PINNs was trained for 30,000 iterations, starting with the second optimizer BFGS or SSBroyden.}
    \label{fig:loss_GL_PINNs}
\end{figure}

\begin{figure}[!tbh]
    \centering
    \begin{minipage}[b]{\linewidth}
        \centering
        \includegraphics[width=\linewidth]{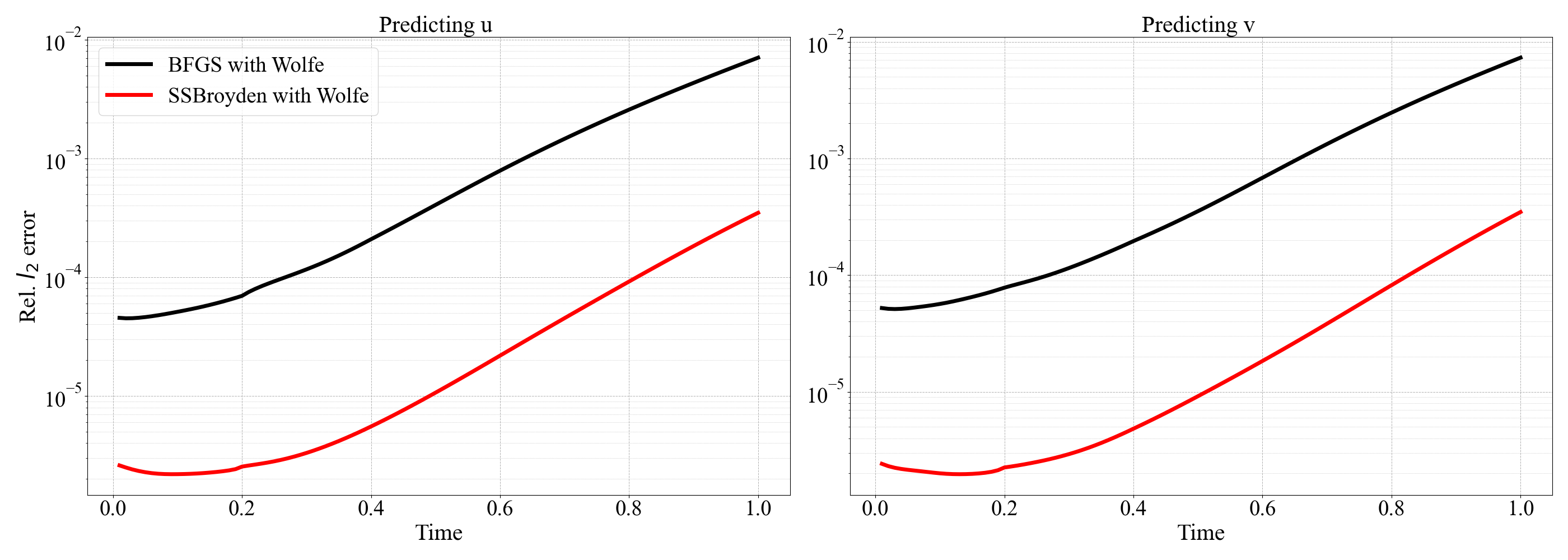}
        \caption{\textbf{Double-Precision PINNs for Ginzburg-Landau equation}: The \( l_2 \) relative error for the Ginzburg-Landau equation, computed using BFGS and SSBroyden over 5 time windows within the time interval \([0, 1]\) with \(\Delta t = 0.2\) for \textbf{Case 3}.}
        \label{fig:PINNs_gl_error}
    \end{minipage}
    \vspace{1em}
    \begin{minipage}[b]{\linewidth}
        \centering
        \rowcolors{2}{cyan!15}{white}
        \scalebox{0.74}{
        \begin{tabular}{|l|r|r|r|r|}
        \hline
        \rowcolor{cyan!40} 
        \textbf{Case} &  \textbf{Optimizer[\# Iters.], Line-search [\#Iters.]} & \textbf{Relative \( l_2 \) error} & \textbf{Training time (s)} & \textbf{Total parameters} \\ 
        \hline
        1 & BFGS with Wolfe [20000] & \(2.73 \times 10^{-2}\)     & 21265  & 3,184 \\ 
        1 & SSBroyden with Wolfe [20000] & \(7.68 \times 10^{-3}\)     & 21833  & 3,184 \\ 
        2 & BFGS with Wolfe [20000] & \(1.05 \times 10^{-2}\)     & 41159 & 6,904 \\ 
        2 & SSBroyden with Wolfe [20000] & \(7.80 \times 10^{-4}\)     & 47131    & 6,904  \\ 
        3 & BFGS with Wolfe [30000] & \(7.19 \times 10^{-3}\)     & 61178   & 6,904 \\ 
        3 & SSBroyden with Wolfe [30000] & \(3.48 \times 10^{-4}\)  & 64883 & 6,904  \\ 
        \hline
        \end{tabular}}
        \captionof{table}{\textbf{Double-Precision PINNNs for the Ginzburg-Landau equation:} 
        Comparison of relative \( l_2 \) errors, training times, and total parameters for PINNs trained using BFGS and SSBroyden optimizers. 
        \textbf{Case 1:} The network architecture consists of five dense layers, each with 30 neurons, designed to predict the solution fields \( u \) and \( v \). The temporal domain is divided into 5 time windows, each of length \( \Delta t = 0.2 \), with a separate PINN trained for each window. Training is performed using 20,000 iterations of BFGS and SSBroyden optimizers. 
        \textbf{Case 2:} An extended network architecture with eight dense layers, each containing 30 neurons, is used to predict the solution fields \( u \) and \( v \). The temporal domain is divided into 5 time windows, and the PINNs are trained using 20,000 iterations of BFGS and SSBroyden optimizers. 
        \textbf{Case 3:} The same architecture as \textbf{Case 1} is trained using 30,000 iterations of BFGS and SSBroyden optimizers. 
        The results highlight that SSBroyden consistently outperforms BFGS across all cases. It is worth noting that the error reported in this table represents the average error for the prediction of the real part \( u \) and the imaginary part \( u \).}        
        \label{tab:PINNs_gl_performance}
    \end{minipage}
\end{figure}

\begin{figure}[!tbh]
    \centering
    \begin{minipage}[b]{\linewidth}
        \centering
        \includegraphics[width=\linewidth]{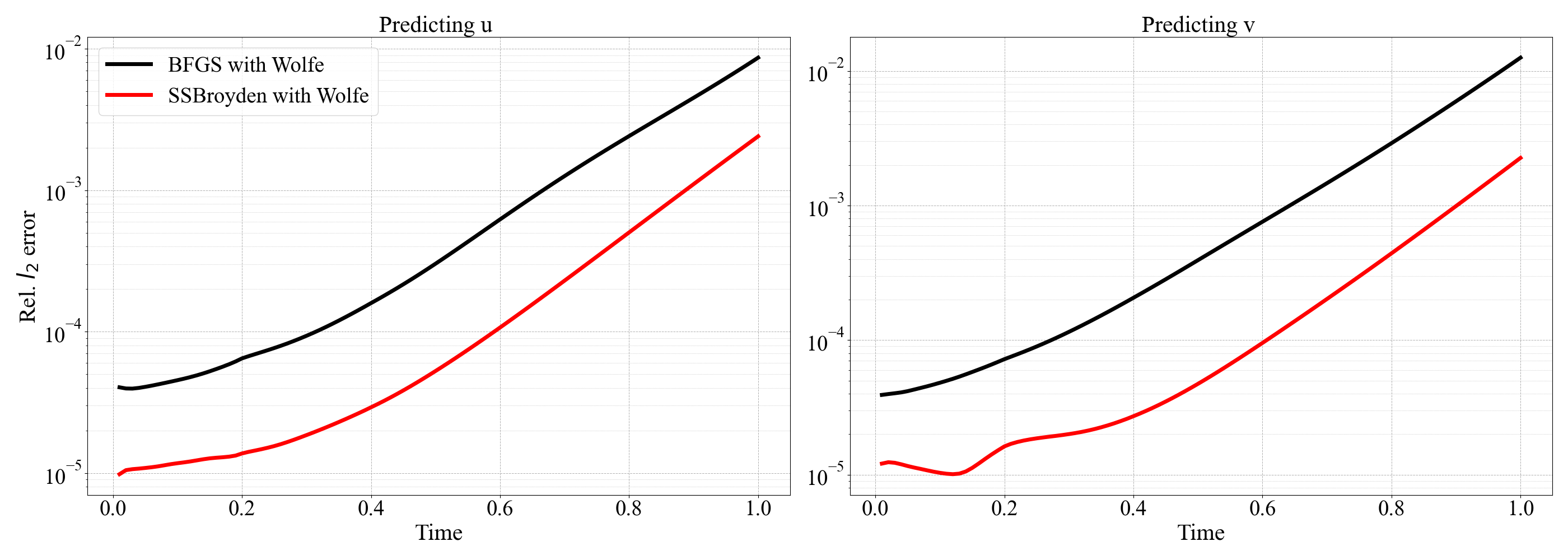}
        \caption{\textbf{Double-Precision PIKANs for the Ginzburg-Landau equation}: The \( l_2 \) relative error over time for the Ginzburg-Landau equation (\textbf{Case 4}), computed using BFGS and SSBroyden optimizers with PIKAN architectures using Chebyshev polynomials.}
        \label{fig:PIKANs_gl_error}
    \end{minipage}
    \vspace{1em}
    \begin{minipage}[b]{\linewidth}
        \centering
        \rowcolors{2}{cyan!15}{white}
        \scalebox{0.75}{
        \begin{tabular}{|l|r|r|r|r|}
        \hline
        \rowcolor{cyan!40} 
        \textbf{Case} &  \textbf{Optimizer[\# Iters.], Line-search [\#Iters.]} & \textbf{Relative \( l_2 \) error} & \textbf{Training time (s)} & \textbf{Total parameters} \\ 
        \hline
        4 & BFGS with Wolfe-Chebyshev[30000](degree 3) & \(7.88 \times 10^{-2}\)     & 14614   & 12,152 \\ 
        4 & SSBroyden with Wolfe-Chebyshev[30000](degree 3) & \(1.10 \times 10^{-3}\) & 26152  & 12,152  \\ 
        5 & BFGS with Wolfe-Chebyshev[30000](degree 5) & \(1.06 \times 10^{-2}\)     & 44151  & 18,212 \\ 
        5 & SSBroyden with Wolfe-Chebyshev[30000](degree 5) & \(2.33 \times 10^{-3}\) & 46625  & 18,212  \\ 
        \hline
        \end{tabular}}
\captionof{table}{\textbf{Double-Precision PIKANs for the Ginzburg-Landau equation}: Comparison of relative \( l_2 \) errors, training times, and total parameters for PIKANs using Chebyshev polynomials of degree 3 (\textbf{Case 4}) and degree 5 (\textbf{Case 5}). Both models employ four dense layers with 30 neurons each and are trained for 30,000 iterations using BFGS and SSBroyden optimizers. The results show that SSBroyden consistently outperforms BFGS. All values are computed across all data points and time steps, averaged over the real (\( u \)) and imaginary (\( v \)) components.}
        \label{tab:PIKANs_gl_performance}
    \end{minipage}
\end{figure}

Figures~\ref{fig:GL_solution} compare the numerical solution of the Ginzburg-Landau equation with the predicted solutions obtained using double-precision PINNs trained with BFGS and SSBroyden, presented as contour plots for \textbf{Case 3}. The absolute errors for both optimizers are shown for (a) the real part \( u \) and (b) the imaginary part \( v \), demonstrating the superior accuracy of SSBroyden over BFGS.
The numerical solution is computed using Chebfun with 200 Fourier modes in each spatial direction, combined with the EDTRK4 algorithm for time integration, employing a step size of \( dt = 10^{-5} \).
Additionally, Figures~\ref{fig:GL_solution_over_time} illustrate the performance of BFGS and SSBroyden at three distinct time steps: \( t = 0.0 \), \( t = 0.5 \), and \( t = 1.0 \), across the spatial domain for \textbf{Case 3}. The plots display (a) the real fields \( u \) and (b) the imaginary fields \( v \), highlighting the evolution and increasing complexity of the solution over time. These results further emphasize the effectiveness of SSBroyden in capturing the solution dynamics accurately.

Figure~\ref{fig:loss_GL_PINNs} presents the evolution of the loss function, which incorporates contributions from both the PDE residual and the initial conditions, across 5 time windows for \textbf{Case 3}. The comparison highlights the performance of the BFGS and SSBroyden optimization algorithms. The results indicate that SSBroyden consistently outperforms BFGS across all time slices, achieving faster convergence and providing more accurate predictions throughout the domain.

Table~\ref{tab:PINNs_gl_performance} summarizes the relative \( l_2 \) error, training time, and total number of training parameters for solving the Ginzburg-Landau equation using different optimizers. The results are computed across all data points and time steps, averaged over the real part (\( u \)) and the imaginary part (\( v \)) for three cases. 
In \textbf{Case 1}, the network architecture consists of five dense layers, each with 30 neurons, designed to predict the solution fields \( u \) and \( v \). In contrast, for \textbf{Case 2} and \textbf{Case 3}, an extended network architecture with eight dense layers, each containing 30 neurons, is utilized. \textbf{Case 2} and \textbf{Case 3} are trained using 20,000 and 30,000 iterations of the BFGS and SSBroyden optimizers, respectively.
Figure~\ref{fig:PINNs_gl_error} illustrates the evolution of the \( l_2 \) relative error over time for both the real part (\( u \)) and the imaginary part (\( v \)) for \textbf{Case 3}. While the errors increase as the prediction complexity grows over time, SSBroyden consistently achieves lower errors compared to BFGS for both components.

Table~\ref{tab:PIKANs_gl_performance} provides a summary of the relative \( l_2 \) error, training time, and total number of training parameters for solving the Ginzburg-Landau equation using PIKANs with Chebyshev polynomials of degree 3 (\textbf{Case 4}) and degree 5 (\textbf{Case 5}). The results are computed across all data points and time steps, averaged over the real part (\( u \)) and the imaginary part (\( v \)). 
In both cases, the network architecture consists of four dense layers, each with 30 neurons, and the models are trained using 30,000 iterations of the BFGS and SSBroyden optimizers. 
Figure~\ref{fig:PIKANs_gl_error} shows the evolution of the \( l_2 \) relative error over time for both the real part (\( u \)) and the imaginary part (\( v \)) for \textbf{Case 5}. Although the errors increase as the prediction complexity grows over time, SSBroyden consistently achieves lower errors compared to BFGS for both components.

\begin{figure}[!tbh]
    \centering
    \begin{minipage}[b]{\linewidth}
        \centering
        \includegraphics[width=\linewidth]{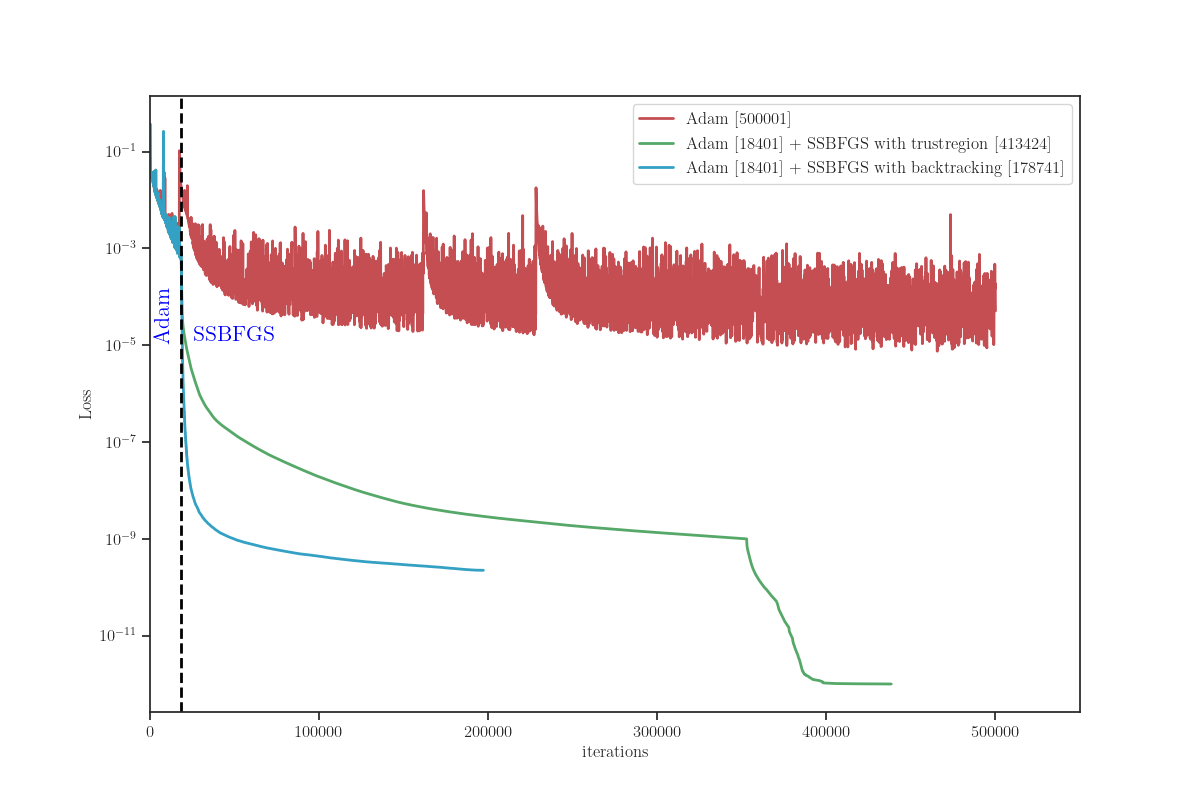}
        \caption{\textbf{Double-precision PINNs for the Stokes equation (~\autoref{eq:Stokes}).} 
        The model is trained using the Adam optimizer followed by SSBFGS with trust-region and backtracking line search strategies. All computations are performed using 64-bit floating-point arithmetic. The results show that SSBFGS combined with the trust-region method significantly outperforms other approaches in terms of convergence.}
        \label{fig:Stokes_loss}
    \end{minipage}

    \vspace{1em}

    \begin{minipage}[b]{\linewidth}
        \centering
        \rowcolors{2}{cyan!15}{white}
        \scalebox{0.78}{
        \begin{tabular}{|l|c|c|}
            \hline
            \rowcolor{cyan!40}
            \textbf{Optimizer [\# Iters.], Line Search [\# Iters.]} & \textbf{Relative \(l_2\) Error (u, v, p)} & \textbf{Total Parameters} \\
            \hline
            Adam (18400) + SSBFGS with Trust-Region (419,965)  & ($3.25 \times 10^{-4}$, $1.57 \times 10^{-4}$, $1.10 \times 10^{-1}$) & 33,667 \\
            Adam (18400) + SSBFGS with Backtracking (178,741)  & ($3.26 \times 10^{-4}$, $1.60 \times 10^{-4}$, $1.10 \times 10^{-1}$) & 33,667 \\
            Adam (500,000)                                     & ($2.70 \times 10^{-2}$, $2.58 \times 10^{-2}$, $1.10 \times 10^{-1}$) & 33,667 \\
            Lion (500,001)                                     & ($1.63$, $1.01$, $9.98$)                                               & 33,667 \\
            AdaBelief (500,001)                                & ($3.11 \times 10^{-2}$, $3.53 \times 10^{-2}$, $1.50 \times 10^{-1}$) & 33,667 \\
            Adam (18,401) + Nonlinear CG (Polak–Ribiere) (500,001)~\cite{polak1969note} & ($4.13 \times 10^{-2}$, $3.96 \times 10^{-2}$, $9.54 \times 10^{-1}$) & 33,667 \\
            \hline
        \end{tabular}}
        \captionof{table}{\textbf{Double-precision PINNs for the Stokes equation.}
        Comparison of relative \(l_2\) errors and total number of model parameters for different optimizers. All models use the same network architecture with 33,667 parameters. The absolute and relative tolerance for SSBFGS convergence is set to \(10^{-14}\). SSBFGS with trust-region line search achieves the best accuracy. For comparison, we also report the performance of Lion, AdaBelief, and Hessian-free Nonlinear Conjugate Gradient (CG) using the Polak–Ribiere variant~\cite{polak1969note}, all of which perform significantly worse.
        All computations are performed on an NVIDIA H100 GPU, with persistent memory usage of 75\% (total 80 GB) and 99\% compute utilization, achieving 25.35 TFLOPs of the theoretical peak of 25.61 TFLOPs.}
        \label{tab:Stokes_equation}
    \end{minipage}
\end{figure}

\subsection{Lid-driven wedge flow}\label{sec:Stokes equation}
To show the efficacy of proposed optimizer, we finally solve the Stokes equation for a viscous flow in a lid-driven wedge. The Stokes equation is given by
\begin{align}\label{eq:Stokes}
\begin{aligned}
\frac{\partial p}{\partial x} -\left(\frac{\partial^2 u} {\partial x^2} + \frac{\partial^2 u} {\partial y^2} \right) = 0,\\
\frac{\partial p}{\partial y} -\left(\frac{\partial^2 v} {\partial x^2} + \frac{\partial^2 v} {\partial y^2} \right) = 0,\\
\frac{\partial u}{\partial x} + \frac{\partial v} {\partial y} = 0,
\end{aligned}
\end{align}

\begin{figure}[!tbh]
\centering
\begin{subfigure}[b]{0.43\linewidth}
\includegraphics[width=\linewidth]{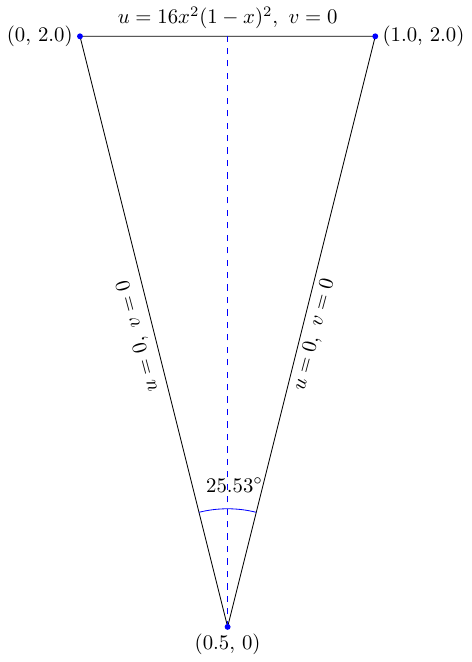}
\caption{Problem setup for \autoref{eq:Stokes}}
\end{subfigure}
\begin{subfigure}[b]{0.51\linewidth}
\includegraphics[trim={0 1cm 0 0},clip, width=\linewidth]{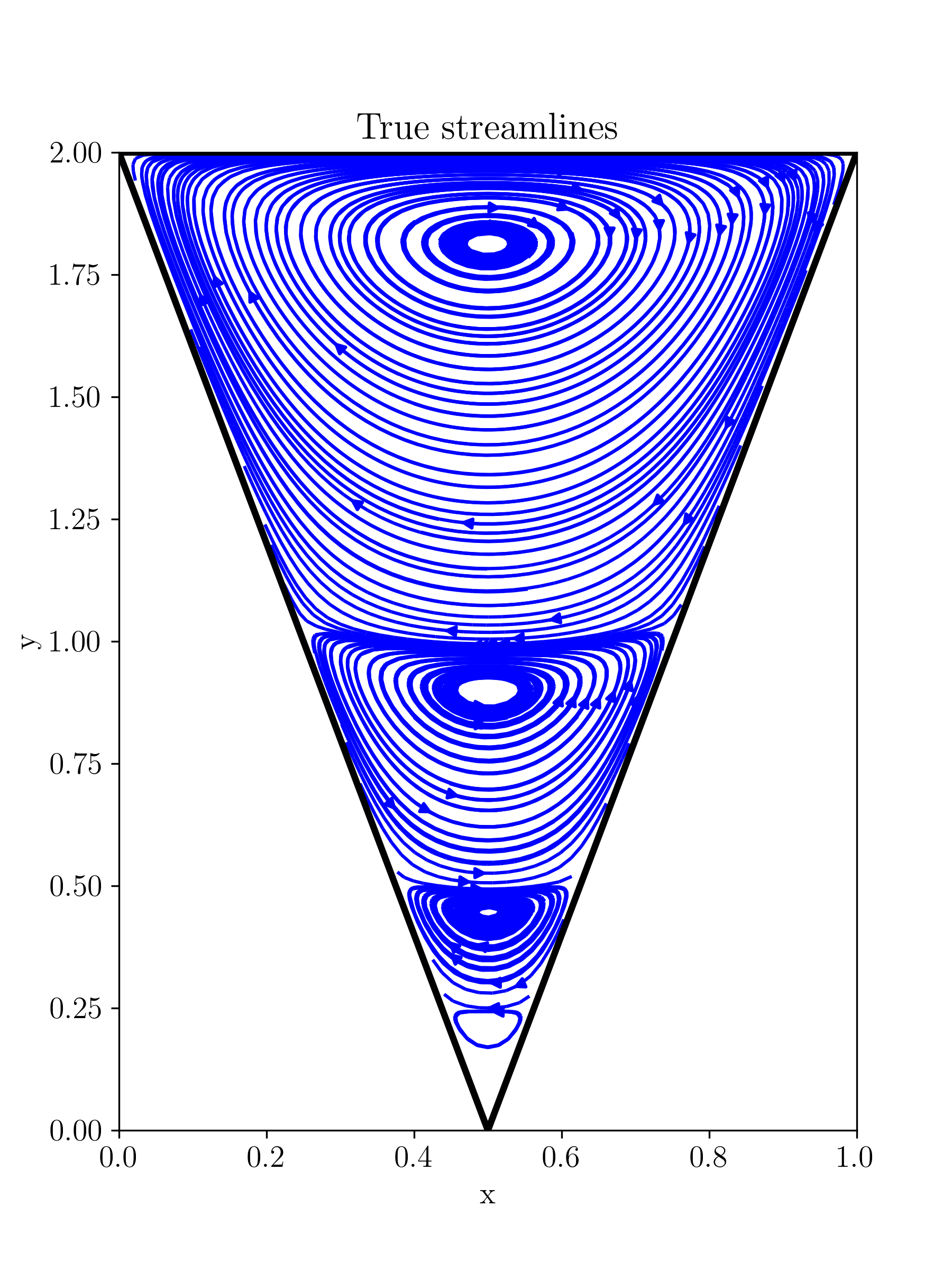}
\caption{Reference solution}
\label{fig:stokes_dom}
\end{subfigure} 
\caption{Problem setup and flow condition used  for solving \autoref{eq:Stokes}. Subfigure (a) shows a fluid domain of wedge shape with an aspect ratio of 2:1 and an angle of 25.53$^\circ$. The transeverse velocity $u$ at top lid shows reqgualrized lid and other two boundaries are rigid wall with $u=v=0$. To validate the solution obtained from PINNs, we solve the \autoref{eq:Stokes} using the spectral element method \cite{karniadakis2005spectral} and present the streamlines in (b). }
\label{fig:Stokes_Domain}
\end{figure}

The boundary conditions and domain configuration are illustrated in \autoref{fig:Stokes_Domain}. This problem is primarily considered to assess the effectiveness of the proposed optimizer for solving PDEs in scenarios involving extremely small values. Moffatt \cite{moffatt1964viscous} derived an asymptotic solution to Equation~\eqref{eq:Stokes} using a similarity approach, demonstrating that the strength of eddies depends on the wedge angle. For the specific case depicted in Figure \ref{fig:Stokes_Domain}, which features a wedge angle of 25.53$^\circ$, the strength of each successive eddy should asymptotically be about 400 times weaker than the preceding one. Accurately resolving such weak eddies requires a highly precise and robust optimizer. This problem has not yet been addressed using the PINNs presented in previous studies, primarily due to two key challenges: the limitations of first-order optimizers and the difficulty of efficiently performing double-precision computations. Even in conventional numerical method these weaker eddies are captured with high-order numerical methods \cite{karniadakis2005spectral}. In this work, we conduct computational experiments to overcome these challenges by comparing our implementation of SSBFGS in JAX—employing both backtracking and trustregion line search methods—with the commonly used Adam optimizer. Therefore, this example also answers the long standing question of PINN recovering the solution to machine precision. The results of this comparison are discussed in the following paragraphs.

In Figure~\ref{fig:Stokes_Domain}, we show the fluid domain and boundary condition on each edge of the domain with non zero transverse velocity $(u)$ at the top edge  and rigid wall (no-slip)  condition on the other two edges. To validate the solution obtained by PINNs we compute the reference solution by solving Equation~\eqref{eq:Stokes} using spectral element method as proposed in \cite{karniadakis2005spectral} and use the Nektar++ codebase \cite{cantwell2015nektar++} to perform the simulation. We show the True streamlines in Figure~\ref{fig:Stokes_Domain}b obtained from the Nektar++. To configure the PINN across all cases, we employ a neural network architecture consisting of 8 fully connected layers, each with 64 neurons, and use the $\tanh$ activation function. A total of 120,000 residual points are sampled uniformly from the domain using $\mathcal{U}[0, 1)$. To enhance convergence speed and accuracy, we adopt residual-based attention as proposed in \cite{Anagnostopoulos2024SNR}, along with the hard imposition of boundary conditions following~\cite{sukumar2022exact}. Training is conducted for three distinct cases. In the first case, the Adam optimizer is used. In the second and third cases, we employ the SSBFGS optimizer in combination with trust-region and backtracking line search strategies, respectively. For the second-order optimization in these cases, the training is initially warmed up with Adam for 18,401 iterations before switching to SSBFGS. The convergence history for all three cases is presented in Figure~\ref{fig:Stokes_loss}. As shown in Figure~\ref{fig:Stokes_loss}, the combination of SSBFGS with the trust-region line search demonstrates superior performance compared to both SSBFGS with backtracking and the Adam optimizer. Specifically, SSBFGS with trust-region converges to a loss value of $10^{-12}$, while SSBFGS with backtracking and the Adam optimizer converge to $10^{-10}$ and $10^{-5}$, respectively.  Training details for all three cases are summarized in Table \ref{tab:Stokes_equation}. To further evaluate the optimizer’s performance, we also include comparison metrics for several niche first-order optimizers—Lion~\cite{chen2023symbolic}, AdaBelief~\cite{zhuang2020adabelief}, and the non-linear conjugate gradient method with the Polak–Ribiere update rule~\cite{polak1969note}—using the same time budget, also presented in Table \ref{tab:Stokes_equation}. Notably, the SSBFGS optimizer with a trust-region approach outperforms all other methods.

\begin{figure}[!tbh]
\begin{subfigure}[b]{0.5\linewidth}
\includegraphics[trim={0 1cm 0 0},clip, width=\linewidth]{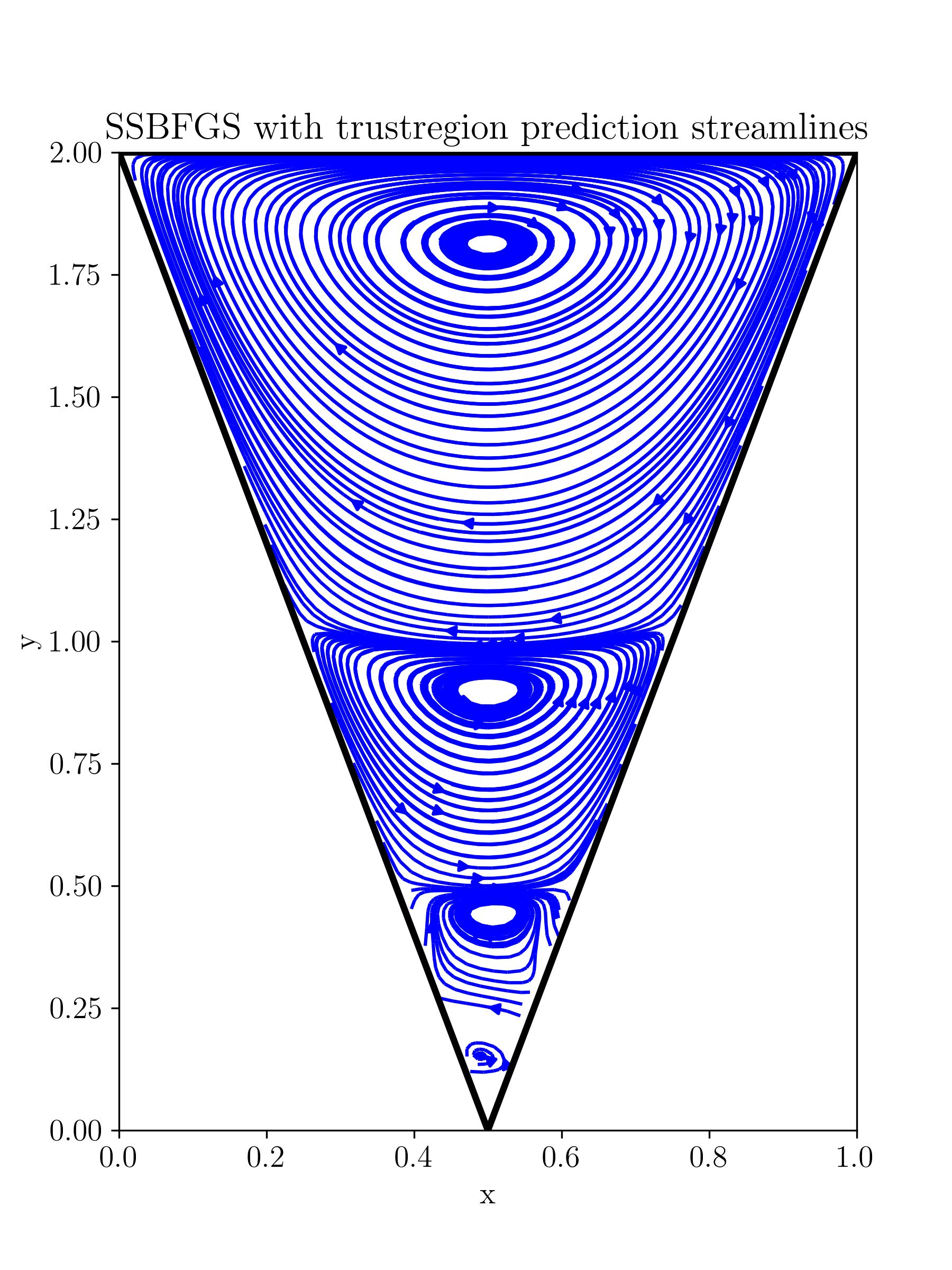}
\caption{SS-BFGS Trustregion}
\end{subfigure}
\begin{subfigure}[b]{0.5\linewidth}
\includegraphics[trim={0 1cm 0 0},clip, width=\linewidth]{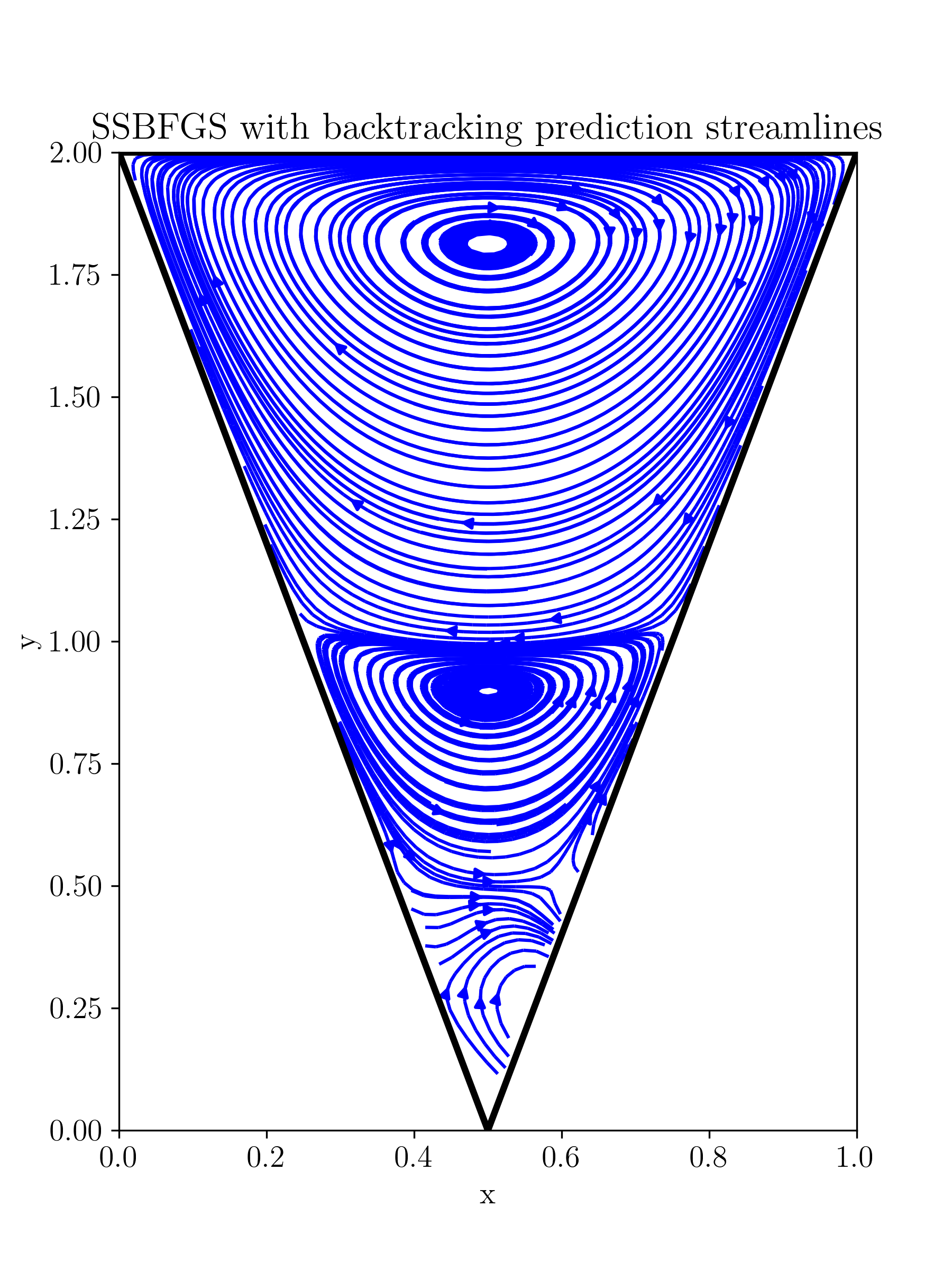}
\caption{SS-BFGS backtracking}
\end{subfigure}
\begin{subfigure}[b]{0.5\linewidth}
\includegraphics[trim={0 1cm 0 0},clip, width=\linewidth]{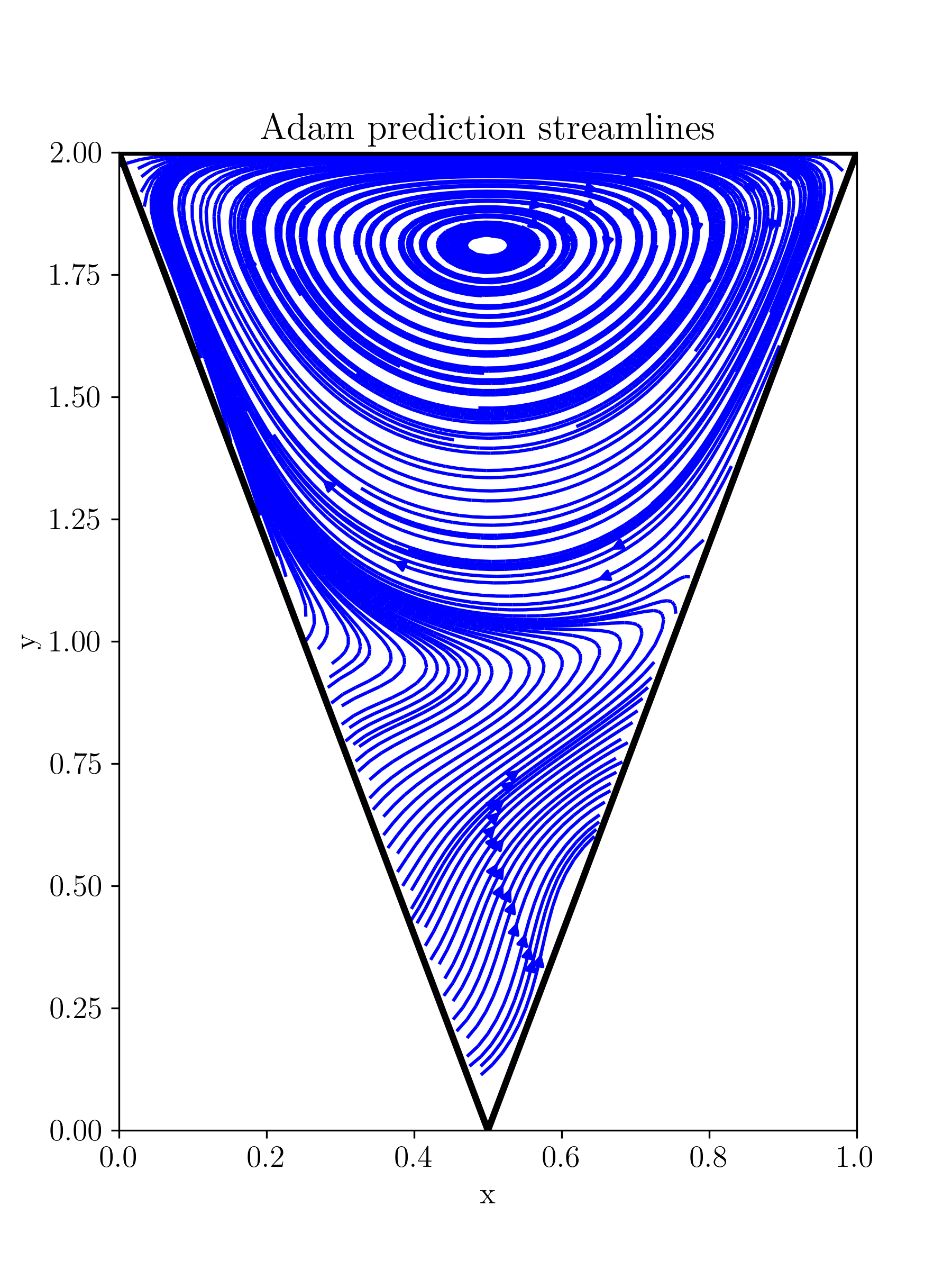}
\caption{Adam optimizer}
\end{subfigure}
\begin{subfigure}[b]{0.6\linewidth}
\includegraphics[trim={0 1cm 0 1cm},clip, width=\linewidth]{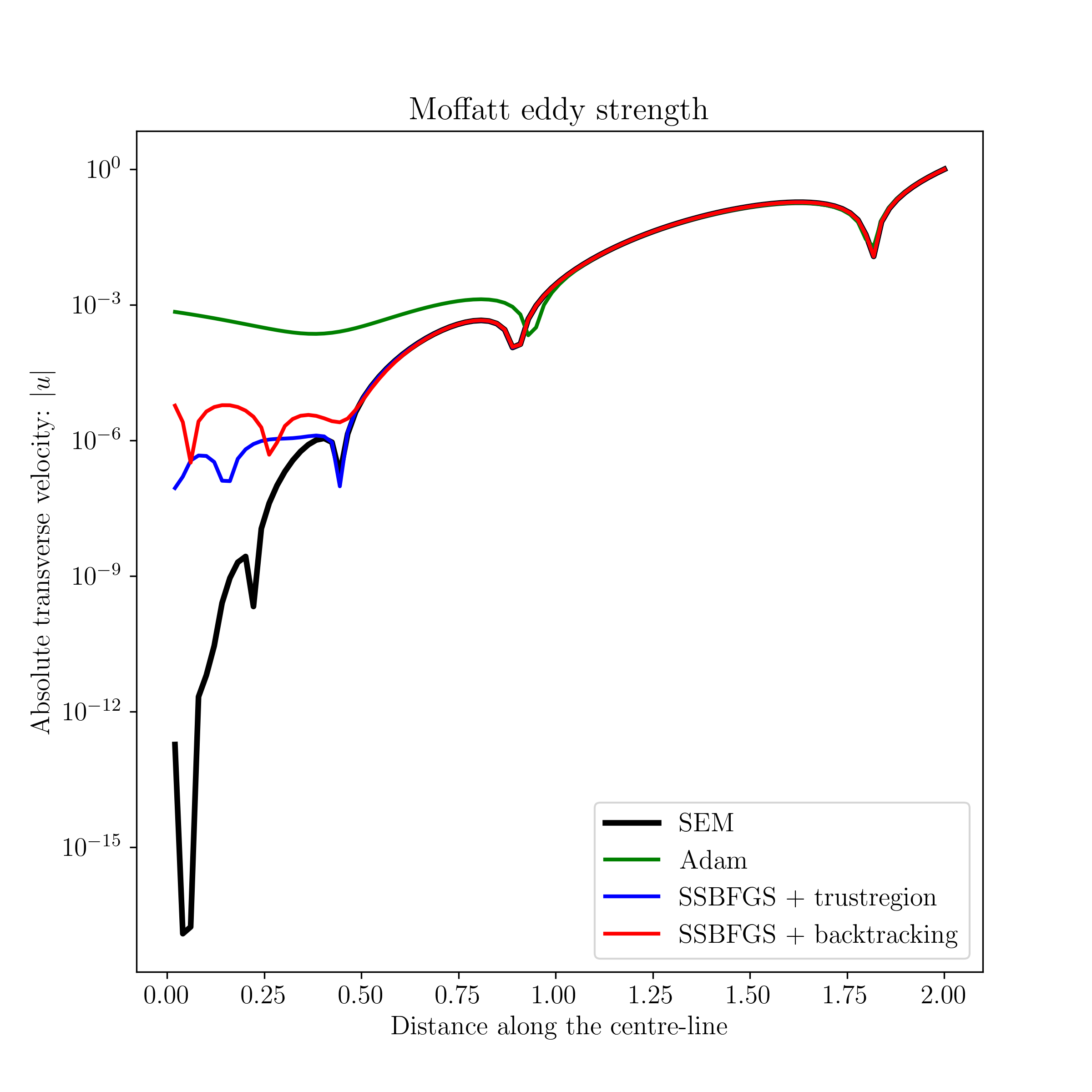}
\caption{Adam optimizer}
\end{subfigure}
\caption{Solutions of \autoref{eq:Stokes} obtained using PINN optimized with (a) SSBFGS optimizer with trustregian based linesearch and subfigure (b) represents the streamlines obtained  by pairing the SSBFGS with backtracking linesearch. (c) represents the streamlines obtained while using Adm optimizer. Subfigure (d) shows comaprison of centre-line (shown with dashed line in \autoref{fig:stokes_dom} transverse velocity $(|u|)$ in (a), (b) and (c) also known as Moffatt eddy strength.}
\label{fig:Stream_Stokes}
\end{figure}

To evaluate the accuracy of each optimizer, we visualize the resulting streamlines in Figure~\ref{fig:Stream_Stokes}. As illustrated (Figure~\ref{fig:Stream_Stokes}a), the combination of SSBFGS with the trust-region line search successfully recovers all four eddies, including the weakest one with a velocity magnitude of $|u| = 10^{-8}$. In contrast, SSBFGS with backtracking (Figure~\ref{fig:Stream_Stokes}b) captures only the dominant eddy accurately, failing to resolve the weaker structures. The streamlines produced using the Adam optimizer (Figure~\ref{fig:Stream_Stokes}c) do not accurately capture any eddies. It is worth noting that although both SSBFGS variants yield similar relative $L_2$ errors, this metric reflects the mean error and does not necessarily indicate uniform convergence. Consequently, the streamlines generated by SSBFGS with backtracking significantly deviate, as streamlines are highly sensitive to the precision of the velocity components $(u, v)$. Finally, in Figure~\ref{fig:Stream_Stokes}d, we present the Moffatt eddy strength~\cite{moffatt1964viscous}, which compares the centerline transverse velocity magnitude $(|u|)$—indicated by the dashed line in Figure~\ref{fig:stokes_dom}—as extracted from Figure~\ref{fig:Stream_Stokes}a, Figure~\ref{fig:Stream_Stokes}b, and Figure~\ref{fig:Stream_Stokes}c. It is evident that the SSBFGS optimizer with trust-region line search most accurately captures the weakest eddies, outperforming both the backtracking variant and the Adam optimizer.

\section{Optimizing Data-Driven DeepONet with Self-Scaled Optimizers}\label{sec:deeponet}

\begin{figure}[!tbh]
    \centering
    \begin{minipage}[b]{\linewidth}
        \centering
        \includegraphics[width=0.99\linewidth]{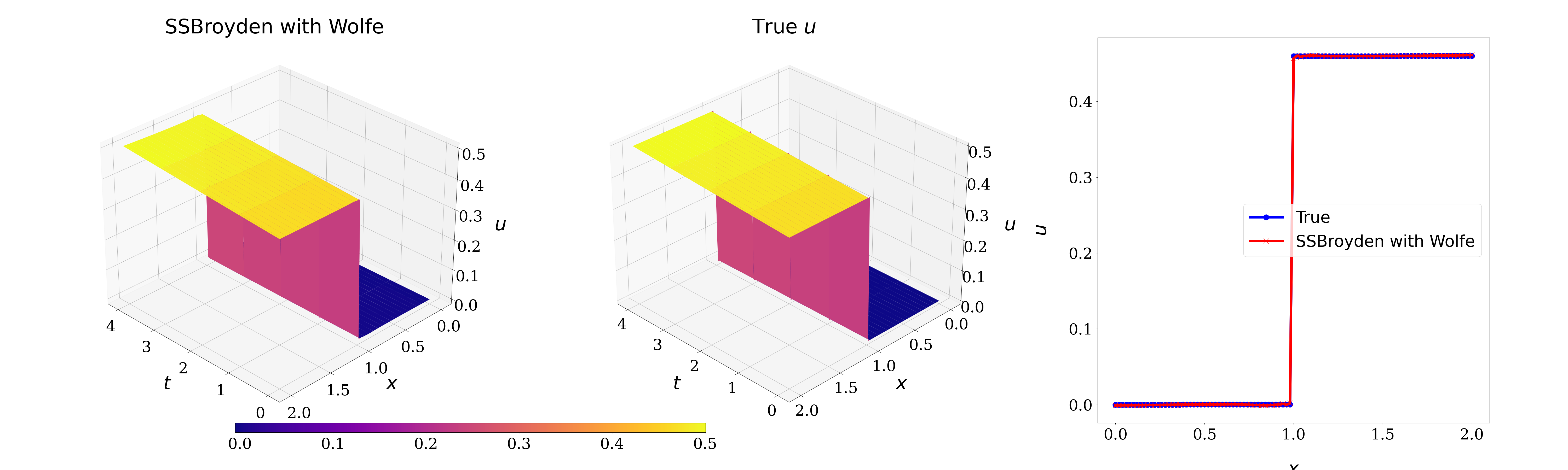}
        \caption*{\textbf{(a)} Comparison between predicted and true displacement $u$ using SSBroyden.}
    \end{minipage}
    
    \vspace{1em}
    \begin{minipage}[b]{0.99\linewidth}
        \centering
        \includegraphics[width=\linewidth]{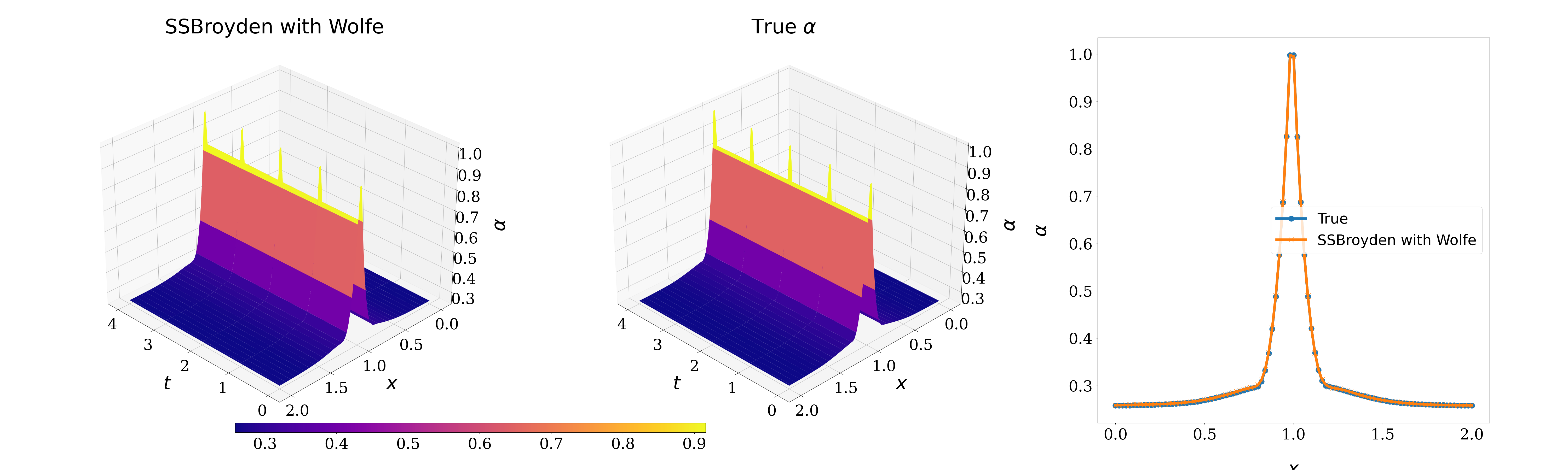}
        \caption*{\textbf{(b)} Comparison between predicted and true damage $\alpha$ using SSBroyden.}
    \end{minipage}
    \caption{\textbf{Double-Precision DeepONet.} Comparison of predicted and true physical fields using the SSBroyden optimizer: (a) displacement $u$, and (b) damage $\alpha$.}
    \label{fig:compare_u_alpha_ssb}
\end{figure}

In this section, we extend our study to evaluate the performance of quasi-Newton optimizers BFGS, SSBFGS, and SSBroyden for DeepONet in a purely data-driven setting. 
Deep Operator Networks (DeepONets) are neural architectures designed to learn operators—mappings between function spaces—and are particularly effective for solving parametric PDEs~\cite{lu2019deeponet}. The model consists of two subnetworks, a branch network that encodes input functions sampled at discrete sensor points, and a trunk network that encodes the spatial coordinates at which the output is evaluated. The final output is obtained via an inner product between the branch and trunk outputs, approximating the operator mapping \( \mathcal{G}: F \mapsto G \). For full architectural and theoretical details, see~\cite{lu2019deeponet}.

We consider a one-dimensional homogeneous bar under increasing applied displacement to study crack nucleation in a simplified setting. Full details of the dataset are available in our previous work~\cite{kiyani2025predicting}. The dataset consists of phase field \( \alpha \) and displacement field \( u \), sampled at 101 equispaced spatial nodes over 50 applied displacements \( U_t \in [0.01, 0.5] \). The first 45 displacement steps are used for training, and the remaining 5 for testing. A DeepONet is employed, where the branch network encodes the applied displacements while the trunk network encodes the spatial coordinates. The branch network has 2 hidden layers and the trunk network has 3 hidden layers, each with 20 neurons. 

\begin{figure}[!tbh]
    \centering
    \begin{minipage}[b]{\linewidth}
        \centering
        \includegraphics[width=0.99\linewidth]{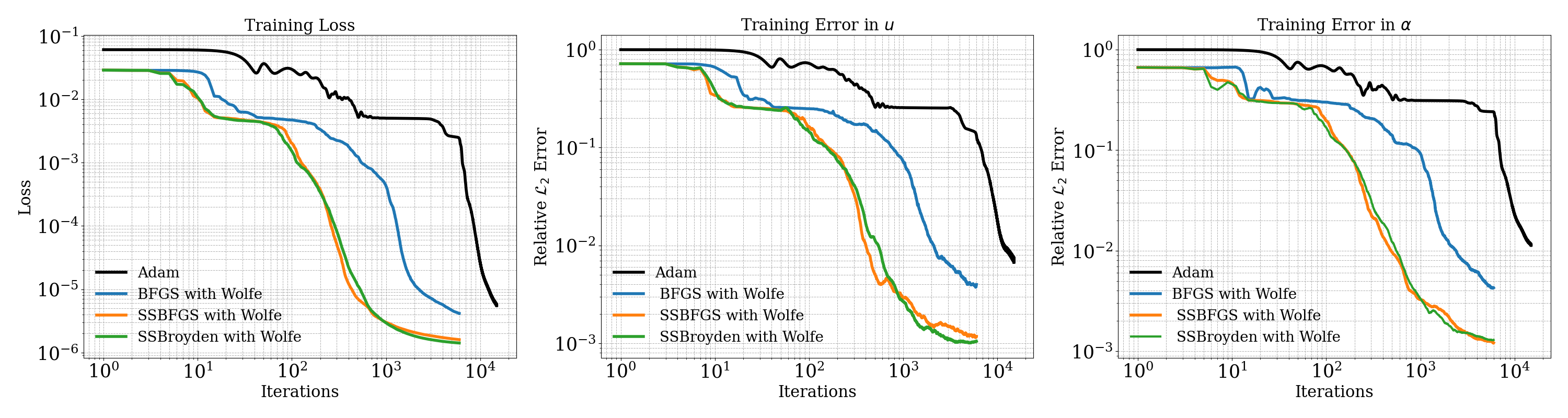}
        \caption*{\textbf{(a)} Training loss and error over iterations.}
    \end{minipage}
    \vspace{1em}
    \begin{minipage}[b]{0.45\linewidth}
        \centering
        \includegraphics[width=\linewidth]{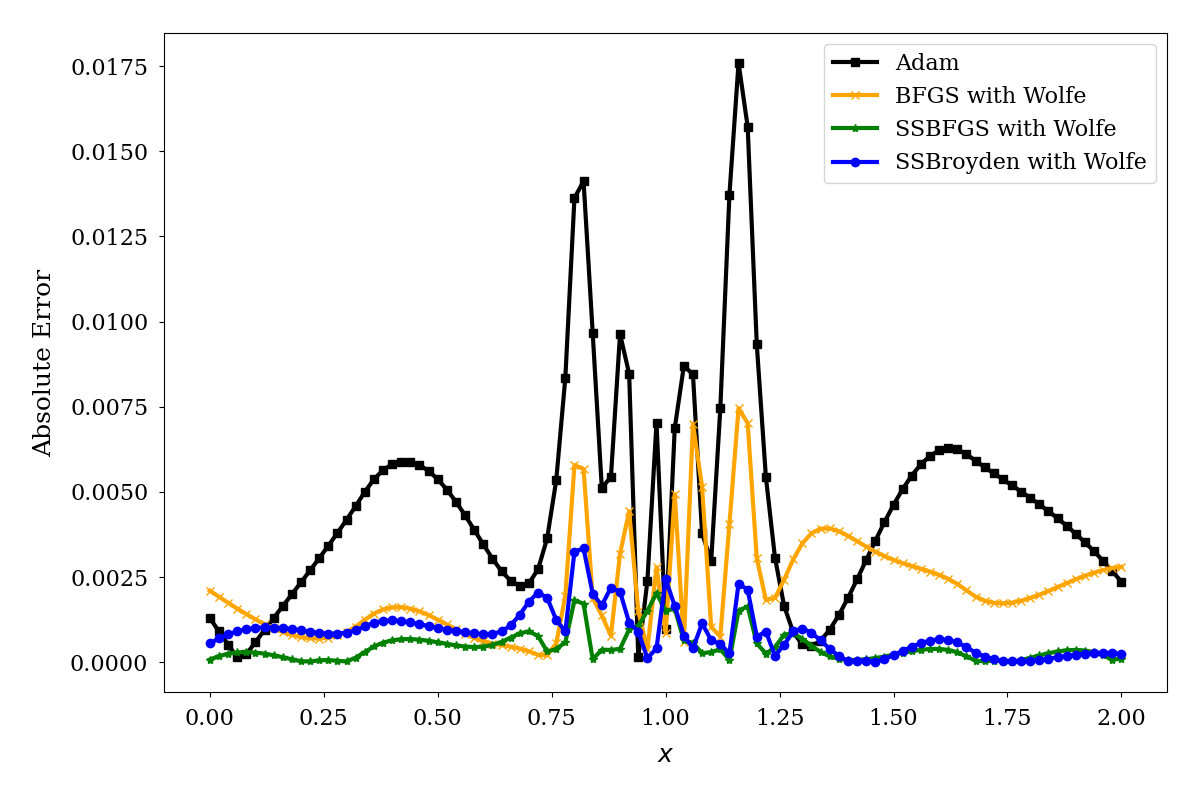}
        \caption*{\textbf{(b)} Relative error in damage $\alpha$.}
    \end{minipage}
    \hfill
    \begin{minipage}[b]{0.45\linewidth}
        \centering
        \includegraphics[width=\linewidth]{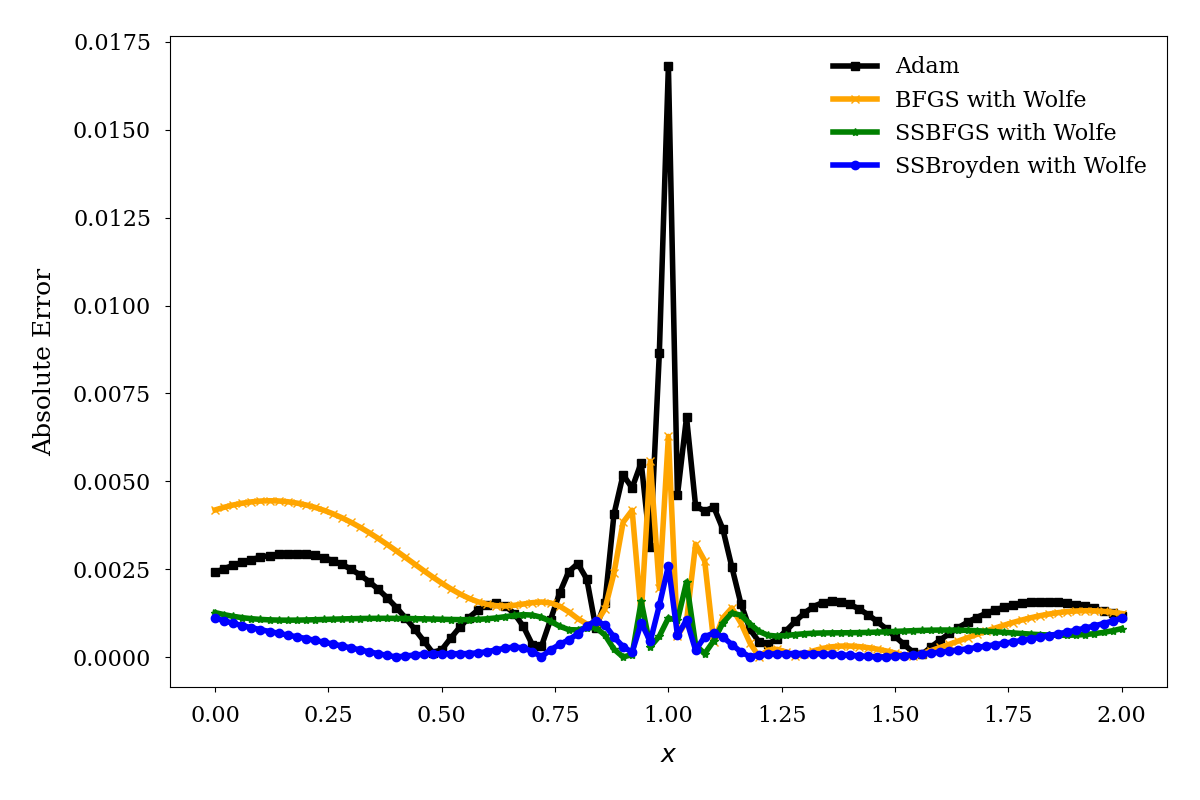}
        \caption*{\textbf{(c)} Relative error in displacement $u$.}
    \end{minipage}
    \caption{\textbf{Double-Precision DeepONet on one-dimensional homogeneous bar.} 
    (a) Loss evolution as well as relative \( l_2 \) error over optimizer iterations. 
    (b) Relative \( l_2 \) error in damage variable $\alpha$. 
    (c) Relative \( l_2 \)  error in displacement field $u$. 
    The table summarizes training performance for different optimization strategies.}
    \label{fig:deeponet_1d_summary}
    \vspace{2em}
    \begin{minipage}[b]{\linewidth}
        \centering
        \rowcolors{2}{cyan!15}{white}
        \scalebox{0.9}{
        \begin{tabular}{|c|l|c|c|}
            \hline
            \rowcolor{cyan!40} 
            \textbf{ID} & \textbf{Optimizer [Adam, BFGS-Type]} & \textbf{Relative \( l_2 \) error (u, $\alpha$)} & \textbf{Training Time (s)} \\
            \hline
            1 & Adam [15000] & $2.1 \times 10^{-2}$, $6.2 \times 10^{-2}$ & 89 \\
            2 & Adam [10] + BFGS with Wolfe [6000] & $5.2 \times 10^{-2}$, $8.6 \times 10^{-2}$ & 399 \\
            2 & Adam [10] + SS-BFGS with Wolfe [6000] & $1.9 \times 10^{-2}$, $1.7 \times 10^{-2}$ & 456 \\
            2 & Adam [10] + SSBroyden with Wolfe [6000] & $5.7 \times 10^{-3}$, $6.9 \times 10^{-3}$ & 444 \\
            3 & Adam [100] + BFGS with Wolfe [6000] & $3.3 \times 10^{-2}$, $4.8 \times 10^{-2}$ & 256 \\
            3 & Adam [100] + SS-BFGS with Wolfe [6000] & $2.9 \times 10^{-2}$, $4.8 \times 10^{-2}$ & 255 \\
            3 & Adam [100] + SSBroyden with Wolfe [6000] & $5.0 \times 10^{-3}$, $8.2 \times 10^{-3}$ & 565 \\
            \hline
        \end{tabular}}
        \captionof{table}{\textbf{Double-Precision DeepONet on one-dimensional homogeneous bar.}: Comparison of optimizer configurations for the 1D fracture problem. Relative \( l_2 \) errors are reported for displacement $u$ and damage field $\alpha$, and total training time is shown in seconds.}
        \label{tab:deeponet_1d_results}
    \end{minipage}
\end{figure}

Figure~\ref{fig:compare_u_alpha_ssb} compares (a) the predicted damage field \( \alpha \) with the true \( \alpha \), and (b) the predicted displacement field \( u \) with the true \( u \), for 5 testing displacement steps.
Figure~\ref{fig:deeponet_1d_summary} presents (a) the training loss for four optimizers—Adam, BFGS, SSBFGS, and SSBroyden—where the loss function includes contributions from both the damage field \( \alpha \) and the displacement field \( u \). It also shows the relative error during training over iterations, demonstrating that SSBroyden and SSBFGS achieve a relative error of \(10^{-3}\) with fewer iterations compared to Adam and BFGS. Subfigures (b) and (c) show the absolute error in predicting the damage field \( \alpha \) and displacement field \( u \), respectively, over the spatial domain. Notably, the error from SSBroyden and SSBFGS remains below 0.0025 for \( \alpha \) and around 0.005 for \( u \), especially near \( x = 1 \), where the phase field transitions sharply from 0 to 1. In contrast, other optimizers exhibit increased errors at this critical region, highlighting the superior stability and accuracy of SSBroyden.
Table~\ref{tab:optimizer_comparison} summarizes the relative errors and training times for several optimization strategies applied to the same network architecture. In all cases, the network size and data remain fixed. 
\textbf{Case 1} uses Adam for 15000 iterations.  
\textbf{Cases 2 and 3} combine a warm-up phase using Adam with 6,000 iterations of quasi-Newton optimizers. Specifically, Case 2 uses only 10 iterations of Adam before switching to BFGS, SS-BFGS, or SSBroyden, while Case 3 uses a longer warm-up of 100 iterations.  
The results show that quasi-Newton methods, particularly SSBroyden, consistently yield lower errors. While SSBroyden requires slightly longer training times, it achieves the best accuracy across both Case 2 and Case 3 configurations.

\section{Summary}\label{summary}

In this study, we conducted a comprehensive comparison of multiple quasi-Newton optimizers for PINNs and PIKANs, focusing on BFGS, SSBFGS, SSBroyden, and L-BFGS with Wolfe line-search conditions, as well as BFGS with Backtracking line search and trust-region methods. Utilizing the \texttt{optax} and \texttt{optimistix} library in JAX, we systematically evaluated these optimization techniques across a range of PDEs. We note that we obtained state-of-the-art results by only employing the best optimizers without fine-tuning our loss functions with self-adaptive or attention-based weights or any other enhancements. We simply pursued a straightforward application of good optimization solvers in order to demonstrate that the big bottleneck for PINNs or PIKANs is the optimization error.

Our initial investigation focused on the Burgers equation, where we evaluated the impact of combining first-order optimizers such as Adam with quasi-Newton optimizers such as BFGS and SSBroyden. We then extended our analysis to more challenging PDEs, including the Allen-Cahn equation, the Kuramoto-Sivashinsky equation, and the Ginzburg-Landau equation, providing a detailed evaluation of efficiency and accuracy.
The results demonstrate that advanced quasi-Newton methods, particularly SSBroyden, SSBFGS and BFGS with Wolfe line-search, significantly improve the convergence rate and accuracy of PINNs, especially for complex and stiff PDEs. Across all cases, SSBroyden consistently outperformed SSBFGS and BFGS, achieving faster convergence and exhibiting greater robustness in handling complex optimization landscapes. These findings highlight the effectiveness of quasi-Newton methods in accelerating the training of PINNs and enhancing numerical stability.
Furthermore, we extended our analysis to PIKANs, replacing traditional multilayer perceptron architectures with KANs utilizing Chebyshev polynomials. In this setting, SSBroyden continued to demonstrate superior optimization performance, reinforcing its robustness across diverse network architectures. Additionally, our study emphasized the importance of implementing PINNs and PIKANs with double-precision arithmetic, which improves numerical stability and enhances optimization efficiency across all tested scenarios.
Moreover, we evaluated the performance of quasi-Newton optimizers —BFGS, SSBFGS, and SSBroyden— for DeepONet in a purely data-driven setting. Specifically, we studied crack nucleation in a one-dimensional homogeneous bar subjected to increasing prescribed displacements.

In Appendix~\ref{appendix:rosenbrock}, we provide a detailed comparison of the number of iterations required to minimize the Rosenbrock function using various optimizers. This analysis highlights how increasing the dimensionality affects the convergence behavior of quasi-Newton methods such as BFGS and SSBroyden. We observed that these methods remain efficient across dimensions, while first-order optimizers like ADAM require significantly more iterations and fail to achieve comparable accuracy.
The errors reported in this study show that using SSBroyden with the Wolfe line search leads to significantly lower relative \( l_2 \) errors in all benchmark problems tested compared to the state-of-the-art SOAP method~\cite{wang2025gradient}. For instance, in the Burgers equation, our approach reduces the error from $8.06 \times 10^{-6}$ to $4.19 \times 10^{-8}$. In the Kuramoto–Sivashinsky equation, we achieve an error of $2.65 \times 10^{-5}$ over the full time interval $t \in [0, 1]$, whereas SOAP reports $3.86 \times 10^{-2}$ over a shorter interval $t \in [0, 0.8]$. Similar improvements are observed for the Allen–Cahn and Ginzburg–Landau equations (see Table~\ref{tab:benchmark_results}). A head-to-head comparison of SSBroyden shows that, unlike other quasi-Newton optimizers, SSBroyden achieves comparable accuracy with dramatically smaller networks—using just five hidden layers of 30 neurons for the Kuramoto-Sivashinsky equation and three hidden layers of 30 neurons for the Allen–Cahn equation—resulting in substantially reduced training time. 

A key question that remains open is the applicability of the proposed family of BFGS-based optimizers for batch training in very large-scale systems. In subsection~\ref{sec:Stokes equation}, we demonstrated the use of the SSBFGS optimizer to train PINNs  with approximately 36,000 parameters, employing 120,000 residual points. However, to fairly compare with first-order optimizers, which efficiently support batch training and thus enable the use of larger neural network architectures, it is important to consider the limitations of BFGS-based methods in such settings. In batch training, the mini-batch changes at each iteration, potentially causing instability in BFGS methods. This is because these methods depend on gradient differences to update Hessian approximations, and the variability in data across batches can introduce inconsistency in those gradients. To address this, \cite{berahas2016multi} proposed the use of overlapping mini-batches for the L-BFGS optimizer to ensure some continuity between iterations and improve the stability of gradient difference estimates. We implemented this multi-batch strategy for the viscous Burgers equation and achieved comparable accuracy to that obtained using full-batch training. Nevertheless, a comprehensive evaluation of multi-batch training for all variants of BFGS optimizers lies beyond the scope of the present study. Such an investigation would require a detailed analysis of how noisy gradient estimates, arising from batch sampling, affect the Hessian approximation, and how the condition number of the Hessian varies with the type of PDE being solved. This is left for future work.

In summary, our study demonstrates that SSBFGS and SSBroyden with Wolfe line-search are effective and reliable optimizers for training PINNs and PIKANs. Their  capacity to handle complex PDEs and their strong convergence properties make them well-suited for advanced applications in scientific machine learning applications. Future research should focus on incorporating domain-specific adaptations to further improve the accuracy, efficiency, and generalization of PINNs and PIKANs, particularly in high-dimensional and multi-scale problem settings.

\begin{minipage}[b]{\linewidth}
  \centering
  \renewcommand{\arraystretch}{1.3}
  \arrayrulecolor{black} 
  {\rowcolors{2}{cyan!15}{white}
  \scalebox{0.85}{ 
  \begin{tabular}{|l|c|c|}
  \hline
  \rowcolor{cyan!40} 
  \textbf{Benchmark} & \textbf{$\mathcal{L}_2$ Error (SOAP~\cite{wang2025gradient})} & \textbf{$\mathcal{L}_2$ Error (SSbroyden with Wolfe)} \\
  \hline
  Burgers Equation & $8.06 \times 10^{-6}$ & $4.19 \times 10^{-8}$ \\
  Allen–Cahn Equation &  $3.48 \times 10^{-6}$ & $9.43 \times 10^{-7}$ \\
  Kuramoto–Sivashinsky Equation &$3.86 \times 10^{-2}$ on $t \in [0, 0.8]$ & $2.65 \times 10^{-5}$ on $t \in [0, 1]$ \\
  Ginzburg–Landau Equation &  $4.78 \times 10^{-3}$ & $2.33 \times 10^{-3}$ \\
  \hline
  \end{tabular}}}    
  \captionof{table}{\textbf{Benchmark comparison of relative $\mathcal{L}_2$ errors.} Results from SSBFGS with Wolfe line-search are compared against SOAP~\cite{wang2025gradient}. SSBFGS demonstrates improved accuracy across all PDE benchmarks.}
  \label{tab:benchmark_results}
\end{minipage}

\section*{Acknowledgments}
This research was primarily supported as part of the AIM for Composites, an Energy Frontier Research Center funded by the U.S. Department of Energy (DOE), Office of Science, Basic Energy Sciences (BES), under Award \#DE-SC0023389 (computational studies, data analysis). Additional funding was provided by the DOE-MMICS SEA-CROGS DE-SC0023191 award, the MURI/AFOSR FA9550-20-1-0358 project, and a grant for 
GPU Cluster for Neural PDEs and Neural Operators  to Support MURI Research and Beyond, under Award \#FA9550-23-1-0671. 
JFU is supported by the predoctoral fellowship
ACIF 2023, cofunded by Generalitat Valenciana and the European Union
through the European Social Fund. JFU acknowledges the support through the grant PID2021-127495NB-I00 funded by MCIN/AEI/10.13039/501100011033 and by the European Union, and the Astrophysics and High Energy Physics programme of the Generalitat Valenciana ASFAE/2022/026 funded by MCIN and the European Union NextGenerationEU (PRTR-C17.I1). 

\appendix

\section{Search Methods and Trust-Region Techniques}
\label{search_methods_trust-region_techniques}

In this section, we provide a detailed overview of search methods and trust-region techniques in the context of quasi-Newton algorithms, focusing on their role in achieving efficient and reliable convergence.

\subsubsection{Line-search methods}
A common approach in optimization is the \textbf{line-search strategy}, where the algorithm first selects a \textbf{search direction} \( \mathbf{p}_k \) that ideally points towards a region of lower function values. Once this direction is determined, the next crucial step is to decide the \textbf{step size} \( \alpha_k \), which dictates how far to move along the chosen direction to achieve sufficient improvement in the objective function \( f \). This step involves solving approximately a one-dimensional minimization problem:

\begin{equation}\label{eq:min_f}
    \min_{\alpha > 0} f(\mathbf{x}_k + \alpha \mathbf{p}_k).
\end{equation}

The selection of both the search direction and step size plays a fundamental role in the convergence behavior and overall effectiveness of the optimization process.
While an exact solution to~\eqref{eq:min_f}  would maximize the benefit of the chosen direction \( \mathbf{p}_k \), finding the exact minimum is often computationally prohibitive. Instead, line search methods typically rely on an approximate solution, evaluating a finite number of trial step lengths \( \alpha_k \) until a suitable reduction in \( f \) is achieved.
This search, which receives commonly the name of \emph{inexact} line search, relies in a series of mathematical conditions to really ensure to obtain an satisfactory step length. 

Once an acceptable step length \( \alpha \) is identified, the algorithm updates the current iterate
\begin{equation}\label{eq:update_x}
    \mathbf{x}_{k+1} = \mathbf{x}_k + \alpha_k \mathbf{p}_k.
\end{equation}
At the new point \( \mathbf{x}_{k+1} \), the process is repeated by selecting a new search direction \( \mathbf{p}_{k+1} \) and step length \( \alpha_{k+1} \). This iterative process continues until convergence criteria are satisfied. 

In this paper, we explored two well-known methodologies that we briefly describe next.
\subsubsection*{Wolfe conditions}
The Wolfe conditions are mathematical criteria used to ensure that a step size in iterative optimization methods satisfies specific properties of sufficiency. Given a point $\textbf{x}_k$ and a direction $\textbf{p}_k$, the Wolfe conditions consist of the following two inequalities:
\begin{align}
    f\left(\textbf{x}_k + \alpha_k \textbf{p}_k \right) & \leq f \left(\textbf{x}_k \right) + c_1 \alpha_k \textbf{p}_k^T \nabla f_k, \label{eq:Wolfe1} \\
    \nabla f\left(\textbf{x}_k + \alpha_k \textbf{p}_k \right)^T \textbf{p}_k & \geq c_2\nabla f \left(\textbf{x}_k \right)^T \textbf{p}_k, \label{eq:Wolfe2}
\end{align}
which commonly receive the names of \emph{Armijo} and \emph{sufficient decrease} conditions, respectively. The quantities $c_1$ and $c_2$ are constants that should follow $0 < c_1 < c_2 < 1$. While the first condition ensures a sufficient decrease in the function $f$, the second one rules out unacceptably short steps. The latter inequality is commonly replaced by taking absolute values at both sides
\begin{equation}
    \left|\nabla f\left(\textbf{x}_k + \alpha_k \textbf{p}_k \right)^T \textbf{p}_k \right|  \leq c_2 \left|\nabla f \left(\textbf{x}_k \right)^T \textbf{p}_k \right|. \label{eq:SWolfe2}
\end{equation}
The conditions \eqref{eq:Wolfe1} and \eqref{eq:SWolfe2} together receive the name of Strong Wolfe conditions.
 For all tests we set $c_1 = 10^{-4}$ and $c_2 = 0.9$, which are the standard choices in the optimization literature.
\subsubsection*{Backtracking}
By only using equation \eqref{eq:Wolfe1}, sufficient progress between iterates is not guaranteed, as it does not exclude unacceptably low values of $\alpha_k$. Because of that, the Wolfe or Strong Wolfe conditions introduce an additional condition given by \eqref{eq:Wolfe2} or \eqref{eq:SWolfe2}, respectively. However, another line-search strategy that has proven numerically to be successful and does not need an additional condition apart from the Armijo one is backtracking~\cite{armijo1966minimization}. In a backtracking strategy the step-length is chosen in a more systematic way; instead of evaluating the objective function multiple times for various step-lengths, we start with a reasonable (but not small) initial value of $\bar{\alpha}$. If the condition is satisfied, the step-size \( \alpha \) is accepted (that is, we set $\alpha_k = \bar{\alpha}$), and the algorithm proceeds to the next iteration. Otherwise, the step size is reduced by multiplying it by the factor \( \rho < 1 \), i.e., \( \bar{\alpha} \leftarrow \rho \bar{\alpha} \). The process is then repeated multiple times until the Armijo condition \eqref{eq:Wolfe1} is met. Additionally, we incorporate the following condition along with \eqref{eq:Wolfe1}:
\begin{align}\label{eq:pred_red}
\textbf{p}_k^ {\top} \nabla f_k \le 0.
\end{align}

\begin{figure}[!tbh]
\includegraphics[width=1\textwidth]{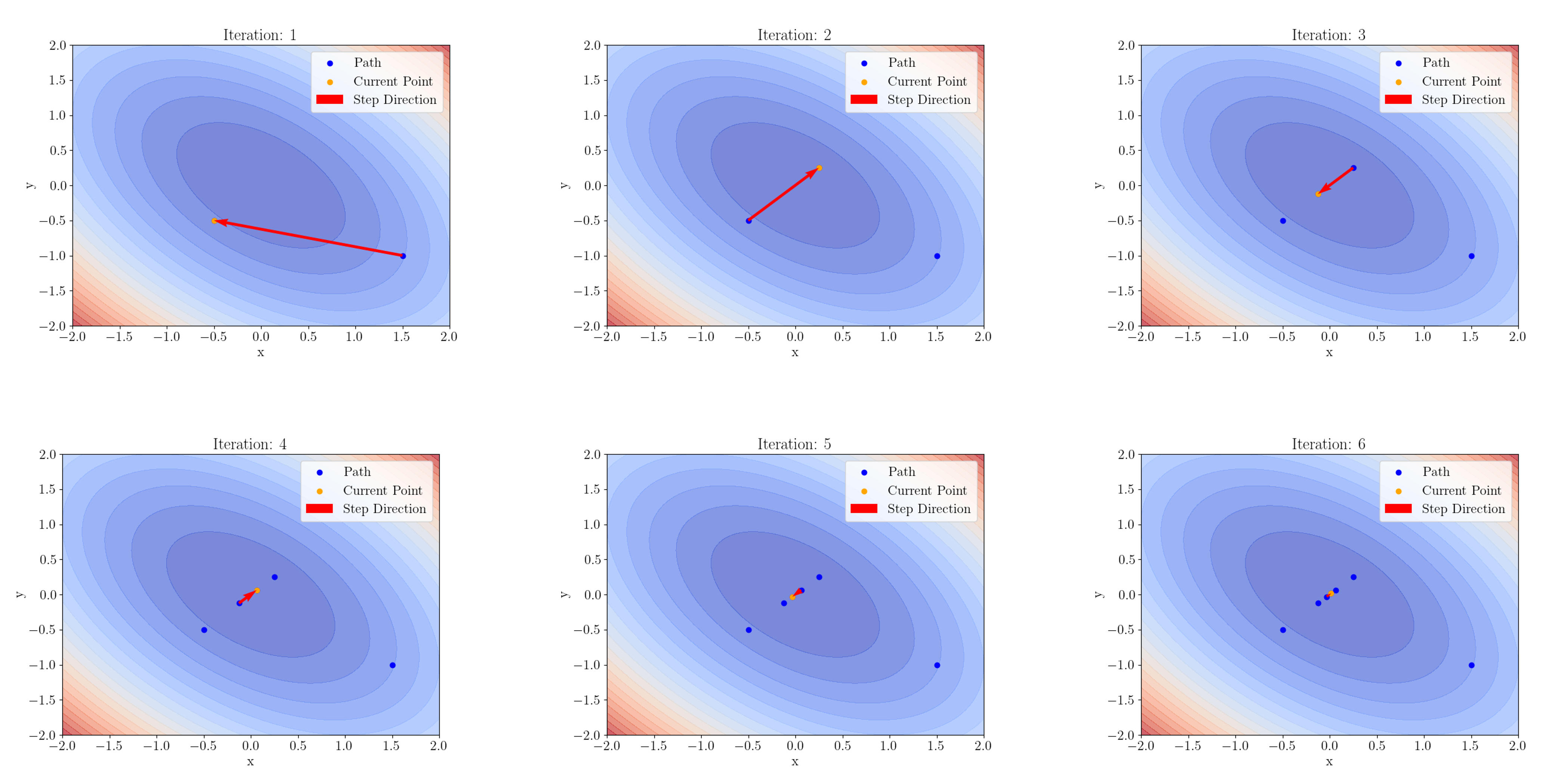}
\caption{An example illustrating the minimization of a quadratic function, \( f(x, y) = x^2 + y^2 + xy \), using the backtracking line-search algorithm is presented. The convergence trajectory is shown over six iterations. Notably, the search direction changes at each iteration, determined by evaluating the inequality condition. It is worth mentioning that backtracking achieves convergence to the minimum value of the level set faster than the trust-region approach.}
\label{fig:back_track}
\end{figure}

\subsubsection{Trust-region methods}
Another class of optimization methods is the family of trust-region methods. The key distinction between trust-region and line-search methods is that the former defines a region (the trust-region) around the current point \( \mathbf{x}_k \) and seeks a suitable point within this region that reduces the objective function. To achieve this, a trust-region algorithm first constructs a local quadratic approximation of the function, denoted as \( m_k \), and then approximately solves the following subproblem:

\[
   \min_\textbf{p} \, m_k(\textbf{p}) = \nabla f_k^\top \textbf{p} + \frac{1}{2} \textbf{p}^\top \textbf{B}_k \textbf{p}
   \]
   subject to:
   \[
   \| p \| \leq \Delta_k,
   \]

where $\Delta_k$ is the radius of the trust-region, and $\textbf{B}_k$ is some approximation of the Hessian matrix at $\textbf{x}_k$. Then, the proposed step \( \textbf{p}_k \) is evaluated. If it results in a significant improvement in the objective function, the step is accepted and the trust-region can be expanded. If the step yields poor results, it is rejected, and the trust-region is reduced. 
Finally, the iterate $\textbf{x}_k$ is updated simply as 
\[
\textbf{x}_{k+1} = \textbf{x}_k + \textbf{p}_k,
\]
that is, in a trust-region algorithm the direction and the step-length are calculated at the same time. 
The choice between \textbf{backtracking line-search} and the \textbf{trust-region method} depends on the problem characteristics and specific implementation details. Backtracking line-search is often preferable for simpler or computationally inexpensive problems due to its straightforward nature and adaptability. In contrast, trust-region methods are more suitable for complex or constrained optimization tasks where careful control over the step size is essential.
A key distinction is that backtracking line-search directly adjusts the step size without solving quadratic subproblems, making it computationally faster in cases where objective function and gradient evaluations are inexpensive. Ultimately, the choice between these methods depends on the specific problem and computational trade-offs.

The final piece of the algorithm is to choose the convergence criteria. The algorithm is considered to have converged if the norm of the gradient satisfies

\begin{align}\label{eq:conv_criteria}
   \|\nabla f(\bm{x}_k)\| \leq \epsilon,
\end{align}

where \( \|\nabla f(x_k)\| \) is the Euclidean norm of the gradient at iteration \(k\), \( \epsilon > 0 \) is a pre-specified tolerance level (a small positive value, e.g., \( 10^{-6} \)).
The gradient of the objective function represents the direction and magnitude of the steepest ascent. At a local minimum, the gradient approaches zero. Thus, the norm of the gradient serves as a natural stopping criterion. A small tolerance (\( \epsilon \)) improves precision but may increase computational cost. 
In some cases, the norm of the gradient stagnates near zero due to numerical precision issues, requiring additional criteria such as step size or objective function change thresholds.

In addition to using the gradient norm as a convergence criterion, the following supplementary criteria can be applied:

\begin{enumerate}
\item If \( |f(\bm{x}_k) - f(\bm{x}_{k-1})| \leq \delta \), where \( \delta \) is a small threshold.
\item If \( \|\bm{x}_k - \bm{x}_{k-1}\| \leq \eta \), where \( \eta \) is a small tolerance.
\item Set a maximum number of iterations to prevent infinite loops in cases of slow convergence.
\end{enumerate}

\begin{figure}[!tbh]
\includegraphics[width=1\textwidth]{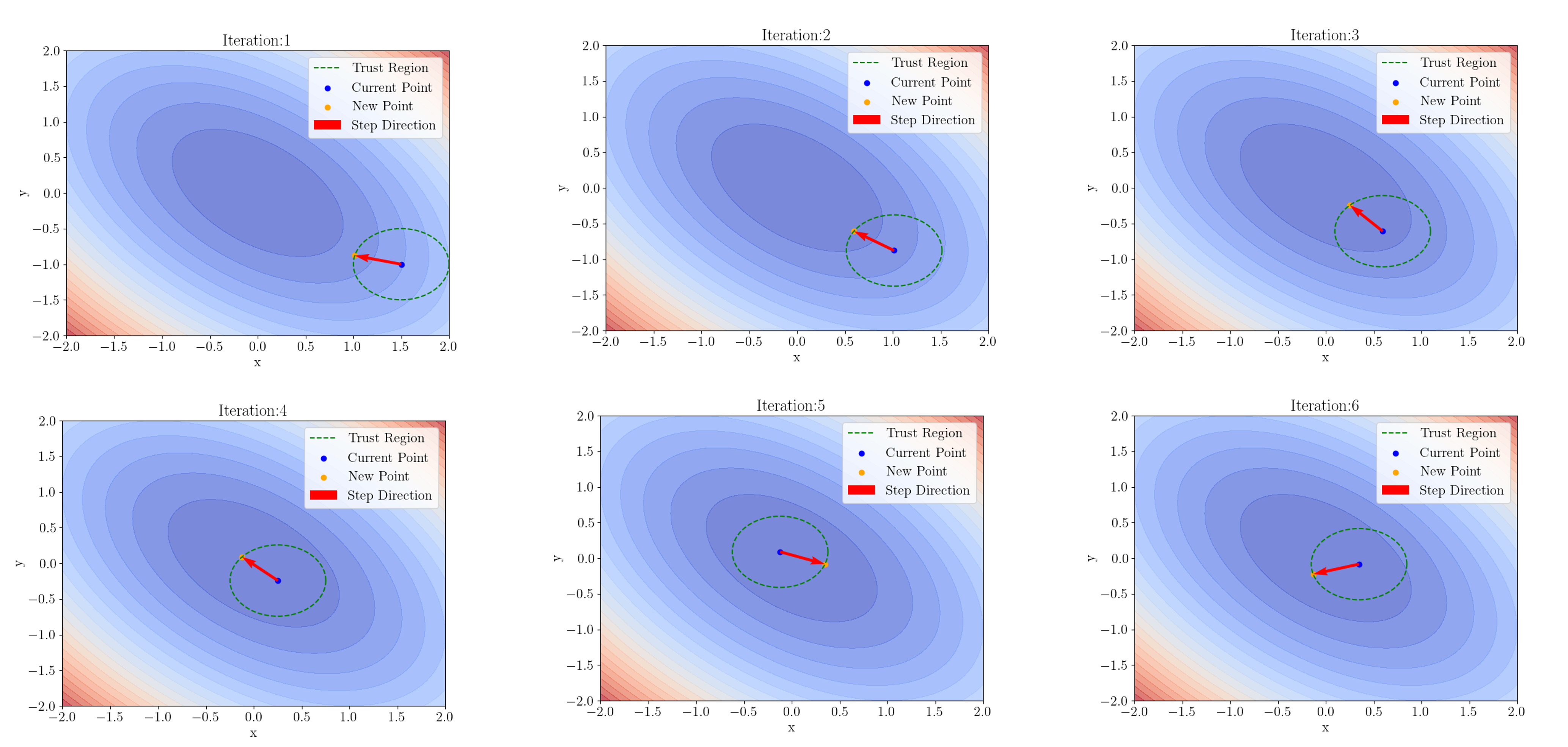}
\caption{An example showing the minimization of a quadratic function \( f(x, y) = x^2 + y^2 + xy\) using trust-region. The convergence is shown for six iteration. It is to be noted that at every iteration search direction changes computed by minimizing the line search criteria. In this example, we show that the radius of trust-region is constant as \( f(x)\) is convex but for PDE problem we have implemented the daptive size of trust-region.}
\label{fig:trust_region_ex}
\end{figure}

To demonstrate the BFGS algorithm with the backtracking line search and trust-region, we present the convergence trajectory for a quadratic function defined as
\[
f(x, y) = x^2 + y^2 + xy,
\]
in Figures~\ref{fig:back_track} and~\ref{fig:trust_region_ex}. We see that from iteration 1 to 6, the algorithm iteratively predicts the search direction. In this example, the trust-region radius is kept constant since \( f(x) \) is convex. However, for PDE problems, we have implemented an adaptive trust-region size.

\section{Detailed Case Study: Burgers Equation in Single Precision}\label{Burgers_Equation_Single_Precision}
In this section, we provide additional details and case studies on the performance of quasi-Newton methods for solving the Burgers equation, with a particular focus on results obtained in single precision.
We begin by examining the performance of SSBroyden, SSBFGS, and BFGS using Wolfe line-search.

\begin{figure}[!tbh]
    \begin{minipage}[b]{\linewidth}
        \centering
        \includegraphics[width=\linewidth]{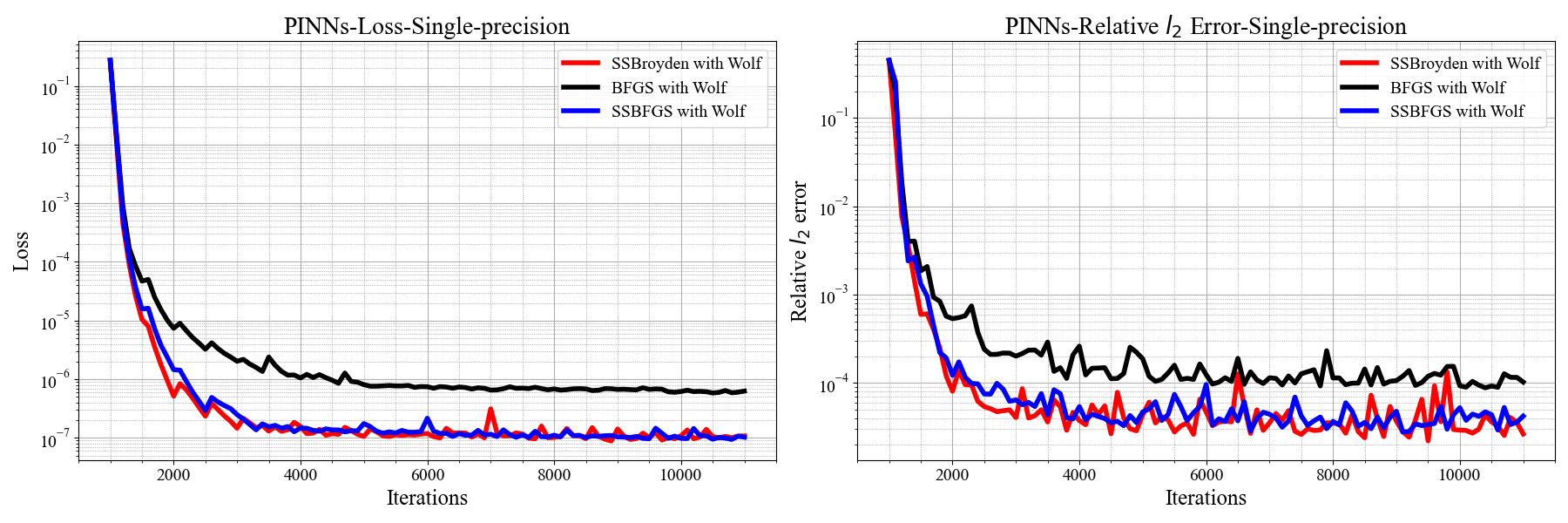}
        \caption{\textbf{PINNs with single precision for the Burgers equation.} 
        Results are shown with Wolfe line search. The plots show the evolution of the loss function (left) and the corresponding \(l_2\) relative error (right) over training iterations for \textbf{Case 1}. After approximately 4,000 iterations, both the loss and error stabilize, indicating no significant further improvement.}
        \label{fig:PINNs_burger_single_combined}
    \end{minipage}
    
    \vspace{1em}
    
    \begin{minipage}[b]{\linewidth}
        \centering
        \rowcolors{2}{cyan!15}{white} 
        \scalebox{0.85}{
        \begin{tabular}{|c|c|c|c|}
            \hline
            \rowcolor{cyan!40}
            \textbf{Optimizer [\# Iters.]} & \textbf{Relative \(l_2\) Error} & \textbf{Training Time (s)} & \textbf{Total Params} \\ \hline
            Adam [1000] + BFGS with Wolfe [10000]      & \(1.04 \times 10^{-4}\) & 263 & 1,341 \\ \hline
            Adam [1000] + SSBroyden with Wolfe [10000] & \(3.60 \times 10^{-5}\) & 209 & 1,341 \\ \hline
            Adam [1000] + SSBFGS with Wolfe [10000]    & \(4.04 \times 10^{-5}\) & 179 & 1,341 \\ \hline
        \end{tabular}
        }
        \captionof{table}{\textbf{PINNs with single precision for the Burgers equation.} 
        Relative \(l_2\) error and training time for solving the Burgers equation using single precision with different optimizers. A PINN with four hidden layers, each containing 20 neurons, is trained for 1,000 iterations using the Adam optimizer, followed by 10,000 iterations using the specified quasi-Newton method.}
        \label{tab:PINNs_burger_single_precision}
    \end{minipage}
\end{figure}


\begin{figure}[!tbh]
  \begin{minipage}[b]{\linewidth}
    \centering
    \includegraphics[trim={0cm 0 0cm 0},clip, width=\textwidth]{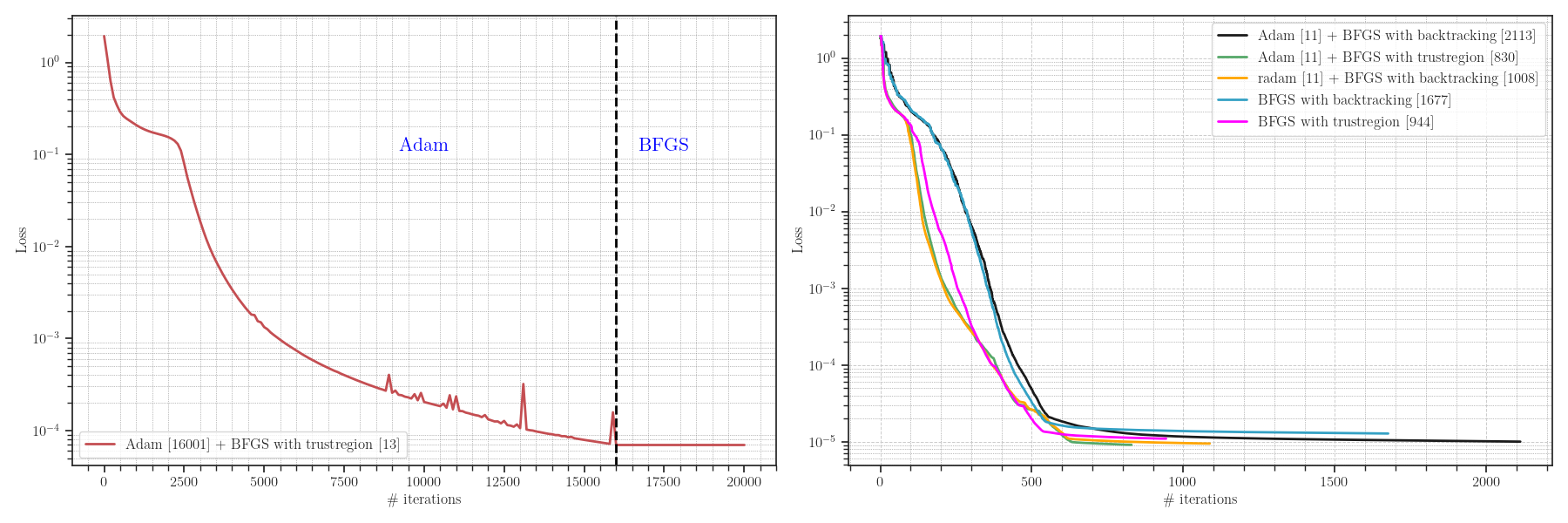}
\captionof{figure}{\textbf{Scenario 1: PINNs with single precision for the Burgers equation.} Results shown with backtracking and trust-region strategies. The GPU is fully saturated with model parameters to minimize latency effects on compute time: The left panel illustrates a scenario where the iteration count of the first-order Adam algorithm dominates, combined with BFGS using the trust-region approach. This pairing exhibits degeneracy due to limited numerical precision during BFGS iterations. The right panel presents the convergence behavior of BFGS paired with trust-region and backtracking line search methods. In this setup, Adam is applied for 11 iterations as a warmup. The BFGS-based optimizer requires significantly more iterations than first-order optimization methods like Adam and Rectified Adam (RAdam) \cite{liu2019variance}.}
    \label{fig:sp_optax_mp}
  \end{minipage}\hfill
  \begin{minipage}[b]{\linewidth}
    \centering
    \rowcolors{2}{cyan!15}{white}
    \scalebox{0.77}{
    \begin{tabular}{|c|c|c|c|}
    \hline
    \rowcolor{cyan!40} 
    \textbf{Optimizer[\# Iters.], Line-search algorithm [\#Iters.]} & \textbf{Relative $l_2$ error} & \textbf{Training time (s)} & \textbf{Total params} \\ 
    \hline
    Adam [16001] + BFGS with trust-region [13] &  \(2.91 \times 10^{-3}\)     & 127    & 3501             \\ 
    \hline
    Adam [11] + BFGS with backtracking [2113] &  \(7.25 \times 10^{-4}\)      &  61               & 3501               \\ 
    \hline
    Adam [11] + BFGS with trust-region [830] & \(\mathbf{6.42} \times \mathbf{10^{-4}}\)     & 56                  & 3501              \\ 
    \hline
    RAdam [11] + BFGS with backtracking [1008]  &         \(7.67 \times 10^{-4}\)  & 62        &   3501              \\ 
    \hline
    BFGS with backtracking [1677]  &         \(2.48 \times 10^{-3}\)  & 45        &    3501            \\ 
    \hline
    BFGS with trust-region [944] & \(1.39 \times 10^{-3}\)     & 49                  & 3501             \\ 
    \hline
    \end{tabular}}
    \captionof{table}{\textbf{[Corresponding to Figure~\ref{fig:sp_optax_mp}] PINNs with single precision for the Burgers equation}: The relative $L_2$ error, training time, and number of training parameters are evaluated for solving the viscous Burgers equation with various optimizers and combinations of line search algorithms. The convergence criteria are based on absolute and relative tolerances in norm of the gradient of the loss function, which is set as $[\text{ATOL}, \text{RTOL}] = [10^{-7}, 10^{-7}]$. This experiments shows that Adam combined with BFGS trust-region provides better accuracy (hghlighted in bold fonts) \label{tab:sp_optax_mp}.}
    \end{minipage}
\end{figure}

\begin{figure}[!tbh]
  \begin{minipage}[b]{\linewidth}
    \centering
    \includegraphics[trim={0cm 0 0 0}, clip, width=\textwidth]{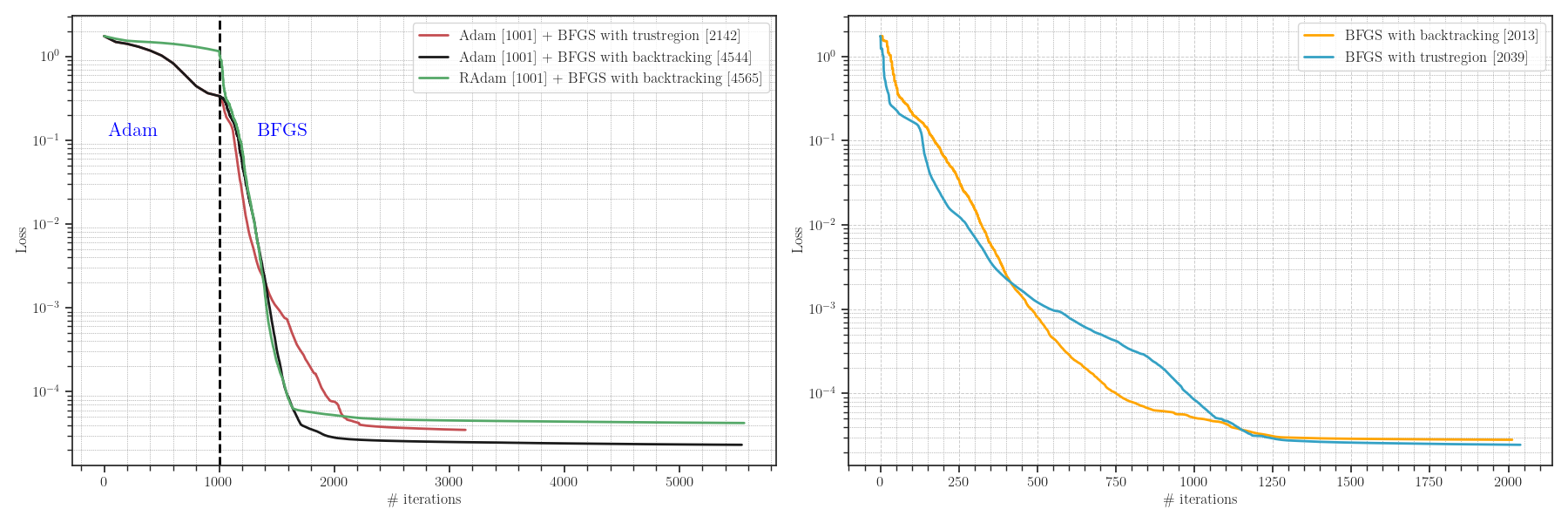}
    \captionof{figure}{\textbf{Scenario 2: PINNs with single precision for the Burgers equation.} 
    Results are shown with backtracking and trust-region strategies. The GPU is fully saturated by collocation points and data points (initial and boundary conditions) to mitigate the impact of latency on computation time. 
    The left panel depicts a scenario where the iteration count for the first-order Adam and RAdam optimizers is fixed at 1001, followed by the quasi-Newton BFGS optimizer using trust-region and backtracking line search algorithms. Similar to the results in Figure~\ref{fig:sp_optax_mp}, this setup also exhibits degeneracy during BFGS iterations due to limited numerical precision.
    The right panel shows the convergence behavior of the standalone BFGS optimizer with trust-region and backtracking strategies, without any warmup from a first-order optimizer. In this configuration, the loss stagnates at $10^{-4}$ due to insufficient precision, particularly affecting the inversion of the Hessian matrix required for descent.}
    \label{fig:sp_optax_cp}
  \end{minipage}\hfill
  \begin{minipage}[b]{\linewidth}
    \centering
\rowcolors{2}{cyan!15}{white} 
\scalebox{0.78}{
\begin{tabular}{|c|c|c|c|}
\hline
\rowcolor{cyan!40} 
\textbf{Optimizer[\# Iters.], Line-search algorithm [\#Iters.]} & \textbf{Relative $l_2$ error} & \textbf{Training time (s)} & \textbf{Total params} \\ 
\hline
Adam [1001] + BFGS with trust-region [2142] &  \(1.51 \times 10^{-3}\)     & 30    & 1341            \\ 
\hline
Adam [1001] + BFGS with backtracking [4544] &  \(1.41 \times 10^{-3}\)      &   43   & 1341               \\ 
\hline
radam [1001] + BFGS with backtracking [4565]  &         \(1.43 \times 10^{-3}\)  & 41    &   1341              \\ 
\hline
BFGS with backtracking [2013]  &         \(1.65 \times 10^{-3}\)  & 18        &    1341          \\ 
\hline
BFGS with trust-region [2039] & \(\mathbf{1.1} \times \mathbf{10^{-3}}\)   & 24               & 1341             \\ 
\hline
\end{tabular}}
\captionof{table}{[Corresponding to Figure~\ref{fig:sp_optax_cp}] Performance metrics for the viscous Burgers equation using single-precision (32-bit) arithmetic for the cases shown in Figure~\ref{fig:sp_optax_cp}: The relative $L_2$ error, training time, and number of training parameters are reported for solving the viscous Burgers equation with various optimizers and combinations of line search algorithms. The convergence criteria are defined by absolute and relative tolerances on the gradient norm of the loss function, set as $[\text{ATOL}, \text{RTOL}] = [10^{-7}, 10^{-7}]$. The results demonstrate that BFGS with the trust-region line search achieves the highest accuracy (highlighted in bold) among all cases presented in Figure~\ref{fig:sp_optax_cp}.}
\label{tab:sp_optax_cp}
\end{minipage}
\end{figure}

Figure~\ref{fig:PINNs_burger_single_combined} illustrates the loss function and relative error over training iterations for SSBroyden, SSBFGS, and BFGS in single precision, corresponding to \textbf{Case 1} in Table~\ref{tab:PINNs_burger_double_precision}. The results are obtained using a PINN with four layers, each comprising 20 neurons, trained with the Adam optimizer followed by 10,000 iterations of the specified optimizer. Table~\ref{tab:PINNs_burger_single_precision} provides a summary of the corresponding relative error and training time. Notably, SSBFGS and SSBroyden achieve a relative error of \(10^{-5}\), while BFGS converges to approximately \(10^{-4}\). It is observed that the loss function and relative error stabilize after roughly 4,000 iterations, indicating no significant further improvement.

The performance of five different optimizer combinations is summarized in Table \ref{tab:sp_optax_mp} for \textbf{Scenario 1}. The convergence history achieved using a neural network with 8 hidden layers, each containing 20 neurons, a \(\tanh\) activation function, 200 randomly sampled data points for the initial and boundary conditions, and 10,000 collocation points to calculate the residual loss. The left panel of Figure \ref{fig:sp_optax_mp} shows a scenario where the iteration count of the first-order Adam algorithm dominates, combined with BFGS using the trust-region line search. The key observation from this setup is that when Adam is run for an extended period, the BFGS optimizer provides no additional advantages due to the degeneration of the loss function caused by single-precision floating-point representation [Row 1 of Table \ref{tab:sp_optax_mp}]. However, when Adam or RAdam is used with a small number of iterations (11 in this case) as a warm-up followed by BFGS with both line search algorithms, the error improves by an order of magnitude [Rows 2, 3, and 4 of Table \ref{tab:sp_optax_mp}]. In contrast, starting directly with the BFGS optimizer results in errors of a similar magnitude to those in the first scenario (left panel of~\ref{fig:sp_optax_mp}). All runs in Table \ref{tab:sp_optax_mp} are terminated once the absolute and relative tolerances of \([\text{ATOL}, \text{RTOL}] = [10^{-7}, 10^{-7}]\) for the $L_{\infty}$ norm of the gradient of the loss function are met. Consequently, the number of iterations reported for the BFGS optimizers corresponds to the step where the specified tolerance is reached. The runtime for each case is also recorded in Table \ref{tab:sp_optax_mp}. For all runs, the runtime is less than one minute, except for the run where Adam is used for a larger number of iterations Figure (\ref{fig:sp_optax_mp}).

In Scenario 2, where the GPU is fully utilized with collocation points and the PINN employs a smaller neural network, the convergence history for single precision arithmetic is shown in Figure~\ref{fig:sp_optax_cp}. Convergence was achieved using a neural network with four hidden layers, each containing 20 neurons, a \(\tanh\) activation function, 250 randomly sampled data points for initial and boundary conditions, and 50,000 collocation points. The performance metrics, including the relative \(L_2\) error and training time for single and double precision, are summarized in Table~\ref{tab:sp_optax_cp}.

\section{ Lorenz system}\label{appendix:Lorenz system}

Considering the the Lorenz system, as described in \cite{LorenzSystem}. It is governed by the following set of coupled ordinary differential equations:

\begin{align}\label{Lorenze_system}
    \frac{dx}{dt} &= \sigma (y - x), \\
    \frac{dy}{dt} &= x(\rho - z) - y, \\
    \frac{dz}{dt} &= xy - \beta z,
\end{align}

where the parameters of the Lorenz system are given as \( \sigma = 10 \), \( \rho = 28 \), and \( \beta = \frac{8}{3} \). The system continues to display chaotic behavior \cite{hirsch2013differential}. The solution is computed up to \( t = 20 \), starting from the initial condition \( \left( x(0), y(0), z(0) \right) = (1, 1, 1) \).

The neural network architecture consists of three hidden layers, each containing 30 neurons. The hyperbolic tangent (tanh) activation function is applied across all hidden layers to capture the system's inherent nonlinearity. 
Training begins with the Adam optimizer, following an exponential decay schedule with an initial learning rate of \( 5 \times 10^{-3} \), which decays by a factor of 0.98 every 1000 iterations. Subsequently, the BFGS and SSBroyden optimizers, incorporating Wolfe line-search, are employed for further optimization.

\begin{figure}[!tbh]
  \begin{minipage}[b]{\linewidth}
    \centering
    \includegraphics[width=0.95\textwidth]{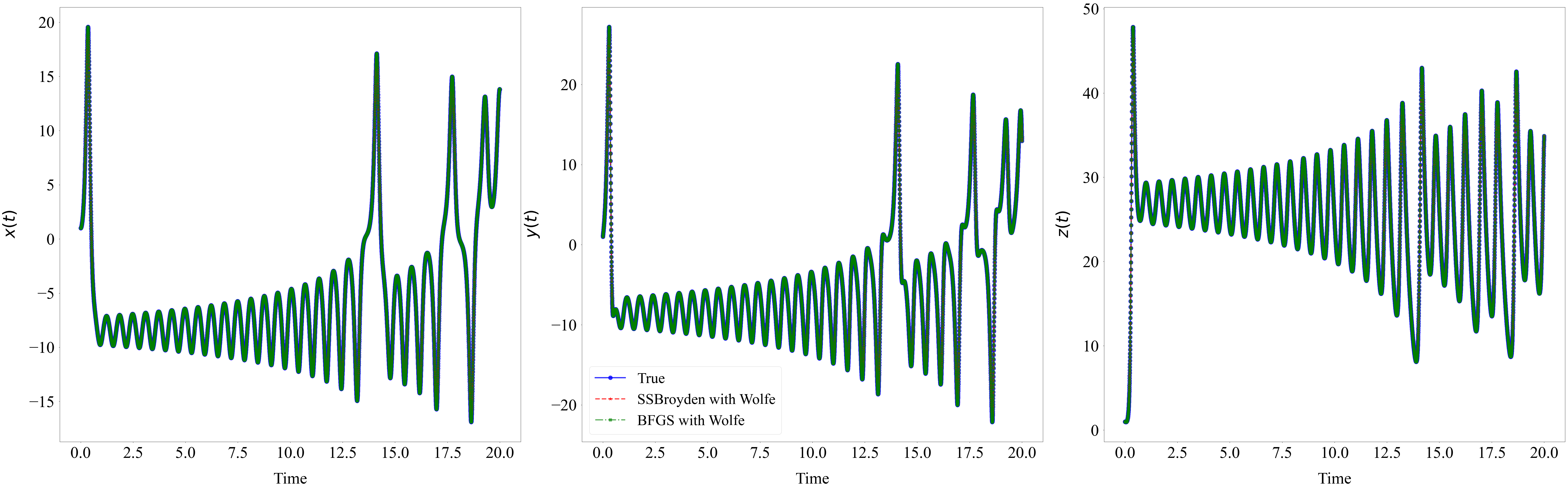}
    \captionof{figure}{\textbf{Double-Precision PINNs for the Lorenz system:} Comparison of PINNs predictions using BFGS and SSBroyden with Strong Wolfe line-search against the numerical solution for the Lorenz system. The plots display \( x(t) \), \( y(t) \), and \( z(t) \) over time.}
    \label{fig:LORENZ_SYSTEM_prediction}
  \end{minipage}\hfill
  \begin{minipage}[b]{\linewidth}
    \centering
    \renewcommand{\arraystretch}{1.3} 
    \arrayrulecolor{black} 
    {\rowcolors{2}{cyan!15}{white} 
    \scalebox{0.85}{ 
    \begin{tabular}{|c|c|c|}
    \hline
    \rowcolor{cyan!40} 
    \textbf{Metric} & \textbf{BFGS with Wolfe} & \textbf{SSBroyden with Wolfe} \\ \hline
    Relative \( l_2 \) error for \( x(t) \) & \( 6.74 \times 10^{-4} \) & \( 9.65 \times 10^{-5} \)  \\ \hline
    Relative \( l_2 \) error for \( y(t) \) & \( 9.81 \times 10^{-4} \) & \( 1.40 \times 10^{-4} \)  \\ \hline
    Relative \( l_2 \) error for \( z(t) \) & \( 4.15 \times 10^{-4} \) & \( 5.94 \times 10^{-5} \)  \\ \hline
    Training time (s)         & 4541.61            & 5022.95            \\ \hline
    \end{tabular}}}
    
    \captionof{table}{\textbf{Double-Precision PINNs for the Lorenz system:} Double-Precision PINN results for the Lorenz system: \( L_2 \) Relative Errors and training times for PINN predictions obtained using BFGS and SSBroyden over 40 time windows in \( [0, 20] \).}
    \label{tab:L2errors_LORENZ_SYSTEM}
  \end{minipage}
\end{figure}

To address the extended time horizon \( [0, 20] \), the time domain is partitioned into 40 time windows, each of length \( \Delta t = 0.5 \). A separate PINNs is trained for each window, using the final state of the previous window as the initial condition for the next.
Figure~\ref{fig:LORENZ_SYSTEM_prediction} compares the numerical solution of the Lorenz system with PINNs predictions obtained via BFGS and SSBroyden over 40 time windows in \( [0, 20] \). The plots show \( x(t) \) (left), \( y(t) \) (middle), and \( z(t) \) (right). Table~\ref{tab:L2errors_LORENZ_SYSTEM} summarizes the $ L_2 $ relative errors for predicting \( x(t) \), \( y(t) \), and \( z(t) \) across all 40 time windows and reports the corresponding training times for BFGS and SSBroyden.

\section{ Multi-dimensional Rosenbrock function}\label{appendix:rosenbrock}

Here, we consider the non-quadratic objective functions Rosenbrock, which is commonly used to evaluate optimization algorithms,  defined in two-dimensions as:

\[
f(x_1, x_2) = 100(x_2 - x_1^2)^2 + (x_1 - 1)^2.
\]

The SSBroyden method with Strong Wolfe line-search demonstrates performance comparable to SSBFGS and BFGS with Strong Wolfe line-search, with the optimization starting from \( x_i = 0.5 \). When applied to challenging optimization problems, including the Rosenbrock function—characterized by a narrow, curved valleys and multi-modal landscapes- SSBroyden efficiently converges to the global minimum with accuracy similar to BFGS. 

\begin{figure}[!tbh]
  \begin{minipage}[b]{\linewidth}
    \centering
    \includegraphics[trim={3cm 0 0 0},clip, width=0.9\textwidth]{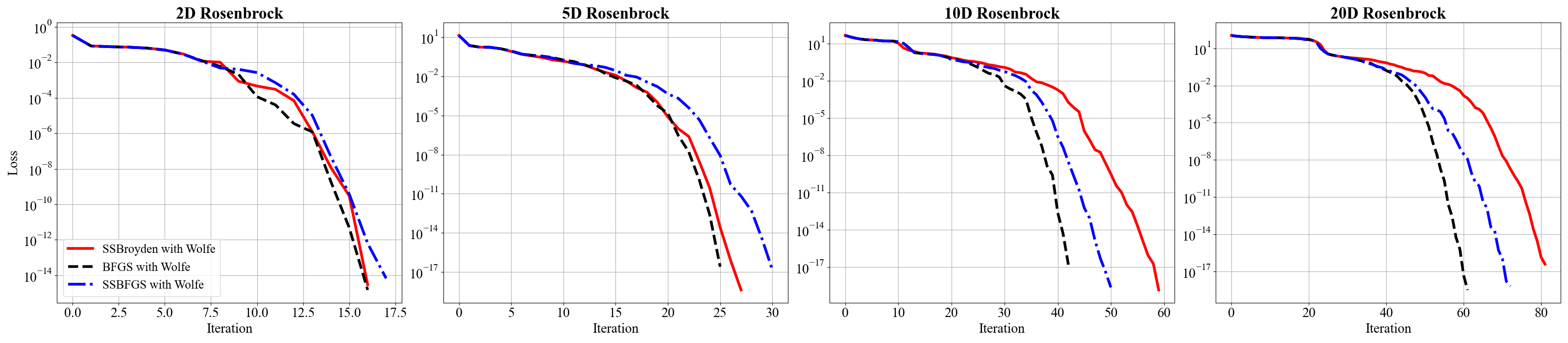}
    \captionof{figure}{\textbf{Convergence comparison} of BFGS, SSBroyden, and SSBFGS with Strong Wolfe line-search on the Rosenbrock function across multiple dimensions (2D, 5D, 10D, and 20D).}
    \label{fig:Rosenbrock_iteration}
  \end{minipage}\hfill

  \vspace{0.6cm}

  \begin{minipage}[b]{\linewidth}
    \centering
    \renewcommand{\arraystretch}{1.3}
    \arrayrulecolor{black} 
    {\rowcolors{2}{cyan!15}{white}
    \scalebox{0.85}{ 
    \begin{tabular}{|c|l|c|p{6.8cm}|l|}
    \hline
    \rowcolor{cyan!40} 
    \textbf{Dimension} & \textbf{Optimizer, Line-search} & \textbf{Iterations} & \textbf{Optimized parameters} & \textbf{Final Loss} \\
    \hline
    2 & BFGS with Wolfe & 17 & {[}1.0, 1.0{]} & $1.50 \times 10^{-15}$ \\
    2 & SSBroyden with Wolfe & 17 & {[}1.0, 1.0{]} & $2.97 \times 10^{-15}$ \\
    2 & SSBFGS with Wolfe & 19 & {[}1.0, 1.0{]} & $3.42 \times 10^{-20}$ \\
    2 & Gradient Descent & 5000 & {[}0.9689, 0.9386{]} & $9.71 \times 10^{-4}$ \\
    2 & Adam & 3899 & {[}1.0, 1.0{]} & $1.24 \times 10^{-12}$ \\
    \hline
    5 & BFGS with Wolfe & 26 & {[}1.0, 1.0, 1.0, 1.0, 1.0{]} & $2.56 \times 10^{-17}$ \\
    5 & SSBroyden with Wolfe & 27 & {[}1.0, 1.0, 1.0, 1.0, 1.0{]} & $7.79 \times 10^{-17}$ \\
    5 & SSBFGS with Wolfe & 31 & {[}1.0, 1.0, 1.0, 1.0, 1.0{]} & $1.44 \times 10^{-17}$ \\
    5 & Gradient Descent & 5000 & {[}0.9970, 0.9939, 0.9879, 0.9759, 0.9522{]} & $7.78 \times 10^{-4}$ \\
    5 & Adam & 5000 & {[}1.0, 1.0, 1.0, 1.0, 1.0{]} & $2.77 \times 10^{-11}$ \\
    \hline
    10 & BFGS with Wolfe & 43 & {[}1.0, ..., 1.0{]} & $1.53 \times 10^{-17}$ \\
    10 & SSBroyden with Wolfe & 57 & {[}1.0, ..., 1.0{]} & $1.31 \times 10^{-15}$ \\
    10 & SSBFGS with Wolfe & 49 & {[}1.0, ..., 1.0{]} & $5.19 \times 10^{-17}$ \\
    10 & Gradient Descent & 5000 & {[}0.9999, ..., 0.9421{]} & $1.15 \times 10^{-3}$ \\
    10 & Adam & 5000 & {[}0.9998, 1.0011, ..., 1.0002{]} & $3.77 \times 10^{-3}$ \\
    \hline
    20 & BFGS with Wolfe & 60 & {[}1.0, ..., 1.0{]} & $6.35 \times 10^{-16}$ \\
    20 & SSBroyden with Wolfe & 81 & {[}1.0, ..., 1.0{]} & $1.40 \times 10^{-16}$ \\
    20 & SSBFGS with Wolfe & 70 & {[}1.0, ..., 1.0{]} & $4.90 \times 10^{-16}$ \\
    20 & Gradient Descent & 5000 & {[}1.0, ..., 0.9312{]} & $1.64 \times 10^{-3}$ \\
    20 & Adam & 5000 & {[}1.0, ..., 0.9999{]} & $6.57 \times 10^{-7}$ \\
    \hline
    \end{tabular}}}    
    \captionof{table}{Final optimization results for the Rosenbrock function using various methods across dimensions 2D, 5D, 10D, and 20D. All quasi-Newton methods recover the global minimum to machine precision. First-order methods (GD, Adam) either converge slowly or stall at suboptimal values in higher dimensions.}
    \label{tab:Rosenbrock_results}
  \end{minipage}
\end{figure}

Figure~\ref{fig:Rosenbrock_iteration} presents a comparison of optimization performance on the Rosenbrock function across four different dimensionalities: 2D, 5D, 10D, and 20D.
To ensure a fair comparison, all optimizers were initialized with the same starting point ($\mathbf{x}_0 = [0.5, \dots, 0.5]$), and the stopping criterion was uniformly defined as the gradient norm falling below $10^{-6}$ or a maximum of 5000 iterations. Both methods used the same initial inverse Hessian approximation ($H_0 = I$) and consistent optimization parameters. The plots highlight the convergence behavior of each method across increasing dimensions, illustrating how the number of iterations and the rate of loss decay vary with problem size. 
The number of iterations required to reach the global minimum increases as the dimensionality of the problem grows. However, for the 20-dimensional case, SSBFGS and SSBroyden requires more iterations to reach global optimization compared to BFGS. While, for the 2D case, BFGS, SSBFGS and SSBroyden require approximately the same number of iterations to achieve the global minimum.

\clearpage
\bibliography{sample}

\end{document}